%% file: telo-main.tex
\documentclass[manuscript,screen,authordraft,nonacm,review=false,timestamp=false]{acmart}
\AtBeginDocument{%
  \providecommand\BibTeX{{%
    \normalfont B\kern-0.5em{\scshape i\kern-0.25em b}\kern-0.8em\TeX}}}

\setcopyright{acmlicensed}
\copyrightyear{2018}
\acmYear{2018}
\acmDOI{XXXXXXX.XXXXXXX}

\usepackage{ amsmath, amsfonts, amsthm,amsbsy}
\usepackage{booktabs} 
\usepackage{graphicx}
\usepackage{bm}
\usepackage{color,colortbl}
\usepackage{float}
\usepackage{cmap}
\usepackage{algorithmicx}
\usepackage{algorithm}
\usepackage{algpseudocode} 
\usepackage{dsfont}
\usepackage{multirow}
\usepackage{placeins}
\usepackage{soul}

\usepackage{enumitem}
\usepackage{tikz}
\usetikzlibrary{patterns}

\usepackage{hyperref}

\newcommand\DrawBox[3][]{%
  \tikz[remember picture,overlay]\draw[ draw=blue!50,  pattern=north west lines, pattern color= blue!30][#1]  (-3.3,-3.3) rectangle (9.1,-0.15);}
\newcommand\DrawBoxed[3][]{%
  \tikz[remember picture,overlay]\draw[ draw=green!60, pattern=north east lines, pattern color= green!35][#1]  (-1.05,-2) rectangle  (11.35,-0.06);}
\newcommand\DrawBoxes[3][]{%
  \tikz[remember picture,overlay]\draw[ draw=yellow,  pattern=horizontal lines, pattern color= yellow!45, fill opacity=0.3][#1]  (-1.5,-1.4) rectangle (11.4,-0.1);}
  \newcommand\DrawBoxesb[3][]{%
  \tikz[remember picture,overlay]\draw[ draw=yellow,  pattern=horizontal lines, pattern color= yellow!45][#1]  (-7.5,-1.05) rectangle (5.9,-0.1);}
  \newcommand\DrawBoxedb[3][]{%
  \tikz[remember picture,overlay]\draw[ draw=green,  pattern=north east lines, pattern color= green!35, fill opacity=0.30][#1]  (-3.9,-0.9) rectangle  (8.9,-0.05);}
  \newcommand\DrawBoxedc[3][]{%
  \tikz[remember picture,overlay]\draw[ draw=red!40,  pattern=grid, pattern color= red!30][#1]  (-0.,-0.3) rectangle  (0.75,0.5);}
  \newcommand\DrawBoxedd[3][]{%
  \tikz[remember picture,overlay]\draw[ draw=gray!60, pattern=dots, pattern color= gray!50][#1]  (2.2,-0.25) rectangle  (12.8,0.45);}

\begin{document}

\title{Analyzing Design Principles for Competitive Evolution Strategies in Constrained Search Spaces}

\author{Michael Hellwig}
\email{michael.hellwig@fhv.at}
\orcid{0000-0002-6731-8166}

 \author{Hans-Georg Beyer}
 \email{hans-georg.beyer@fhv.at}
  \orcid{0000-0002-7455-8686}
\affiliation{
  \institution{Vorarlberg University of Applied Sciences, Research Centre Business Informatics, Josef Ressel Centre for Robust Decision Making}
  \streetaddress{Campus V, Hochschulstrasse 1}
  \postcode{6850}
  \city{Dornbirn}
  \country{Austria}
}

\renewcommand{\shortauthors}{M. Hellwig and H.-G. Beyer}

\begin{abstract}
In the context of the 2018 IEEE Congress of Evolutionary Computation, the Matrix Adaptation Evolution Strategy for constrained optimization turned out to be notably successful in the competition on constrained single objective real-parameter optimization. Across all considered instances the so-called $\epsilon$MAg-ES achieved the second rank. However, it can be considered to be the most successful participant in high dimensions. Unfortunately, the competition result does not provide any information about the modus operandi of a successful algorithm or its suitability for problems of a particular shape. To this end, the present paper is concerned with an extensive empirical analysis of the $\epsilon$MAg-ES working principles that is expected to provide insights about the performance contribution of specific algorithmic components.
To avoid rankings with respect to insignificant differences within the algorithm realizations, the paper additionally introduces significance testing into the ranking process. 
\end{abstract}





\maketitle

\section{Introduction}
\label{sec01}
Constrained optimization problems are found in a vast 
number of real-world applications that deal with unknown parameters, e.g. Financial Mathematics,
 Operations Research, Engineering, or Machine Learning. Due to the rising complexity of the corresponding models, the analytical derivation of solutions is only rarely possible. 
In such situations, efficient
 and effective numerical methods are required to provide reasonable
 solutions.

Evolutionary Algorithms (EAs) can be considered successful heuristic approaches for this kind of tasks~\cite{collange2010multidisciplinary,Zhang2011,mora2015applications}. 
These algorithms do not make any assumption about the underlying problem features. Yet, due to their inherent working principles gleaned from nature, EAs are in many cases able to realize great performance where classical methods are stretched to their limits.

However, the success of probabilistic search heuristics depends strongly on the problem-specific fitness environment and is usually hard to predict. This is particularly true in the field of real-valued constrained optimization, where the theoretical background is comparably scarce. 
A theoretical performance analysis is often manageable only on rather simple test problems that are regarded to emulate the original problem locally. On the other hand, the algorithm development on a specific real-world problem must be considered costly.
For this reason, benchmark environments play an important role in the development of new algorithmic ideas.
They are essential for the assessment and the comparison of contemporary algorithms. 
To this end, benchmarks are considered to establish well-defined experimental settings that aim at providing reproducible and comparable algorithmic results.

The state-of-the-art benchmark environments for EAs dedicated to constrained optimization have been introduced in the context of the \emph{Competition on Constrained Single Objective Real-Parameter Optimization} which is regularly organized during the \emph{IEEE Congress on Evolutionary Computation (CEC)}. Until today, the constrained CEC benchmarks are used for algorithm testing in a large number of publications. 
The latest collection of constrained test functions and ranking principles was introduced for CEC 2017, cf.~\cite{CEC2017}. For brevity, we will refer to this benchmark environment as the CEC2017 benchmarks.

Due to the success of various Differential Evolution (DE) methods in the context of the constrained CEC benchmarks, DE is considered to be especially useful EA variants for constrained optimization problems. Yet, there is also progress in the EA subclass of Evolution Strategies (ES). A novel MA-ES variant for constrained optimization~\cite{HellwigB2018} is applied to the benchmark problems specified for the CEC 2018 Competition on Constrained Single Objective Real-Parameter Optimization. By the combination of well-known constraint handling techniques with the Matrix Adaptation Evolution Strategy (MA-ES)~\cite{BeyerS2017}, a novel ES for constrained optimization, the $\epsilon$MAg-ES has been presented. The algorithm turned out to be quite successful on the CEC2017 problems and was ranked runner-up in the respective competition. While the algorithm can find feasible solutions on more than 80\% of the CEC benchmark problems with high accuracy, the reasons for its success have not been disclosed within the conference paper~\cite{HellwigB2018}, nor in the report on the competition results~\cite{CEC2018comp}.  

Only recently, several papers are submitted that build upon the $\epsilon$MAg-ES. These papers simply vary the repair operator or the selection scheme and claim to represent reasonable advancements of the original strategy. 
However, these papers usually also remain short on the investigation of the algorithmic component's contribution to the overall performance.
While this is in order in a competition paper, a full paper publication should also give some insight into the working mechanisms of an algorithm. Consequently, obtaining detailed insight into the $\epsilon$MAg-ES is not only desirable for an increase in knowledge, but also
for the identification of true strategy advancements.

According to~\cite{HellwigB2019a}, the CEC benchmarks reveal some drawbacks concerning the interpretation of the algorithm results.
The performance indicators only provide a measure of algorithm effectiveness, while efficiency (e.g. in terms of running time) is not considered. Being organized in large tables the performance indicators of two distinct algorithms are difficult to analyze.
The recommended ranking procedure depends strongly on the number of the compared algorithms.
Moreover, the aggregation of algorithm results over different dimensions in the final ranking can cause a loss of information. 
This might be appropriate for a competition, but impedes the investigation of an algorithm for its strengths and weaknesses on certain problems.
To deal with this flaw, the present investigation will additionally take into account the algorithmic mean value dynamics on specific constrained problems.
The empirical $\epsilon$MAg-ES analysis aims at  
\begin{enumerate}[label=(\roman*)] 
 \item  providing a detailed assessment of the contribution of single components to the $\epsilon$MAg-ES performance on the CEC2017 benchmarks problems,
 \item  identifying problem features that correspond to particularly good or bad search behavior
 \item  determining beneficial strategy parameter settings for the $\epsilon$MAg-ES on the constrained CEC2017 benchmarks, and 
 \item  demonstrating a sound way of algorithm assessment that can be adopted in future studies. 
\end{enumerate}

The remainder of the paper is organized as follows:
Section~\ref{sec02} gives the general problem formulation used in the constrained CEC2017 benchmarks as well as some related definitions.
In Sec.~\ref{sec03} the $\epsilon$MAg-ES is described in detail and its core components are highlighted. Afterward, the experimental investigation in Sec.~\ref{sec04} compares the performances of different algorithm increments on the whole CEC2017 benchmark suite as well as on particular subgroups of constrained problems.  In Sec.~\ref{sec05} the algorithm dynamics are examined on six individual problems in greater detail. This is followed by an empirical examination of the $\epsilon$MAg-ES strategy parameter configuration in Sec.~\ref{sec06}. In particular, this paper considers the choice of an appropriate offspring population size and the number of consecutive repair repetitions of the Jacobian-based repair step, respectively.
The paper is concluded in Sec.~\ref{disc} with a thorough discussion of the empirical observations and provides suggestions for future research directions.

\section{General problem formulation}
\label{sec02}
The constrained test problems considered in the present paper have been introduced in the technical report~\cite{CEC2017} for the CEC2017 competition on constrained optimization. They have the general form
  \begin{equation}
     \begin{aligned}
       \min   \quad & f(\bm{y})  & &\\
        s.t.   \quad & g_i(\bm{y}) \leq 0,  & i=1,\dots,l, &\\
	     & h_j(\bm{y}) = 0,     & j=1,\dots,k, &\\
	     & \bm{y} \in S \subseteq \mathds{R}^N. & & 
      \end{aligned}
      \label{cop}
    \end{equation} 
 Without loss of generality, the optimization goal is 
the minimization of the real-valued objective function $f(\bm{y})$.
Here, $\bm{y} \in S$ denotes the $N$-dimensional search space parameter vector. 
The set $S$ usually comprises several box constraints specifying reasonable 
intervals of the parameter vector components.
We refer to the $N$-dimensional vectors that specify the lower and upper box constraints of each parameter component as $\bm{\check{y}}$, and $\bm{\hat{y}}$, respectively.
Additionally, the feasible region of the search space is restricted by $m=l+k$ real-valued constraint functions.
These constraint functions are separated into inequality constraints $g_i(\bm{y}), \: i=1,\dots,l$ and 
equality constraints $h_j(\bm{y}), \: j=1,\dots,k$.
A vector $\bm{y}\in S$ that satisfies all constraints simultaneously is called feasible. The set of all feasible parameter vectors is denoted
\begin{equation}
 \label{feasSet}
  M \coloneqq \left\{ \bm{y} \in S \colon g_i(\bm{y}) \leq 0 \wedge  h_j(\bm{y}) = 0, \forall i,j \right\}. 
 \end{equation}
The global optimum of~\eqref{cop} is denoted by $\bm{y}^* \in M$.
Note, that the objective function $f(\bm{y})$ subject to the related constraints {\color{black}will in the following be} referred to as \textit{constrained function}, cf. Eq.~\eqref{cop}.

When considering problem~\eqref{cop} a measure of infeasibility is useful for ranking potentially infeasible candidate solutions.
To this end, we compute the constraint violation $\nu({\bm{y}})$ of a candidate solution $\bm{y}$ as 
  \begin{equation}
   \nu({\bm{y}}) = \sum_{i=1}^{l} G_i(\bm{y}) + \sum_{j=1}^k H_j(\bm{y}),
   \label{violation}
  \end{equation}
  with functions $G_i(\bm{y})$ and $H_j(\bm{y})$ defined by
  \begin{equation}
    G_i(\bm{y}) \coloneqq \max\left(0,g_i(\bm{y})\right),
  \end{equation}
  and
  \begin{equation}
    		  H_j(\bm{y})\coloneqq \left\{ \begin{matrix} \lvert h_j(\bm{y})\rvert, &\textrm{if} \:\: \lvert h_j(\bm{y})\rvert-\delta > 0\\
							      0, &\textrm{if} \:\: \lvert h_j(\bm{y})\rvert-\delta \leq 0\\ 
							 \end{matrix} \right. .
							\label{eqcon}
  \end{equation}
	To be able to satisfy the equality constraints, the $\delta$ term introduces the necessary error margin. 
	This paper considers the CEC2017 standard recommendation of $\delta = 10^{-4}$.
 
   In accordance to the CEC2017 specifications, the corresponding mean constraint violation $\overline{\nu}({\bm{y}})$ is calculated as the quotient of the constraint violation and the number of constraint functions
  \begin{equation}
   \overline{\nu}({\bm{y}}) = \cfrac{\sum_{i=1}^{l} G_i(\bm{y}) + \sum_{j=1}^k H_j(\bm{y})}{l+k}.
	\label{eqcon2}
  \end{equation}
    
  {
  As candidate solutions are associated with their individual fitness (objective function) as well as constraint violation values, 
  there is need of an order relation suitable to distinguish their quality.
  Considering two candidate solutions $\bm{y}_i\in\mathds{R}^N$ and $ \bm{y}_j\in\mathds{R}^N$ of problem~\eqref{cop}, these are compared as follows
    \begin{equation}
	    \begin{split}
	     \bm{y}_i  \leq_{lex} \bm{y}_j   
		      \Leftrightarrow  \left\{ \begin{matrix}
			       f(\bm{y}_i) \leq f(\bm{y}_j), & \textrm{if }& \nu(\bm{y}_i) =\nu(\bm{y}_j), \\
                               \nu(\bm{y}_i) < \nu(\bm{y}_j), & & \textrm{otherwise. }
                         \end{matrix}  \right.
              \end{split}
                         \label{lexo}
	    \end{equation}
    That is, the ordering $\leq_{lex}$ primarily ranks two candidate solutions according to their constraint violations and secondly with respect to their objective function values. We refer to the order relation~\eqref{lexo} as lexicographic ordering. Note that this order relation is also commonly known as \emph{superiority of feasibility} or \emph{Deb's rule}~\cite{DEB2000}.
   }
    
   {\color{black} According to the competition guidelines~\cite{CEC2018comp}, a budget of $20000 N$ function evaluations is available for each constrained problem. Each evaluation of a complete constrained function~\eqref{cop} is counted as a single function evaluation. That is, it is irrelevant whether the algorithm uses only the values of the target function and/or the values of the constraint functions.}

\input{telo-body}

\begin{acks}
The financial support by the Austrian Federal Ministry of Labour and Economy, the National Foundation for Research, Technology, and Development and the Christian Doppler Research Association is gratefully acknowledged.
\end{acks}

\bibliographystyle{ACM-Reference-Format}
\bibliography{DEreview,mybib}

\newpage

\appendix
\section*{Supplementary Material}

\input{telo-appendix}

\end{document}

%% file: telo-body.tex
\section{Algorithm description}
\label{sec03}
  This section recaps the $\epsilon$MAg-ES for constrained real-parameter optimization~\cite{HellwigB2018}. In the context of the 2018 IEEE CEC  Competition on Constrained Single Objective Real-Parameter Optimization, this algorithm finished runner-up and excelled in higher dimensions. Its pseudo-code is displayed in Algorithm~\ref{alg_MAES}. The strategy is based on the MA-ES~\cite{BeyerS2017} which represents an algorithmically simplified CMA-ES variant with comparable performance in unconstrained environments. To deal with constrained problems, three constraint handling techniques are incorporated into the MA-ES. These techniques include a method for treating box-constraints, a specific ordering relation
  as well as a projection method to prevent the algorithm from premature convergence. Below, all techniques are presented in detail. Their  impact on the $\epsilon$MAg-ES performance is examined in Sec.~\ref{sec04}.
  
  Given the black-box scenario of the competition, the algorithm has no knowledge about the appropriate step-size at a random location in the search space.  Instead of starting from a single point, the algorithm initially samples a uniformly distributed population $\mathcal{P}$ of $\lambda$ candidate solutions $\bm{y}_j \in \mathbb{R}^N, \: j=1,\dots,\lambda$ within the predefined box-constraints $\bm{\check{y}}$ and $\bm{\hat{y}}$, respectively.
  {
  }
  \begin{algorithm}[h]
  \caption{Initialization of the starting population within the given box-constraints.}
   \begin{algorithmic}[]
		\For {$j \gets 1 \colon \lambda$} 
  \State $\bm{y}_j \gets \bm{\check{y}} + (\bm{\hat{y}}-\bm{\check{y}}) \circ \bm{u}(0,1)$
  \State $\mathcal{P} \gets \mathcal{P} \cup \{\bm{y}_j\}$
  \EndFor
  \end{algorithmic}  
  \label{popInit}
  \end{algorithm}
  
  The initial parental recombinant $\bm{y}^{(0)}$ is then obtained by weighted recombination of the $\mu$ best candidate solutions in line 5.\footnote{Notice, that $\bm{y}_{m;\lambda}$ denotes the $m$th best out of $\lambda$ candidate solutions with respect to the order relation $\leq_{\epsilon}$, see Eq.~\eqref{epso}. }
  The standard weights $w_i$ of the MA-ES as described in~\cite{BeyerS2017} are used.

  \begin{algorithm}[htbp]
 \caption[Pseudo code of the $\epsilon$MAg-ES.]{ Pseudo code of the Matrix Adaptation Evolution Strategy variant for constrained single objective real-parameter optimization: 
    the $\epsilon$MAg-ES. }
  \begin{algorithmic}[1]
  \State \textbf{Initialize: } Generate starting population within the box-constraints, see Alg.~\ref{popInit}.
  \State \textbf{Initialize: } $\mu$, $\lambda$, $\sigma^{(0)}$, $\bm{p}_\sigma^{(0)} \gets \bm{0}$  , $\bm{M}^{(0)} \gets\bm{I}$,  $g\gets 0$ , $T$\DrawBoxesb{a}{b}
   \State $\epsilon^{(0)} \gets \sum_{i=1}^{\lfloor \theta_t \lambda \rfloor} {\nu}(\bm{y}_{i;\lambda}) / \lfloor \theta_t \lambda \rfloor  $ 
   \Comment{$\epsilon$ threshold initialization}
  \State $\gamma \gets \max\left(\gamma_{\min},(-5 - \log(\epsilon^{(0)}) ) / \log(0.05) \right) $
	\State $\bm{y}^{(0)} \gets \sum_{i=1}^\mu{w_i\bm{y}_{i;\lambda}}$ \Comment{Recombination according to $ \leq_\epsilon $}
	  \State $ fevals \gets \lambda$
	\State $\bm{y}_\textrm{bsf} \gets \bm{y}_{1;\lambda}  $ \Comment{Initialize best solution found}
	\While {$fevals < fevals_{\max}$} \DrawBoxedb{a}{b}
	  \State $\bm{M}^{-1} \gets \texttt{PseudoInverse}(\bm{M}^{(g)})$
		\State Reset $\bm{M}^{(g)} \gets \bm{I} $ if $\texttt{PseudoInverse}(\bm{M}^{(g)})$ fails
	  \For {$l \gets 1\colon \lambda$} { $fevals \gets fevals +1$}
	  \State $\bm{z}_l^{(g)} \gets \mathcal{N}(\bm{0},\bm{I})$
	  \State $ \bm{d}_l^{(g)} \gets \bm{M}^{(g)}\bm{z}_l^{(g)}$
	    \State $\bm{\bar{y}} \gets \bm{y}^{(g)} + \sigma^{(g)} \bm{d}_l^{(g)}  $
	    \State $\bm{y}_l^{(g)} \gets \texttt{KeepRange}(\bm{\bar{y}})$
	    \DrawBox{a}{b} \If {$ \textrm{mod}(g, N) = 0 \: \wedge \: u(0,1)<\theta_p$} {$h\gets 1$ } \Comment{Repair step}   
		  \While {$ h \leq \theta_r \wedge \nu(\bm{y}_l^{(g)}) > 0$} { $h \gets h+1$} 
		    \State $\bm{\tilde{y}} \gets \texttt{GradientBasedRepair}(\bm{y}_l^{(g)})$
		    \State $\bm{y}_l^{(g)} \gets \texttt{KeepRange}(\bm{\tilde{y}})$
		    \State $fevals \gets fevals + N+1 $
		  \EndWhile
	      \EndIf \DrawBoxed{a}{b}
         \If {$\bm{\bar{y}} \neq  \bm{y}_l^{(g)}$ } {} \Comment{Recalculation of mutation vectors}
	         \State $ \bm{d}_l^{(g)} \gets \big(\bm{y}_l^{(g)} - \bm{y}^{(g)}\big) / \sigma^{(g)}  $
	         \State $ \bm{z}_l^{(g)} \gets {\bm{M}^{-1}} \bm{d}_l^{(g)}  $
	      \EndIf
	  \EndFor
		\If {$ \bm{y}_{1;\lambda}^{(g)}  \leq_{\epsilon} \bm{y}_\textrm{bsf}$} {$\bm{y}_\textrm{bsf} \gets \bm{y}_{1;\lambda}^{(g)} $ }  \Comment{Update best solution found}
		\EndIf
		\State $\bm{y}^{(g+1)} \gets \bm{y}^{(g)} + \sigma^{(g)} \sum_{i=1}^{\mu} {w_i \bm{d}_{i;\lambda}^{(g)}}$ \Comment{Selection and recombination according to $ \leq_\epsilon $}
		\State $\bm{p}_\sigma^{(g+1)} \gets (1-c_\sigma) \bm{p}_\sigma^{(g)} + \sqrt{\mu_w c_\sigma (2-c_\sigma)} \sum\limits_{i=1}^{\mu} {w_i \bm{z}_{i;\lambda}^{(g)}} $ \Comment{Search path update}
\State \DrawBoxedd{r}{s}$\bm{M}^{(g+1)} \gets \bm{M}^{(g)}\left( \bm{I} + \frac{c_1}{2} \left[ \bm{p}_\sigma^{(g)} (\bm{p}_\sigma^{(g)})^\top - \bm{I} \right]  +\frac{c_\mu}{2} \left[ \sum\nolimits_{i=1}^\mu w_i \bm{z}_{i;\lambda}^{(g)} (\bm{z}_{i;\lambda}^{(g)})^\top  -\bm{I}\right]\right) $\Comment{$M$ update}
\State $\sigma^{(g+1)} \gets \min \Big( \sigma^{(g)}\textrm{exp}\left[\frac{c_\sigma}{2}\left(\frac{\left\|\bm{p}^{(g+1)}_\sigma\right\|^2}{N} -1\right)\right]   ,\DrawBoxedc{r}{s} \:\sigma_{\max} \: \Big)$ \Comment{Mutation strength update}
	\State $g\gets g+1$\DrawBoxes{a}{b}
	\If {$g > T$} { $\epsilon^{(g)} \gets 0$}\Comment{Update $\epsilon$ threshold} 
	\Else  { $\epsilon^{(g)} \gets \epsilon^{(0)} (1-\frac{g}{T})^{\gamma} $ }
      \EndIf
	\EndWhile 
	\State \Return $\left[ \bm{y}_\textrm{bsf} , \:f(\bm{y}_\textrm{bsf}) ,\: \nu(\bm{y}_\textrm{bsf}) \right]$
	\end{algorithmic}
  \label{alg_MAES}
\end{algorithm}

  Regarding $\bm{z}_l^{(g)}$ and $\bm{d}^{(g)}_l$, the vector that contributes to the $m$th best candidate solution $\bm{y}_{m;\lambda}$ is considered the $m$th best, i.e. $\bm{z}_{m;\lambda}$ and $\bm{d}_{m;\lambda}$, respectively.
  The comparison of the candidate solutions always involves the evaluation of the constrained problem and consumes function evaluations. 
  {\color{black} As suggested in the technical report~\cite{CEC2018comp}, we account one function evaluation per call to the constrained function. That is, one function evaluation is consumed each time a candidate solution is evaluated regardless of whether only the objective function, only the constraints, or all components of the constrained function are to be used.
  }

  The selection of the $\mu$ best among $\lambda$ candidate solutions is based on the so-called $\epsilon$-level order relation~\cite{TakahamaS06}.
  The order relation $\leq_{\epsilon}$ is described in more detail in Sec.~\ref{epslevel}. All parts of Alg.~\ref{alg_MAES} that are related to the control of the $\epsilon$-level are highlighted by yellow boxes.
  The starting population is also used to determine the initial $\epsilon^{(0)}$ value in line 3, and the associated parameter $\gamma$, in line 4, that controls the $\epsilon^{(g)}$ decrease. Here, $\theta_t$ specifies the percentage of considered candidate solutions and $\left\lfloor \cdot \right\rfloor$ denotes the floor function. 
  In line 7, the best individual of the initial population $\bm{y}_{1;\lambda}$ becomes the best solution found so far $\bm{y}_{\textrm{bsf}}$.

  Contrary to CMA-ES, the MA-ES replaces the covariance matrix update as well as the adaptation of the related search path by an updated transformation matrix $\bm{M}^{(g)}$. In the offspring procreation loop of each generation $g$, $\lambda$ offspring are generated out of the recombined $\mu$ best candidate solution of the previous generation. To this end, the mutation direction vector $\bm{d}^{(g)}_l$ of each offspring is obtained by multiplication of this matrix $\bm{M}^{(g)}$ and a vector $\bm{z}_l^{(g)}$ with standard normally distributed components, lines 12 and 13. By adding the product of the mutation strength $\sigma^{(g)}$ and $\bm{d}^{(g)}$ to the recombinant $\bm{y}^{(g)}$ of the previous generation, an offspring is generated in line 14.

  In cases where offspring individuals are generated outside the box constraints, these candidate solutions are reflected inside the box. According to Eq.~\eqref{reflect}, the routine $\texttt{KeepRange}(.)$ recomputes each offspring that does not satisfy all box constraints.
  Regarding the upper and lower parameter bounds ($\bm{\hat{y}},\bm{\check{y}}\in \mathds{R}^N$) of a constrained function, each exceeding component $i\in\{1,\dots,N\}$ of $\bm{y}$ is reflected into the box according to 
      \begin{equation}
	y_i = \left\{ \begin{matrix}
			 \check{y}_i + \left( (\check{y}_i-y_i) - \left\lfloor\frac{\check{y}_i-y_i}{\hat{y}_i-\check{y}_i}\right\rfloor (\hat{y}_i-\check{y}_i)\right), & \textrm{if } y_i < \check{y}_i, \\[1ex]
			 \hat{y}_i - \left( (y_i-\hat{y}_i) - \left\lfloor\frac{y_i-\hat{y}_i}{\hat{y}_i-\check{y}_i}\right\rfloor (\hat{y}_i-\check{y}_i) \right), & \textrm{if } y_i > \hat{y}_i,\\
	                 y_i , & \textrm{else}.
	                \end{matrix} \right.
	                \label{reflect} 
      \end{equation}
  This step is performed before each evaluation of the constrained function.
  
  As suggested by~\cite{TakahamaS10}, in generations $g$ that are multiples of the dimensionality $N$, an additional repair step is performed with probability $\theta_p$ (lines 16 to 22).  
    In this scenario, infeasible offspring candidate solutions $\bm{y}_l^{(g)}$ are repaired by the application of a step that involves the approximation of the Jacobian. The repair step requires $N$ function evaluations per execution plus one single evaluation of the repaired candidate solution.
    While this step is rather costly, it can potentially guide the search towards the feasible region of the search space. Since $\texttt{GradientBasedRepair}(.)$ does not guarantee feasibility in a single step, it is repeated up to $\theta_r=3$ times~\cite{HellwigB2018}.
    The details of this repair technique are provided in Sec.~\ref{repair}. Within Alg.~\ref{alg_MAES} the lines related to the 
    {\color{black} Jacobian-based} mutation step are highlighted by the blue box.
    
  Offspring candidate solutions $\bm{y}_l^{(g)}$ that are adjusted by application of the repair algorithm components $\texttt{KeepRange}(.)$, or $\texttt{GradientBasedRepair}(.)$, are expected to differ from the 
  originally sampled $\bm{\bar{y}}$ in at least one component. The corresponding mutation vector $\bm{d}_l^{(g)}$ and $\bm{z}_l^{(g)}$ of $\bm{y}_l^{(g)}$ have to be readjusted in order to adequately take into account the correct quantities in the update of the transformation matrix $\bm{M}$. This readjustment referred to as \emph{back-calculation} is executed in lines 24 and 25 of Alg.~\ref{alg_MAES}. While the repair of $\bm{d}_l^{(g)}$ is straight forward, that of $\bm{z}_l^{(g)}$ involves the inverse of the transformation matrix $\bm{M}^{(g)}$. Since $\bm{M}^{(g)}$ has no fixed properties, it can easily become singular. Hence, the pseudo inverse ({\color{black}or Moore-Penrose inverse}) $\bm{M}^{-1}$ of $\bm{M}^{(g)}$ is used in line 25. It is computed in line 9 at the beginning of each generation.
  If the transformation matrix update results in a matrix $\bm{M}^{(g)}$ that is ill-suited for the determination of the pseudo inverse $\bm{M}^{-1}$, a reset step has to be applied in line 10 to prevent the algorithm from breaking off due to numerical instabilities. As a first workaround, we simply reset the transformation matrix to $\bm{I}$. This way the transformation matrix adaptation is restarted at the current location in the search space. All steps corresponding to the back-calculation are emphasized in Alg.~\ref{alg_MAES} by use of green boxes. 
  
  After the computation of all $\lambda$ offspring, the best offspring is compared to the best solution found so far $\bm{y}_\textrm{bsf}$ in line 28.
  The parental recombinant $\bm{y}^{(g)}$ is updated in line 30. Its update involves the selection of the best $\mu$ mutation vectors with respect to $\leq_{\epsilon}$. 
  In line 31, the $\epsilon$MAg-ES adapts the search path $\bm{p}_\sigma^{(g)}$ known from CMA-ES~\cite{Hansen2003}. Its length indicates whether the mutation strength $\sigma^{(g)}$ should be decreased or increased in the next generation. Further, it contributes to the transformation matrix update which is performed in line 32. The corresponding strategy parameters are chosen according to the recommendations in~\cite{BeyerS2017}.  
  $\sigma^{(g)}$ is then updated in line 33. It is bounded from above by the parameter $\sigma_{\max}$ (indicated by the red box in Alg.~\ref{alg_MAES}). 
  The use of $\sigma_{\max}$ was simply motivated by empirical observations where the $\epsilon$MAg-ES began to gradually increase the mutation strength towards infinity on some constrained test functions and requires further examinations.
  
  Finally, the $\epsilon^{(g)}$-threshold is gradually decreased in lines 35 to 36 until it reaches zero, or a predefined number of generations $T$ is reached, respectively. For $g > T$, $\epsilon^{(g)}$ is directly set to zero. This procedure continuously increases the accuracy with which the $\leq_\epsilon$ order relation distinguishes feasible from infeasible candidate solutions (see Eq.~\eqref{epso} below).
  
  These steps are repeated until $fevals$ exceeds the budget of function evaluations $fevals_{\max}$. 
  After termination, the algorithm returns the best-found solution $\bm{y}_\textrm{bsf}$ together with its objective function value $f(\bm{y}_\textrm{bsf})$, and the corresponding constraint violation $\nu(\bm{y}_\textrm{bsf})$, respectively.

  In the following section, the original $\epsilon$MAg-ES~\cite{HellwigB2018} is compared to six algorithmically reduced versions of itself. That way, one can expect to gain insights with respect to the contribution of the individual algorithm components.  Ideally, one aims to identify those problem characteristics for which a specific algorithm component works best.
  The six reduced $\epsilon$MAg-ES versions are obtained by abandoning (combinations of) the colored algorithm parts:
  \begin{enumerate}
   \item \textbf{$\bm{\epsilon}$MA-ES} -- A variant that omits the {\color{black} Jacobian-based} repair step, see Sec.~\ref{repair}. That is, all {\color{black}blue diagonally hatched} parts of Alg.~\ref{alg_MAES} are ignored, i.e. lines 16 to 22.
   \item \textbf{$\bm{\epsilon}$MAg-ES w/o} -- This modification of the $\epsilon$MAg-ES goes without the back-calculation step. Accordingly, the green {\color{black}counterdiagonally hatched} algorithm parts must be omitted (lines 9,10, and 23 --26).
   \item \textbf{$\bm{\epsilon}$MAg-ES nl} -- The original $\epsilon$MAg-ES without considering a bound on the maximal mutation strength of Alg.~\ref{alg_MAES}. Hence, the red box {\color{black}in chequer pattern} in line 33 of the pseudo-code is omitted and no limitation (nl) of $\sigma$ is assumed.
   \item \textbf{$\bm{\epsilon}$SAg-ES} -- This variant of the $\epsilon$MAg-ES neglects the update of the transformation matrix $\bm{M}$, i.e. in each generation the offspring candidate solutions are sampled from an isotropic normal distribution with standard deviation $\sigma$. Hence, the gray {\color{black}dotted} part in line 32 of Alg.~\ref{alg_MAES} is neglected.
   \item \textbf{lexMAg-ES} -- This version replaces the $\epsilon$-level ordering~\eqref{epso} with the lexicographic ordering approach~\eqref{lexo}. This can easily be obtained by setting $T$ to zero in line 2 of Alg.~\ref{alg_MAES}. {\color{black}Consequently, all lines in the pseudo-code that are highlighted by yellow horizontal lines can be ignored (i.e. lines 3,4, and 35 to 37).}
   \item \textbf{lexMA-ES} -- A modification of the lexMAg-ES that comes without the {\color{black} Jacobian-based} repair step (Sec.~\ref{repair}). {\color{black}That is, the lines 3,4, 16 to 22, and 35 to 37 in Alg.~\ref{alg_MAES} must be omitted.}
  \end{enumerate}
    
    {\color{black}
   The fixed strategy parameter settings used for the comparison of the $\epsilon$MAg-ES variants are outlined in Table~\ref{config}. 
   Most of these parameters are based on standard configuration of the MA-ES~\cite{LoshchilovGlasmachersBeyer2018} or the original recommendations of the respective constraint handling methods~\cite{TakahamaS10}. Others, like $\sigma_\textrm{max}$ represent the best guess settings used in the constrained CEC2018 competition, cf.~\cite{HellwigB2018}. Current experiments have shown that the number of parameters and their settings can be simplified. But changing the algorithm and tuning its strategy parameter is out of the scope of this paper.
   
   \begin{table}[htbp]\color{black}
    \caption{Strategy parameter configuration used for the subsequent investigations.}
    \resizebox{0.95\textwidth}{!}{%
    \begin{tabular}{p{4.2cm}l||p{4.2cm}l}
     \textbf{Description} & \textbf{Value} & \textbf{Description} & \textbf{Value} \\\hline
      Search space dimensions  & $N=10$ and $N=100$                 & Budget & $fevals_{\textrm{max}} = 20000 \cdot N$ \\
      Offspring population size & $\lambda = 4N$                     &  Initial mutation strength & $\sigma_{\textrm{Init}}=1$\\
      Parental population size & $\mu = \lfloor \lambda/3 \rfloor$  & Maximum mutation strength & $\sigma_{\textrm{max}} = 100$ \\ 
      Recombination weigths   & $w_i =  \frac{\ln(\mu+\frac{1}{2})-\ln i}{\sum_{j=1}^\mu (\ln(\mu+\frac{1}{2})-\ln j)}, \: i =1,\dots,\mu$   & $\epsilon$-level control parameter & $\theta_t = 0.9$ \\
      Effective population size & $\mu_w = \left( \sum_{i=1}^\mu w_i^2 \right)^{-1}$        & $\epsilon$-level control parameter& $T= 1000$\\
      Rank-1 learning rate  &  $c_1 = \frac{2}{(N+1.3)^2+\mu_w}$                             & $\epsilon$-level control parameter & $\gamma_{\textrm{min}} = 3$\\
      Rank-$\mu$ learning rate  &  $c_\mu = \min\left(1-c_1,\frac{2(\mu_w-2+1/\mu_w)}{(N+2)^2+\mu_w} \right) $                       & Repair repetitions & $\theta_r = 3$   \\
      $\sigma$ learning rate    & $c_\sigma = \frac{\mu_w+2}{N+\mu_w+5} $                    &  Repair probability & $\theta_p = 0.2$
    \end{tabular}
    }
    \label{config}
   \end{table}
    }
       
      \subsection{The \texorpdfstring{$\epsilon$}{epsilon}-level ordering} 
      \label{epslevel}
      The population is driven towards the optimum of~\eqref{cop} by variation and selection. Regarding selection, the generated offspring individuals of a single generation have to be ranked. Feasible solutions are considered superior to infeasible solutions. The usual lexicographic ordering primarily ranks two candidate solutions according to their constraint violations and secondly with respect to their objective function values.
      The $\epsilon$MAg-ES uses another ordering relation: the $\epsilon$-level order relation introduced by~\cite{TakahamaS06} in the context of DE.
      The $\epsilon$-level ordering represents a relaxation that enables the algorithm to treat infeasible candidate solutions with constraint violation below a specific $\epsilon^{(g)}$ threshold as feasible. A candidate solution $\bm{y}$ is said to be $\epsilon$-feasible, if its constraint violation $\nu(\bm{y})$ does not exceed a predefined constraint violation threshold $ \epsilon^{(g)}$ in generation $g$. The threshold $\epsilon^{(g)}$ is continuously reduced to zero with the number of generations. Hence, the strategy can move outside the feasible region within the early phase of the search process which can potentially support the convergence to the optimizer $\bm{y}^*$. 
      
      {
      Given two candidate solutions $\bm{y}_i\in\mathds{R}^N$ and $ \bm{y}_j\in\mathds{R}^N$, the pairs $\left(f_i, \nu_i\right) \coloneqq\left(f(\bm{y}_i), \nu(\bm{y}_i)\right) $, and $(f_j,\nu_j)$ respectively, represent the corresponding objective function values as well as the related constraint violations. The $\epsilon$-level order relation denoted by $\leq_\epsilon$ is then defined as
      }
	    \begin{equation}
	    \begin{split}
	     \bm{y}_i  \leq_\epsilon \bm{y}_j   
		      \Leftrightarrow  \left\{ \begin{matrix}
                               f_i \leq f_j, & \textrm{if }& (\nu_i \leq \epsilon^{(g)}) \wedge (\nu_j \leq \epsilon^{(g)}),\\
			       f_i \leq f_j, & \textrm{if }& \nu_i =\nu_j, \\
                               \nu_i < \nu_j, & & \textrm{otherwise. }
                         \end{matrix}  \right.
              \end{split}
                         \label{epso}
	    \end{equation}
	    Candidate solutions are compared according to the following criteria:
	    Two $\epsilon$-feasible solutions are ranked with respect to their objective function values.
	    Two $\epsilon$-infeasible solutions are ordered based on their constraint violations. 
		    Ties are resolved by considering the objective function values.
		    
	    The initial $\epsilon^{(0)}$ is determined as the average constraint violation of $\theta_t$ (percent) of the best candidate solution within the initial population (see Alg.~\ref{alg_MAES}, line 3). During the search process, it is gradually reduced with each generation until it is set to zero after a fixed number of generations $T$. For $\epsilon=0$, the $\epsilon$ level ordering is equal to the lexicographic ordering mentioned above.
	   
      \subsection{{\color{black} Jacobian-based} repair} 
      \label{repair}
      In addition to $\epsilon$-level order relation, the $\epsilon$MAg-ES makes use of the {\color{black} Jacobian-based} repair approach introduced in~\cite{TakahamaS06,TakahamaS10}. 
      Let the vector of all $m$ constraint values according to problem~\eqref{cop} be denoted by
      \begin{equation}
          \bm{C}(\bm{y})=\left(g_1(\bm{y}),\ldots,g_l(\bm{y}),h_1(\bm{y}),\ldots,h_k(\bm{y})\right)^\top.
      \end{equation}			
      By determining the degree of violation of an infeasible candidate solution $\bm{y}$, the vector of constraint violations reads
      \begin{equation}
      \bar{\bm{C}}(\bm{y})= 
			\big(\max(0,g_1(\bm{y})),\ldots,\max(0,g_l(\bm{y})),h_1(\bm{y}),\ldots,h_k(\bm{y})\big)^{\!\top}.
      \end{equation}
       The infeasible candidate solution $\bm{y}$ can then be repaired by addition of a correction vector $\bar{\bm{y}}$
		\begin{equation}
		  \bm{\tilde{y}} = \bm{y} + \bar{\bm{y}}.
		\end{equation}
		The term $\bar{\bm{y}}$ is calculated by solving the linear system
		\begin{equation}
		  \mathcal{J}_{\bm{C}}(\bm{y}) \cdot \bar{\bm{y}} = - \bar{\bm{C}}(\bm{y}).
		  \label{eq018}
		\end{equation}
		To this end, the Jacobian matrix $\mathcal{J}_{\bm{C}}(\bm{y})$ with respect to $\bm{C}(\bm{y})$ has to be determined. Since constrained black-box optimization is considered, the Jacobian needs to be approximated, e.g. by using finite differences.
		Equation~\eqref{eq018} can approximately be solved by making use of the pseudo inverse $\mathcal{J}_{\bm{C}}(\bm{y})^{-1}$
		\begin{equation}
		   \bar{\bm{y}} = - \mathcal{J}_{\bm{C}}(\bm{y})^{-1} \bar{\bm{C}}(\bm{y}).
		  \label{eq019}
		\end{equation}
	In the case that no feasible solution is found after the correction, the {\color{black} Jacobian-based} repair step is repeated at most $\theta_r$ times. 
	Usually, the constraint violation is gradually decreased with every repair step. 
	If the strategy cannot find a feasible solution,
	the last infeasible candidate solution is considered as the new offspring candidate solution.

\section{Experimental algorithm assessment}
\label{sec04}

This section is concerned with empirical investigations of the core components of the $\epsilon$MAg-ES.
   To this end, the individual impact of the single algorithm components on the overall $\epsilon$MAg-ES performance is validated on the constrained CEC2017 benchmarks. The six different algorithm variants introduced in Sec.~\ref{sec03} are compared against the original $\epsilon$MAg-ES which is considered the baseline algorithm. The aim is to discover those problem properties for which a certain algorithm component is particularly well suited. 
   
   As a first step, the contribution of the individual components is assessed concerning the performance indicators recommended by {\color{black}the constrained CEC2017 benchmarks}. Therefore, all algorithm variants performed 25 independent runs on all 28 constrained CEC2017 test problems. The reduced algorithm variants are then separately compared to the $\epsilon$MAg-ES in dimensions $N=10$ and $N=100$.
    
%
    The CEC2017 benchmark performance results of the original $\epsilon$MAg-ES algorithm~\cite{HellwigB2018} for dimension $N=10$ and $N=100$ are provided in Sec.~\ref{AppendixB} of the supplementary material, see Tables~\ref{CEC18_MAESD10} and~\ref{CEC18_MAESD100}. 
    {Due to space limitations, the tables for the other algorithm variants are not provided. Yet, the corresponding performance     data are included in the supplementary material.}
    
    To provide a full picture of the algorithm performance, the distinct $\epsilon$MAg-ES variants are benchmarked on the complete set of CEC2017 test functions. On each particular constrained problem, they are separately compared to the baseline algorithm ($\epsilon$MAg-ES) by making use of the performance indicators recommended in~\cite{CEC2017}. That is, for each pair of algorithms one takes into account the \emph{median} as well as the \emph{mean} fitness and constraint violation values over the 25 independent runs. 
    
    The CEC2017 benchmark approach for ranking algorithms on a single problem is dyadic. A first ranking is based on the median solution returned by the algorithms. As all final candidate solutions come with its fitness value as well as its constraint violation value, the lexicographic ordering is used to rank the 25 algorithm realizations.
    Two distinct algorithm variants are then compared in terms of their median candidate solutions by the following rules:
    \begin{itemize}
     \item[(i)] feasible solutions are better than infeasible solutions,
     \item[(ii)] infeasible solutions are ordered according to their constraint violation amounts, and 
     \item[(iii)] feasible solutions are ranked based on their objective function values.
    \end{itemize}
    \begin{figure}[b]
     \includegraphics[width=0.98\textwidth]{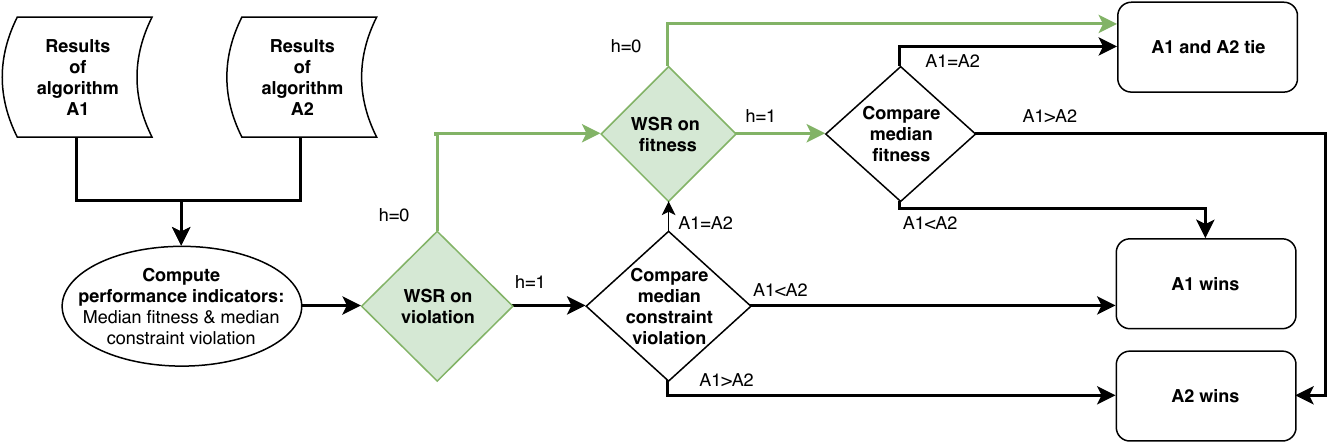}
     \Description{Median ranking approach with respect to the technical report ofthe constrained CEC2017 benchmarks.}
      \caption{Median ranking approach according to the technical report of {\color{black}the constrained CEC2017 benchmarks~\cite{CEC2017}}. The green parts display the newly integrated aspect of statistical testing into the ranking flow. To this end, the Wilcoxon Signed Rank (WSR) test has been used: a value of $h=1$ is indicating that both algorithm realizations are significantly different at a 5\% significance level and $h=0$ is indicating the opposite, respectively. }
     \label{MedianRanking}
    \end{figure}
    For a graphical representation of the ranking procedure with respect to the median performance indicators recommended for {\color{black}the constrained CEC2017 benchmarks}, refer to Fig.~\ref{MedianRanking}. 
    {
    Once the median solution among all 25 realizations of algorithm A1 has been identified, its fitness and constraint violation values are compared to those of algorithm A2. In the first step, the constraint violation values are considered and the algorithm with lower constraint violation is elected as the winner strategy. In case of zero or equal median constraint violations the fitness values are taken into account. Again, due to minimization, the algorithm with lower fitness value is chosen as the winner strategy of the one-on-one comparison. Equal constraint violation and equal fitness values of two algorithms result in a tie. 
    Note that the original CEC2017 ranking comes without the green diamonds that represent decisions based on significance testing, i.e. the 
    test results must be ignored in that respective context and $h=1$ can always be assumed.
    }

    The ranking that is based on the mean performance indicators additionally includes the feasibility rate $FR$ of the respective algorithm variants. It works as follows:
    \begin{itemize}
     \item[(i)]  rank the algorithms based on their feasibility rates $FR$,
     \item[(ii)]  algorithms with equal feasibility rates $FR$ are ordered according to the mean constraint violation amounts, and 
     \item[(iii)]  ties with respect to the mean constraint violation values are broken by comparison of the mean objective function values.
    \end{itemize}
    Note that the feasibility rate $FR$ is defined as the ratio of the number of algorithm runs resulting in a feasible candidate solution over the number of all 25 algorithm runs. For brevity, a graphical representation in the style of Fig.~\ref{MedianRanking} is omitted for the mean ranking. Except for the consideration of the feasibility rate, and the use of the mean performance indicators, it would look similar.
    
    Yet, looking at the tabular algorithm results, it appears that the median and mean performance indicators of many strategy variants do only differ slightly from the baseline results. To avoid preference based on insignificant differences, this paper introduces the idea to additionally including a statistical test into the median ranking procedure. To this end, the ranking procedure is adjusted by performing a \emph{Wilcoxon Signed-Rank test} before ranking decisions that involve the median fitness or the median constraint violation realizations. In Fig.~\ref{MedianRanking}, this additional step is illustrated by the green diamonds.

    The Wilcoxon signed rank test represents a non-parametric test for two paired algorithm realizations. The test statistic is the sum of the ranks of positive differences between the observations, {
    By rejecting the null hypothesis at a 5\% significance level, the algorithm realizations can be regarded as sufficiently dissimilar to use them within the ranking process.
    It has to be noted that the \emph{Wilcoxon Signed-Rank test} is not able to make a statement on the mean values or the variance of the algorithm realization. That is, due to single outliers, the mean values might turn out to be reasonably different even if the \emph{Wilcoxon Signed-Rank test} would accept the null hypothesis. 
    Yet, aiming at a consistent ranking methodology that integrates signifiance testing into the {\color{black}ranking recommendations of the constrained CEC 2017 benchmarks}, we propose to use the non-parameterized \emph{Wilcoxon Signed-Rank test} also for the mean ranking. Since one cannot make any assumptions concerning the distribution, making use of the Signed-Rank test allows to provide a notion of statistical significance concerning differences in the distributions of the algorithm realizations. 

    The present paper uses the updated median and mean performance ranking for pairwise comparison with the baseline algorithm only. 
    The detailed comparison results corresponding to Table~\ref{tab1} are included in Sec.~\ref{AppendixC} of the supplementary material, while Table~\ref{tab1} presents the aggregated results for dimensions $N=10$ and $N=100$.
    For each algorithm variant, the tables indicate whether neglecting a specific component does affect the algorithm performance or not.         
    The outcome is displayed by use of the notation $\bm{+}$/$\bm{=}$/$\bm{-}$, with
    \begin{itemize}\itemsep3pt
     \item[$\bm{+}$] referring to the number of test problems on which $\epsilon$MAg-ES exhibits \textbf{superior performance} with respect to the considered indicator,
     \item[$\bm{=}$] indicating the number of benchmarks for which \textbf{a tie} is observed based on either insignificant differences within the algorithm realization or equal performance indicators, and
     \item[$\bm{-}$]  denoting the number of test problems on which $\epsilon$MAg-ES shows \textbf{inferior performance}.
    \end{itemize}
    Accordingly, the $\bm{+}$ symbol corresponds to a positive contribution of the neglected component within the $\epsilon$MAg-ES, and the $\bm{-}$ symbol to a negative contribution, respectively.
    The $\bm{=}$ symbol implies that the component has no significant effect on the algorithm effectiveness on the given number of problems.
    
    For {\color{black}the constrained CEC2017 benchmarks}, the total ranks are built by aggregating both the median as well as the mean ranks. By doing this, in a few cases, advantages in terms of median performance and disadvantages with respect to mean performance may cancel out, and vice versa. 
    In this situation, both algorithms are considered to show equal performance following the CEC2017 benchmark ranking. For an illustration of the rank aggregation refer to Table~\ref{tabA}.
    
    \newcolumntype{g}{>{\columncolor{gray!50}}c}   
\begin{table}[b]
\centering
\caption{Results of the significance test based comparison of $\epsilon$MAg-ES with six different reduced algorithm variants of itself on the constrained CEC2017 benchmark functions. For a comparison with the ranking approach recommended for CEC2018 refer to Sec.~\ref{AppendixA} of the supplementary material.}
\label{tab1}
\renewcommand{\arraystretch}{1.1}
\resizebox{0.6\textwidth}{!}{%
\begin{tabular}{lccgccg}
  $\bm{\epsilon}$\textbf{MAg-ES}          & \multicolumn{3}{c}{$\bm{N=10}$} & \multicolumn{3}{c}{$\bm{N=100}$} \\
            & \multicolumn{3}{c}{Ranking} & \multicolumn{3}{c}{Ranking} \\
 $\bm{+}$/$\bm{=}$/$\bm{-}$      & Median & Mean & Total & Median & Mean   & Total\\\hline
$\epsilon$MA-ES       & 7/19/2    & 7/17/4     & 7/18/3           & 5/12/11  & 7/11/10 & 5/13/10\\
$\epsilon$MAg-ES w/o  & 8/17/2    & 10/15/3     & 10/15/3          & 10/8/10  & 13/6/9 & 11/8/9\\
$\epsilon$MAg-ES nl   & 12/14/2   & 14/12/2    & 14/12/2         & 5/15/8   & 7/14/7 & 5/16/7\\
$\epsilon$SAg-ES      & 18/9/1    & 20/7/1     & 20/7/1          & 18/8/2   & 18/8/2 & 17/10/1\\
$lex$MAg-ES      & 6/20/2    & 7/18/3          & 7/18/3          & 9/10/9   & 10/9/9 & 9/11/8\\
$lex$MA-ES       & 8/16/4    & 8/14/6          & 8/14/6          & 14/7/7   & 14/7/7 & 13/9/6\\\hline
\end{tabular}%
}
\end{table}

        In small dimensions ($N=10$), the results in Table~\ref{tab1} indicates that the original $\epsilon$MAg-ES equipped with all algorithm components demonstrates the best comparison results over the whole benchmark set. The neglection of either component results in inferior performance with respect to both recommended indicators. The only exception is the mean performance comparison with the $lex$MA-ES which is basically indistinguishable. Yet, considering the aggregated median and mean ranks, a tendency towards $\epsilon$MAg-ES can be observed. Anyway, the use of the lexicographic ordering appears to be sufficient on many test problems in dimension $N=10$.      
    
    \begin{table}[t]
     \centering
\caption{Total aggregated ranking according to the CEC2017 recommendations and augmented by the statistical significance testing, see Fig.~\ref{MedianRanking}.}
\label{tabA}
\renewcommand{\arraystretch}{1.1}
\resizebox{0.25\textwidth}{!}{%
\begin{tabular}{rl|ccc}
 \multicolumn{2}{c|}{\tiny Total} & \multicolumn{3}{c}{\tiny Mean Rank} \\[-1ex]
  \multicolumn{2}{c|}{\tiny Rank}  & $\bm{+}$ & $\bm{=}$ & $\bm{-}$ \\ \hline
\multirow{3}{*}{\rotatebox[origin=c]{90}{\tiny Median Rank}}  & $\bm{+}$ & $+$ &$+$ & $=$\\
    & $\bm{=}$ & $+$ & $=$ & $-$ \\
   & $\bm{-}$ & $=$ & $-$ & $-$
\end{tabular}%
}
    \end{table}
    
    The image changes somehow when taking into account high dimensions ($N=100$). There, the $\epsilon$MAg-ES comparison with $\epsilon$MA-ES points towards a negative influence of {\color{black} Jacobian-based} repair on the overall algorithm ranking.  
    Further, the positive effect of the \emph{back-calculation} appears to be decreasing with growing dimensionalities. 
    Regarding the limitation of the mutation stength $\sigma$, it seems to make no difference whether this specific algorithmic component is included or not. The clear positive contribution observed in $N=10$ is lost. The same applies to the choice of the order relation. Yet, in case that the lexicographic order relation is used, the need of {\color{black} Jacobian-based} repair is supported by the relative performance of the $lex$MA-ES.
    
    In both considered dimensionalities, the bad results of the $\epsilon$SAg-ES support the effectiveness of the matrix adaptation component within the $\epsilon$MAg-ES. Accordingly, this well-known result from the context of unconstrained optimization is also recognized on the constrained problems of the CEC2017 benchmark suite. 
        
    However, the performance assessment of the CEC2017 test suite concentrates only on the effectiveness of competing algorithms.
    An performance measure with respect to algorithm efficiency is missing~\cite{HellwigB2019a}. 
    We propose to introduce an additional measure of algorithm running time to deal with the effectiveness ties. 
    The running time of probabilistic search algorithms can directly be identified with the number of function evaluations that is consumed until the algorithm reaches a certain target value.
    As the globally optimal solutions of the constrained test problems are generally not known, it is not immediately obvious how such a target value could be defined. 
    
    As a first approach, we suggest to measure the accumulated number of function evaluations needed to find the best-so-far candidate solution that is finally returned in each independent algorithm run. This way, the mean effectiveness indicators (mean objective function and mean constraint violation) are complemented by a mean efficiency indicator: the mean running time to approach the algorithm's globally best observation \emph{meanRTgb}. This indicator cannot be regarded as a stand-alone performance measure, but it can be used to break ties with respect to algorithm realizations that are only insignificantly different or equal in terms of their median and mean effectiveness indicators.
    An illustration of this comparison method is provided in Table~\ref{tab2}. There, the $\bm{+}$ (and $\bm{-}$) sign again indicate those constrained benchmark problems on which the $\epsilon$MAg-ES exhibits superior (and inferior) performance in terms of effectiveness. For all ties, the mean running time \emph{meanRTgb} to reach the best candidate solution observed is considered.
    The \emph{meanRTgb} of an algorithm variant is then displayed relative to that of the baseline algorithm. Example given, when taking into account test problem C01, the effectiveness of $\epsilon$MAg-ES and $\epsilon$MA-ES is only insignificantly different. Yet, $\epsilon$MA-ES only uses 98\% of the function evaluations consumed by $\epsilon$MAg-ES to approach candidate solutions of similar fitness. While this difference in the running time may be of no consequence, other problems reveal greater deviations, see $C17$ or $C19$.
    
\begin{table}[t]
\centering
\caption{Total ranking comparison results according to the constrained CEC2017 benchmark functions.}
\label{tab2}
\resizebox{0.98\textwidth}{!}{%
\begin{tabular}{l|cccccc|cccccc}
\multicolumn{1}{l}{} & \multicolumn{6}{c}{$\bm{N=10}$} & \multicolumn{6}{c}{$\bm{N=100}$} \\[0ex]
\multicolumn{1}{l}{}               & \multicolumn{6}{c}{$\epsilon$MAg-ES vs.}              & \multicolumn{6}{c}{$\epsilon$MAg-ES vs.}            \\[1ex]
\multirow{ 2}{*}{Problem}        & \multirow{ 2}{*}{$\epsilon$MA}     &  $\epsilon$MAg  & $\epsilon$MAg    & \multirow{ 2}{*}{$\epsilon$SAg}  & \multirow{ 2}{*}{lexMAg}   & \multirow{ 2}{*}{lexMA}    & \multirow{ 2}{*}{$\epsilon$MA}    & $\epsilon$MAg  & $\epsilon$MAg & \multirow{ 2}{*}{$\epsilon$SAg}  & \multirow{ 2}{*}{lexMAg }  & \multirow{ 2}{*}{lexMA }   \\[-1ex] 
           & &  w/o & nl & & & & & w/o & nl & & & \\ \hline 
C01 & 0.98 & 0.96 & 1.01 & + & 0.94 & 0.93 & 1.05 & + & 1.05 & + & 1.12 & 1.01 \\
C02 & 0.94 & 1.09 & 0.96 & + & 0.94 & 1.44 & 1.06 & + & 0.97 & + & 1.19 & 0.97 \\
C03 & 0.95 & 1.05 & 1.06 & + & + & + & 0.86 & + & 1.01 & + & + & + \\
C04 & - & - & 0.51 & 0.85 & + & + & - & 1.35 & 1.10 & 0.97 & + & + \\
C05 & 0.98 & 1.01 & 0.97 & + & 1.00 & 1.09 & - & + & - & + & - & 1.00 \\
C06 & + & + & + & + & + & + & + & + & 1.04 & + & + & + \\
C07 & + & + & + & + & + & + & 1.02 & 1.24 & 1.37 & 0.60 & + & + \\
C08 & 0.66 & 0.99 & 1.00 & + & 0.25 & 0.22 & - & 1.00 & 1.00 & + & - & - \\
C09 & 0.67 & + & + & + & 0.35 & 0.29 & - & - & + & 1.06 & - & - \\
C10 & 0.69 & + & 1.00 & + & 0.27 & 0.26 & - & - & - & + & - & - \\
C11 & + & + & + & + & 0.59 & 0.32 & - & + & - & + & - & - \\
C12 & 0.83 & 1.16 & + & 0.79 & 0.79 & 0.67 & 0.71 & 0.86 & 0.86 & + & 0.63 & 0.58 \\
C13 & 0.74 & 0.90 & 0.84 & + & 0.74 & 0.66 & - & 0.99 & - & + & + & + \\
C14 & + & + & + & + & - & - & 0.40 & - & + & 0.55 & 0.13 & + \\
C15 & + & + & + & + & - & - & + & + & + & + & 0.64 & 0.75 \\
C16 & 0.78 & 0.97 & 0.98 & 1.08 & + & + & 0.88 & + & 0.95 & + & + & + \\
C17 & 1.46 & 0.68 & 0.73 & 0.24 & 0.79 & 1.56 & 0.71 & - & - & 0.72 & 0.57 & + \\
C18 & 0.84 & + & + & + & 0.80 & 0.65 & 0.76 & 1.00 & 1.00 & + & 0.87 & 0.61 \\
C19 & 0.55 & 1.39 & + & + & 0.81 & 0.67 & 1.00 & - & 1.00 & 0.85 & 1.00 & 1.00 \\
C20 & - & - & - & - & 1.05 & - & - & - & - & - & - & - \\
C21 & 0.99 & 1.09 & 0.95 & 0.79 & 0.66 & - & 0.80 & + & 0.98 & + & 0.73 & 0.67 \\
C22 & 0.82 & 0.87 & - & + & 0.97 & 0.80 & - & - & 0.97 & 1.01 & - & - \\
C23 & - & - & + & + & - & - & + & 0.47 & + & 0.71 & 0.24 & + \\
C24 & + & + & + & + & + & + & + & + & + & + & + & + \\
C25 & 0.76 & 0.96 & 0.95 & 1.17 & + & + & 0.82 & + & 0.92 & + & + & + \\
C26 & 1.72 & 1.30 & + & 0.59 & 1.10 & 2.28 & + & - & - & 0.63 & - & + \\
C27 & + & + & + & + & 0.84 & + & - & 1.00 & 1.00 & + & + & + \\
C28 & 0.86 & 1.18 & + & + & 0.99 & - & 0.92 & - & 0.99 & 0.76 & 1.00 & 0.92
\end{tabular}%
}
\end{table}

  Unfortunately, the CEC2017 benchmark suite does not provide a clear grouping of problems according to common characteristics~\cite{HellwigB2019a}. Hence, the interpretation of the detailed results in Table~\ref{tab2} is rather complicated. 
  In order to provide some insights, we will focus on the differences with respect to separability and rotation of the problems.
  Additionally, we take a look on those constrained problems that have only equality constraints and we distinguish between those problems that include at least one equality constraint and those that do only consider inequality constraints.
  
  \paragraph{Fully separable problems}
  Regarding {\color{black}the constrained CEC2017 benchmarks}, only four constrained problems are fully separable, i.e. these problems come with a separable objective functions as well with separable constraint functions. These problems are $C04$, $C06$, $C07$, $C12$. Interestingly, the original $\epsilon$MAg-ES obtained relatively poor results on $C04$ and $C12$ when compared to the Differential Evolution variants during the CEC2018 competition. This advantage of Differential Evolution algorithms might be attributed to the crossover operator which is known to promote a search in  direction of the coordinate axes~\cite{SuttonLW2007}.
  In both problems, the objective is given by Rastrigin's function equipped with two non-linear constraints.
  
  On the other hand, the $\epsilon$MAg-ES was performing good on the other two problems ($C06$ and $C07$) which have only equality constraints. Here, the {\color{black} Jacobian-based} repair method might be considered to have a beneficial contribution to the algorithm performance. However, having a closer look on those problems with only equality constraints reveals a more complex situation.

    \begin{table}[tp]
\centering
\caption{Comparison of $\epsilon$MAg-ES with the six algorithm variants on those seventeen constrained CEC2017 benchmark functions that include at least one equality constraint.}
\label{tab1eq}
\renewcommand{\arraystretch}{1.1}
\resizebox{0.6\textwidth}{!}{%
\begin{tabular}{lccgccg}
  $\bm{\epsilon}$\textbf{MAg-ES}          & \multicolumn{3}{c}{$\bm{N=10}$} & \multicolumn{3}{c}{$\bm{N=100}$} \\
            & \multicolumn{3}{c}{Ranking} & \multicolumn{3}{c}{Ranking} \\
 $\bm{+}$/$\bm{=}$/$\bm{-}$      & Median & Mean & Total & Median & Mean   & Total\\\hline
$\epsilon$MA-ES       & 7/10/0    & 7/8/2     & 7/9/1           & 5/7/5  & 6/6/5 & 5/7/5\\
$\epsilon$MAg-ES w/o  & 9/8/0    & 10/6/1     & 10/6/1          & 6/7/4  & 7/5/5 & 7/5/5\\
$\epsilon$MAg-ES nl   & 9/8/0   & 11/6/0    & 11/6/0        & 5/9/3   & 5/8/4 & 5/8/4\\
$\epsilon$SAg-ES      & 11/6/0    & 13/4/0     & 13/4/0          & 12/5/0   & 11/5/1 & 11/6/0\\
$lex$MAg-ES      & 5/10/2    & 6/8/3          & 6/8/3          & 7/5/5   & 7/4/6 & 7/5/5\\
$lex$MA-ES       & 7/8/2    & 7/7/3          & 7/7/3          & 12/1/4   & 11/1/5 & 11/2/4\\\hline
\end{tabular}%
}
\end{table}
  
  \paragraph{Equality constraints}
  
  Table~\ref{tab1eq} is considering all those seventeen constrained problems that include at least one equality constraint.
  
  We observe that the $\epsilon$MAg-ES is exhibiting superior performance with respect to all other algorithm variants considered in dimension $N=10$. The back-calculation approach, the limitation of the mutation strength and the transformation matrix adaptation do realize the biggest advantages. On the other hand, the performance is quite balanced regardless of using the $\epsilon$-level or the lexicographic ordering relation. 
  
  The situation changes with growing dimensionality: The performance advantages are less evident. Unsurprisingly, the transformation matrix adaptation must be considered useful. Yet, all the other algorithm components appear to have their strengths and weaknesses depending on the problem characteristics. Yet, if the lexicographic ordering is used, the application of {\color{black} Jacobian-based} repair turns out to be beneficial on many problems.

  \paragraph{Equality constraints only}
  Only three constrained problems are defined with equality constraints only: $C06$, $C07$, $C08$, and $C10$.
  While Rastrigin's function is accompanied by six non-linear equality constraints in $C06$, the objective of constrained problem $C07$ consists of a sum of sinusoidal terms and comes with two non-linear equality constraints. Problems $C08$ and $C10$ represent a Min-Max problem with distinct sets of two quadratic non-separable constraints each.
  
  Taking into account the low dimensional problems ($N=10$), only $\epsilon$MAg-ES and $lex$MAg-ES are able to realize a feasibility rate of 100\% on problem $C06$. On problem $C07$, $\epsilon$SAg-ES also reaches the feasible region in every run. Among all algorithm variants, $\epsilon$MAg-ES exhibits the best performance on both problems. When considering $N=100$, releasing the mutation strength slightly accelerates the algorithm on $C06$. However, unreasonably large $\sigma$ values can be observed within the mean value dynamics. That is, single runs do exhibit deteriorating mutation strength dynamics.  Apart from $\epsilon$MAg-ESnl , the $\epsilon$MAg-ES again shows the best results.
  For $N=100$, problem $C07$ is solved in equal quality by all variants that use the $\epsilon$-level order relation. Again, the deterioration of the mutation strength is observed in single runs of the $\epsilon$MAg-ESnl. The $lex$MAg-ES and $lex$MA-ES perform significantly worse. 
    
  On problem $C10$, all 7 algorithm variants realize feasible solutions within 100\% of the algorithm runs. Except for $\epsilon$MAg-ES w/o and $\epsilon$SAg-ES, all variants show similarly good performances. Yet, those versions that use the lexicographic order relation improve the mean running time by a factor of $4$ in small dimensions.
  In high dimensions, one observes that all variants approach the feasible region. Except for $\epsilon$SAg-ES, all other algorithm variants outperform the $\epsilon$MAg-ES in terms of efficiency and effectiveness. Those variants that reach the feasible region first, on average, are able to realize significantly lower objective function values. The {\color{black} Jacobian-based} repair operator appears to slow down the progress towards the optimizer. The situation on problem $C08$ is very similar.
  
   \paragraph{Rotated problems}
  The last eight constrained optimization problems of the CEC2017 suite include a rotation of the search space: $C21$ to $C28$. Note that problems $C26$ and $C28$ must be considered to be infeasible. 
  
  Taking into account problem $C21$, the lexicographic algorithm versions show the best performance. The constrained problem consists of Rastrigin's function as objective and two non-linear inequality constraints, see Eq.~\eqref{C21}. Especially in high dimensions, the \emph{back-calculation} appears to have a positive contribution. On the other hand, the versions that use {\color{black} Jacobian-based} repair ($\epsilon$MAg-ES and $lex$MAg-ES) are outperformed by those strategies that omit the repair component ($\epsilon$MA-ES and $lex$MA-ES).

  Due to the immense differences in the problem characteristics, it is not possible to identify any general tendencies with respect to the separate algorithm components, see  Table~\ref{tab1rot}. The only exception being that the matrix adaptation is consistently a beneficial algorithm component.
  
     \begin {table}[tp]
        \centering
        \caption{Comparison of $\epsilon$MAg-ES with the six algorithm variants on the eight constrained CEC2017 benchmark functions that are subject to a search space rotation.}
        \label{tab1rot}
        \renewcommand{\arraystretch}{1.1}
        \resizebox{0.6\textwidth}{!}{%
        \begin{tabular}{lccgccg}
        $\bm{\epsilon}$\textbf{MAg-ES}          & \multicolumn{3}{c}{$\bm{N=10}$} & \multicolumn{3}{c}{$\bm{N=100}$} \\
                    & \multicolumn{3}{c}{Ranking} & \multicolumn{3}{c}{Ranking} \\
        $\bm{+}$/$\bm{=}$/$\bm{-}$      & Median & Mean & Total & Median & Mean   & Total\\\hline
        $\epsilon$MA-ES       & 2/6/0    & 2/5/1     & 2/5/1           & 3/3/2  & 3/3/2 & 3/3/2\\
        $\epsilon$MAg-ES w/o  & 2/6/0     & 2/5/1    & 2/5/1          & 3/2/3  & 3/2/3 & 3/2/3\\
        $\epsilon$MAg-ES nl   & 4/3/1   & 5/2/1    & 5/2/1        & 2/4/2   & 3/4/1 & 2/5/1\\
        $\epsilon$SAg-ES      & 4/4/0    & 5/3/0     & 5/3/0          & 4/3/1   & 5/3/0 & 4/4/0\\
        $lex$MAg-ES           & 2/6/0     & 2/5/1    & 2/5/1           & 2/4/2   & 3/3/2 & 3/3/2\\
        $lex$MA-ES            & 3/4/1    & 3/2/3          & 3/2/3          & 5/2/1   & 5/2/1 & 5/2/1\\\hline
        \end{tabular}%
        }
    \end{table}

\section{Algorithm dynamics on selected problems}
\label{sec05}
 This section presents a selection of test functions that are considered useful for a more detailed investigation of the distinct ES algorithm components. The selection focuses on specific test problems of {\color{black}the constrained CEC2017 benchmarks} where the $\epsilon$MAg-ES performance was observed to be particularly good and bad, respectively.
 
 Taking into account the benchmark functions for larger dimensions ($N \geq 30$), it turns out that the $\epsilon$MAg-ES performed comparatively poorly on test function $C04$, $C12$, $C20$, and $C21$. \textbf{Notice that these specific constrained problems have inequality constraints only.} Constrained test function $C21$ resembles $C12$ except for the application of a search space rotation by the matrix $\bm{R}$. Due to this similarity and because the rotated problem $C21$ is considered to be more complicated, our investigations will focus on $C21$ instead of $C12$.
 According to the comprehensive results of the CEC2018 competition~\cite{CEC2018comp}
 the $\epsilon$MAg-ES received below-average ranks on these constrained functions. 
 
 In contrast, the $\epsilon$MAg-ES performed comparatively good on test function $C03$, $C06$, $C07$, $C16$, $C18$, $C25$, and $C27$ (for $N \geq 30$). \textbf{It should be noted that all these constrained benchmark problems have at least one equality constraint.} Among these test functions, the constrained problems $C16$ and $C25$, as well as $C18$ and $C27$, differ only by the application of a search space rotation. Due to the repeated use of Rastrigin's function as the objective, the constrained test problem $C06$ is omitted in the following considerations. Below, three selected problems are presented: $C03$, $C07$, and $C27$.
 
 The following paragraphs present the individual test problems and their main features. 
 {
  The rotations and translations occurring in this context are indicated by $\bm{R}\in\mathds{R}^{N\times N}$ and the vector $\bm{o}\in\mathds{R}^N$. The transformations are executed according to the specifications of the constrained CEC2017 benchmark provided in~\cite{CEC2017}.
  }
 
 By taking into account the dynamics of the algorithm variants, the ranking results can be validated and additional insights to the benefits of specific algorithmic components of the $\epsilon$MAg-ES on these particular problems are gained.
 To this end, we consider the mean value dynamics of the sum of the best-so-far fitness and constraint violation values over 25 independent runs.
 {
 These are obtained by recording the respective values in each generation of the algorithm run and averaging over all 25 independent runs. Any differences with regard to the generations required within algorithm runs are treated by cutting off longer runs according to the shortest generation number observed. This approach is appropriate to represent the qualitative dynamic behavior of the algorithms. However, due to the above mentioned procedure, it should not be expected to  observe the exact final mean values that are reported in Tables~\ref{CEC18_MAESD10} and~\ref{CEC18_MAESD100} of the supplementary material.
 }
 The aggregated values are plotted against the number of function evaluations. For each of the curves, the circular marker is indicating the number of function evaluations where the mean constraint violation is zero for a particular algorithm variant. That is, the markers indicate the average runtime needed to reach a feasibility ratio of 100\%. Note that the dark blue curve always represents the dynamics of the $\epsilon$MAg-ES which is the baseline algorithm of this analysis.
 
  {
  In addition to the displayed mean value dynamics, we recap the median and mean performance of all algorithm variants relative to the $\epsilon$MAg-ES for each individual problem. In case of ties with respect to the final (or total) rank, the relative average running time is displayed in accordance with the presentation in Table~\ref{tab2}.
  }
 
  \subsection{Problems of good relative performance}    
 \subsubsection*{Problem $C03$}
 
 The constrained test function $C03$ is composed of the Shifted Schwefel's function and two non-linear constraints: one inequality constraint $g(\bm{x})$ and one equality constraint $h(\bm{x})$.  While both constraints are separable, the objective is non-separable. The constrained problem $C03$ reads 
        \begin{equation}
        \begin{aligned}
          \min  \quad & f(\bm{x}) = \sum\nolimits_{i=1}^{N} \left( \sum\nolimits_{j=1}^{i} z_j \right)^2  \qquad \textrm{with} \quad \bm{z}=\bm{x}-\bm{o} \\
            s.t. \quad  & g(\bm{x}) = \sum\nolimits_{i=1}^{N} \left( z_i^2 -5000 \cos(0.1 \pi z_i) - 4000 \right) \leq 0\\
            & h(\bm{x}) =  -\sum\nolimits_{i=1}^{N}  z_i \sin(0.1 \pi z_i) = 0\\
            & \bm{x} \in [-100,100]^N
        \end{aligned} 
        \label{C03}
        \end{equation}

\begin{table}[t]
\centering
\caption{Results of the comparison of $\epsilon$MAg-ES with six different reduced algorithm variants of itself on the constrained CEC2017 benchmark function $C03$.}
\label{tabC03}
\renewcommand{\arraystretch}{1.1}
\resizebox{0.6\textwidth}{!}{%
\begin{tabular}{lccgccg}
  \multicolumn{1}{l}{$\bm{\epsilon}$\textbf{MAg-ES}}         & \multicolumn{3}{c}{$\bm{N=10}$} & \multicolumn{3}{c}{$\bm{N=100}$} \\
    \multicolumn{1}{l}{}         & \multicolumn{3}{c}{Ranking} & \multicolumn{3}{c}{Ranking} \\
       & Median & Mean & Total & Median & Mean & Total \\ \hline
$\epsilon$MA-ES       & $=$   & $=$        & $0.95$             & $=$  & $=$ & $0.86$\\
$\epsilon$MAg-ES w/o  & $=$   & $=$    & $1.05$                 & $+$  & $+$ & $+$ \\
$\epsilon$MAg-ES nl   & $=$   & $=$        & $1.06$             & $=$  & $=$ & $1.01$\\
$\epsilon$SAg-ES      & $+$   & $+$    & $+$                 & $+$  & $+$ & $+$\\
$lex$MAg-ES      & $+$   & $+$     & $+$                & $+$  & $+$ & $+$\\
$lex$MA-ES       & $+$   & $+$     & $+$                & $+$  & $+$ & $+$\\\hline
\end{tabular}%
}
\end{table}

From Table~\ref{tabC03} one can infer that those algorithm variants that used matrix adaptation and $\epsilon$-level constraint handling do realize equally good results on $C03$ for $N=10$. Considering the relative running times, one observes only rather small deviations of about 5\% compared the baseline $\epsilon$MAg-ES runtime. With growing dimension (see $N=100$), the use of \emph{back-calculation} becomes beneficial on this kind of constrained problem. 

This is substantiated by taking into account the aggregated fitness and constraint violation dynamics in Fig.~\ref{c03-fit}. While the lexicographic versions show a considerably worse performance, the $\epsilon$SAg-ES also is yielding a significantly inferior performance. 
The differences get more pronounced with increasing dimensionality. For $N=100$, one observes that omitting the \emph{back-calculation} has a clear negative effect on the $\epsilon$MAg-ES on problem $C03$.
Looking at the running times, the {\color{black} Jacobian-based} repair components appears to slow down the progress in both cases. In dimension $N=100$, the number of function evaluations is reduced by about 14\%.

 \begin{figure}[t]\centering
           \includegraphics[clip,trim= 80 180 120 180, width=0.4\textwidth,height=0.35\textwidth]{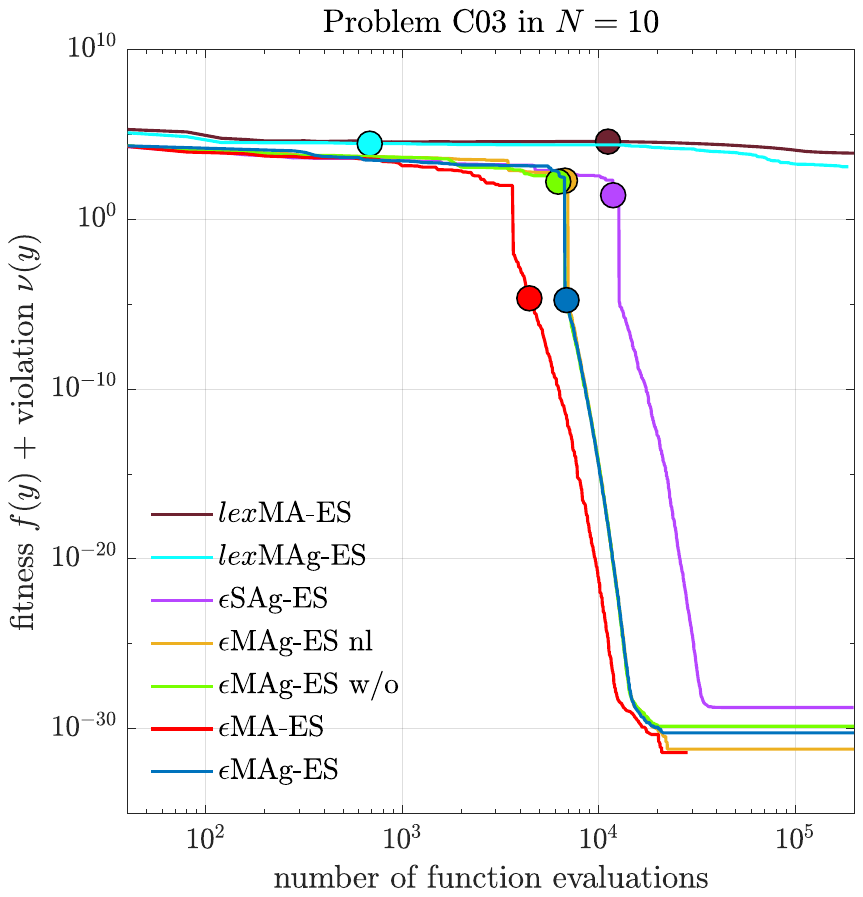} \quad
        \includegraphics[clip,trim= 80 180 120 180, width=0.4\textwidth,height=0.35\textwidth]{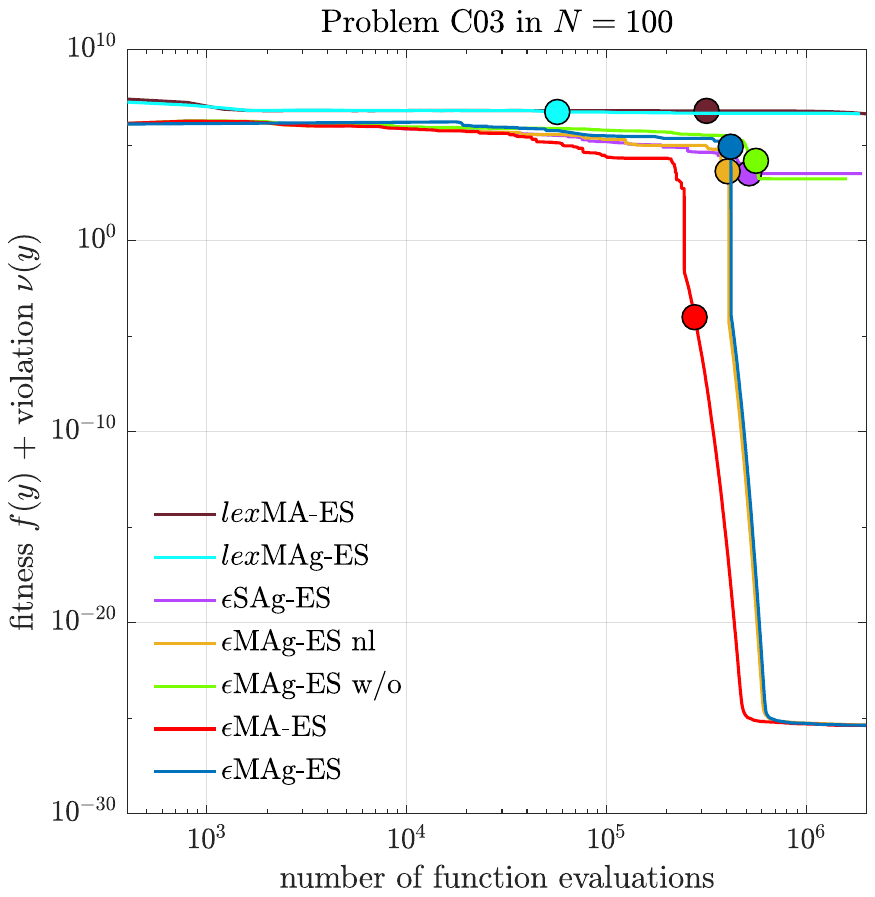} \quad
        \Description{Mean value dynamics on problem $C03$ in dimensions $N=10$ and $N=100$.}
  \caption{Fitness plus constraint violation dynamics on problem $C03$ in dimensions $N=10$ and $N=100$. All curves represent the mean values of 25 independent algorithm runs. The circular markers display the number of function evaluations needed to satisfy all constraints on average.}
  \label{c03-fit}
 \end{figure}
Taking into account the mean mutation stength dynamics corresponding to Fig.~\ref{c03-fit}, one observes that the poorly performing algorithm versions do not sufficiently decrease their mutation strength $\sigma$. These strategies maintain rather diverse offspring populations that are not able to concentrate on the search space region around the optimal solution. Both strategies are not able to further improve on their suboptimal but feasible best found candidate solutions within the allocated budget of function evaluations. On the other hand, the $\epsilon$SAg-ES tends to reduce its mutation strength too rapidly to approach candidate solutions of equal quality in the low dimensional case of $N=10$.

 \begin{figure}[t]\centering
           \includegraphics[clip,trim= 80 180 120 180,width=0.4\textwidth,height=0.35\textwidth]{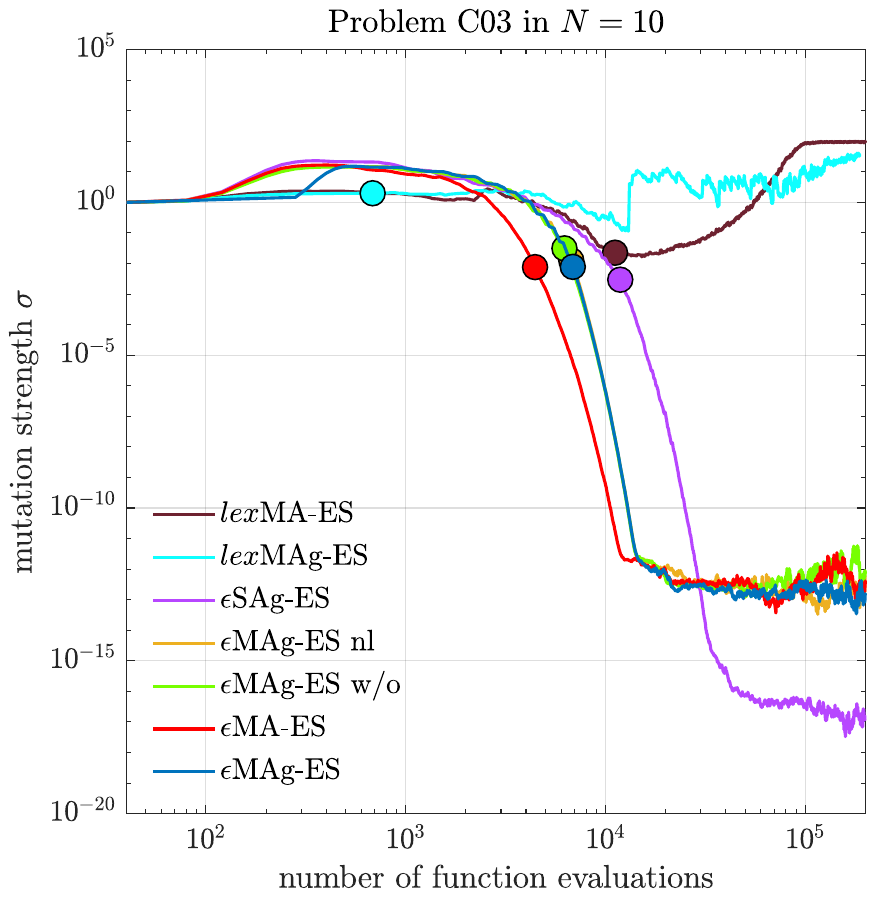}
        \quad\includegraphics[clip,trim= 80 180 120 180,width=0.4\textwidth,height=0.35\textwidth]{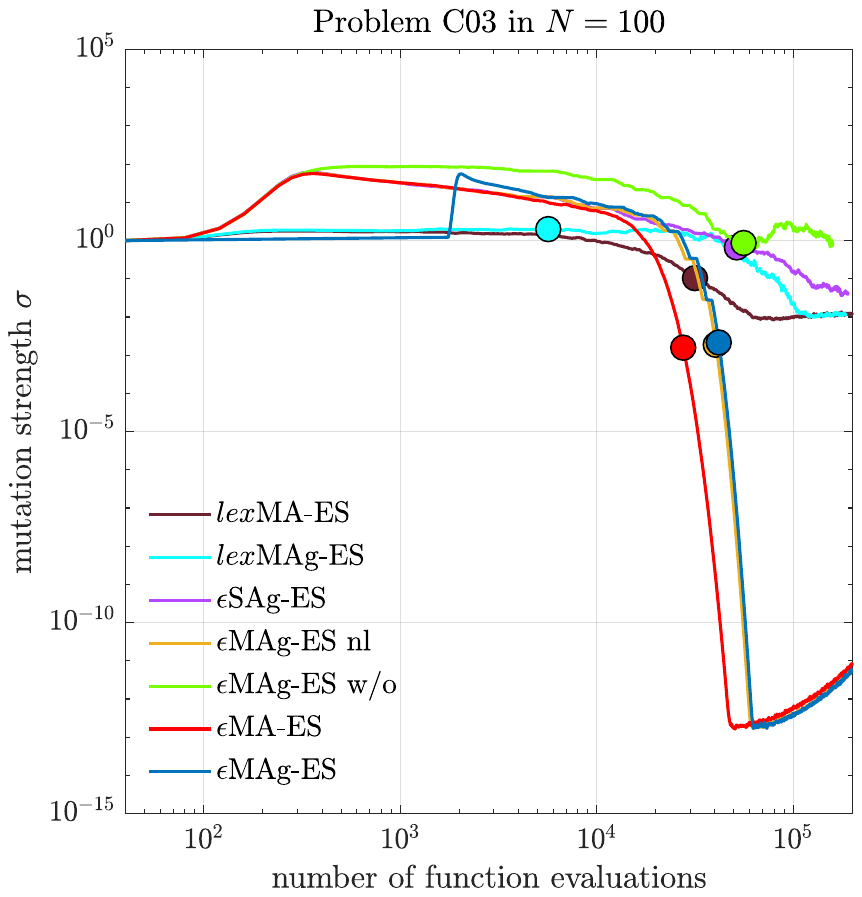} 
        \Description{Mutation strength dynamics on problem $C03$ in dimensions $N=10$ and $N=100$.}
  \caption{Mutation strength dynamics on problem $C03$ in dimensions $N=10$ and $N=100$.}
  \label{c03-sig}
 \end{figure}
\FloatBarrier
  \subsubsection*{Problem $C07$}
   Constrained test function $C07$ consists of a non-linear, but separable, objective function that is subject to two non-linear equality constraints $h_1(\bm{x})$ and $h_2(\bm{x})$. Both constraints are separable. 
   The constrained problem is shown in Eq.~\eqref{C07}. The parameter vector is also subject to a translation by the vector $\bm{o}$.  
  \begin{equation}
        \begin{aligned}
          \min  \quad & f(\bm{x}) = \sum\nolimits_{i=1}^{N}   z_i \sin(z_i)  \qquad \textrm{with} \quad \bm{z}=\bm{x}-\bm{o} \\
            s.t. \quad  & h_1(\bm{x}) = \sum\nolimits_{i=1}^{N} \left( z_i -100 \cos(0.5 z_i)+100\right) = 0\\
            & h_2(\bm{x}) =  -\sum\nolimits_{i=1}^{N} \left( z_i -100 \cos(0.5 z_i)+100\right) = 0\\
            & \bm{x} \in [-50,50]^N
        \end{aligned} 
        \label{C07}
        \end{equation} 
    {
    }
    
    The performance comparison of the algorithm variants relative to $\epsilon$MAg-ES is provided in Table~\ref{tabC07}.
    For low dimensions, a significant advantage of the original $\epsilon$MAg-ES can be observed. This result is also supported by the dynamics illustrated in Fig.~\ref{c07-fit}.
    Only three strategies are able to return a feasible solution in every single run. All of them are using the {\color{black} Jacobian-based} repair approach. Interestingly, in terms of effectiveness the $\epsilon$SAg-ES is closest to the performance of the $\epsilon$MAg-ES. 
    While the $lex$MAg-ES is able to reach the feasible region fastest, it is not able to attain the same solution quality.
    In dimension $N=100$, except for $lex$MA-ES, all variants realize a feasibility ratio of 100\%. The advantage of the $\epsilon$MAg-ES 
    over the reduced versions is not that pronounced anymore.  In terms of running time, there are only marginally performance deviations when disregarding the repair step in $\epsilon$MA-ES.
    Yet, the average $\epsilon$MAg-ES running time is considerably faster than that of the $\epsilon$MAg-ESw/o and $\epsilon$MAg-ESnl variants. This indicates a positive impact of the back-calculation as well as of the $\sigma$ limitation on $C07$ also in high dimensional search spaces.
    Still, $\epsilon$MAg-ES performs significantly better than the lexicographic algorithm variants.

    \begin{table}[t]
\centering
\caption{Results of the comparison of $\epsilon$MAg-ES with six different reduced algorithm variants of itself on the constrained CEC2017 benchmark function $C07$.}
\label{tabC07}
\renewcommand{\arraystretch}{1.1}
\resizebox{0.6\textwidth}{!}{%
\begin{tabular}{lccgccg}
  \multicolumn{1}{l}{$\bm{\epsilon}$\textbf{MAg-ES}}         & \multicolumn{3}{c}{$\bm{N=10}$} & \multicolumn{3}{c}{$\bm{N=100}$} \\
    \multicolumn{1}{l}{}         & \multicolumn{3}{c}{Ranking} & \multicolumn{3}{c}{Ranking} \\
       & Median & Mean & Total & Median & Mean & Total \\ \hline
$\epsilon$MA-ES       & $+$   & $+$    & $+$                  & $=$  & $=$ & $1.02$\\
$\epsilon$MAg-ES w/o  & $+$   & $+$    & $+$                  & $=$  & $=$ & $1.24$ \\
$\epsilon$MAg-ES nl   & $+$   & $+$    & $+$                  & $=$  & $=$ & $1.37$\\
$\epsilon$SAg-ES      & $+$   & $+$    & $+$                  & $=$  & $=$ & $0.60$\\
$lex$MAg-ES      & $+$   & $+$     & $+$                      & $+$  & $+$ & $+$\\
$lex$MA-ES       & $+$   & $+$     & $+$                      & $+$  & $+$ & $+$\\\hline
\end{tabular}%
}
\end{table}   
\begin{figure}[b]\centering
           \includegraphics[clip,trim= 80 180 120 180, width=0.4\textwidth,height=0.35\textwidth]{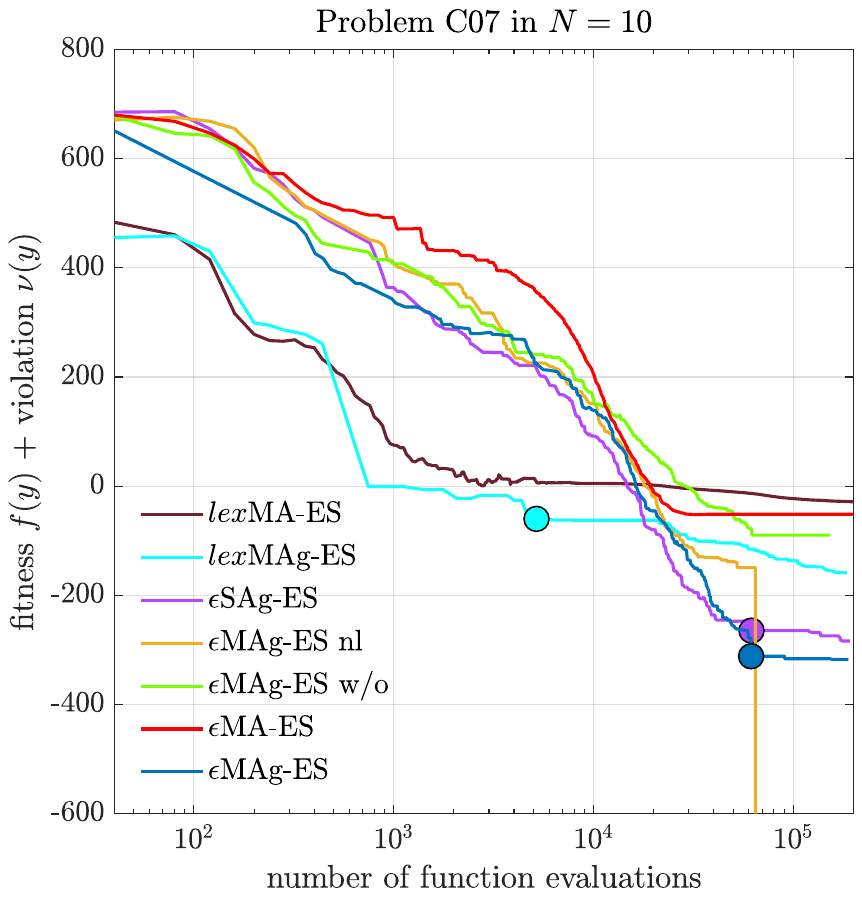} \quad
        \includegraphics[clip,trim= 80 180 120 180, width=0.4\textwidth,height=0.35\textwidth]{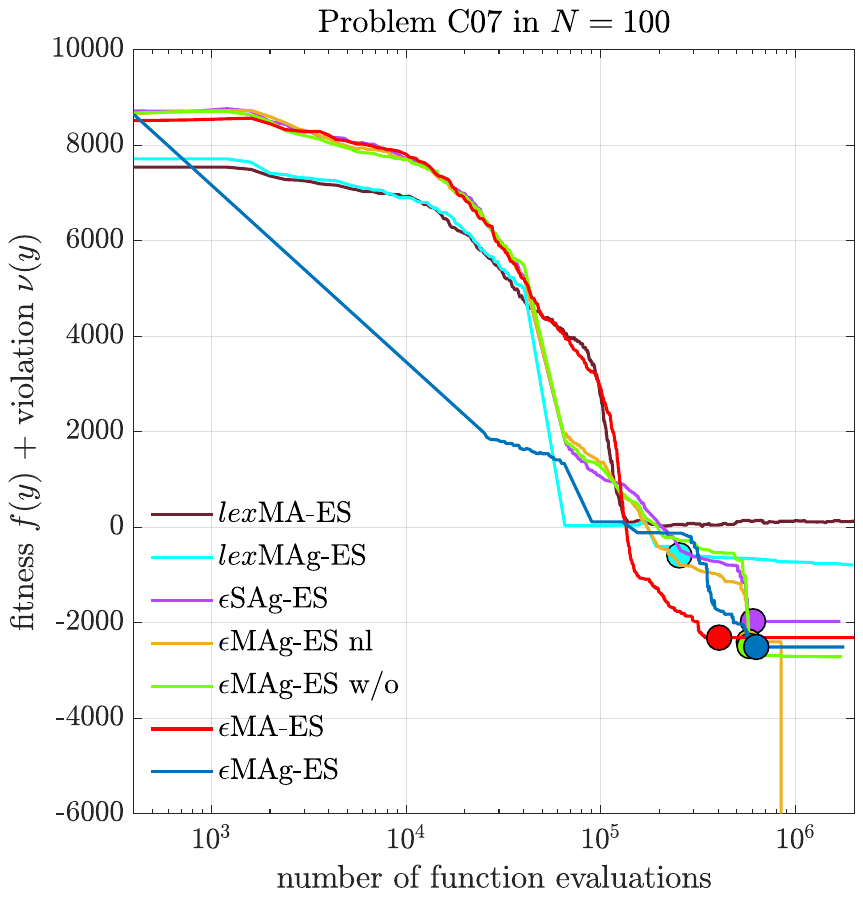} 
        \Description{Mean value dynamics on problem $C07$ in dimensions $N=10$ and $N=100$.}
  \caption{Fitness plus constraint violation dynamics on problem $C07$ in dimensions $N=10$ and $N=100$. All curves represent the mean values of 25 independent algorithm runs. The circular markers display the number of function evaluations needed to satisfy all constraints on average.}
  \label{c07-fit}
 \end{figure}
    
    Interestingly, the matrix adaptation is causing a significant degradation of the running time in $N=100$.
    The success of the SAg-ES indicates that the global structure of the problem has a relatively small conditioning. Accordingly, the use of isotropic mutations is sufficient in this scenario. In combination with the separability of the objective function, the continuous learning of the transformation matrix appears to cause some overhead and might thus simply be disregarded in large dimensions.

   {
   In both dimensions, the limitation of the mutation strength is necessary in the context of problem $C07$. This is underlined by the corrupted mutation strength dynamics (displayed as yellow curves in Fig.~\ref{c07-sig}) that are caused by single runs with deteriorating mutation strength values. In dimension $N=10$, the mutation strength values become NaN and the generated solutions remain infeasible in most runs. In dimension $N=100$, the mutation strength is drastically increased. This way, candidate solutions are generated with components outside the box in individual cases. Due to the numerical problems, the blow-up of individual components cannot be detained by the box-constraint handling. In case of problem $C07$, such impractical candidate solutions outside the box-constraints come with extremely low fitness values. That is, the numerical instabilities also affect the mean fitness and constraint violation dynamics (see the yellow curves in Fig.~\ref{c07-fit}). As a result, the mean fitness values exhibit a sharp drop towards extremely negative values.
   }
   \begin{figure}[t]\centering
           \includegraphics[clip,trim= 80 180 120 180, width=0.425 \textwidth,height=0.35\textwidth]{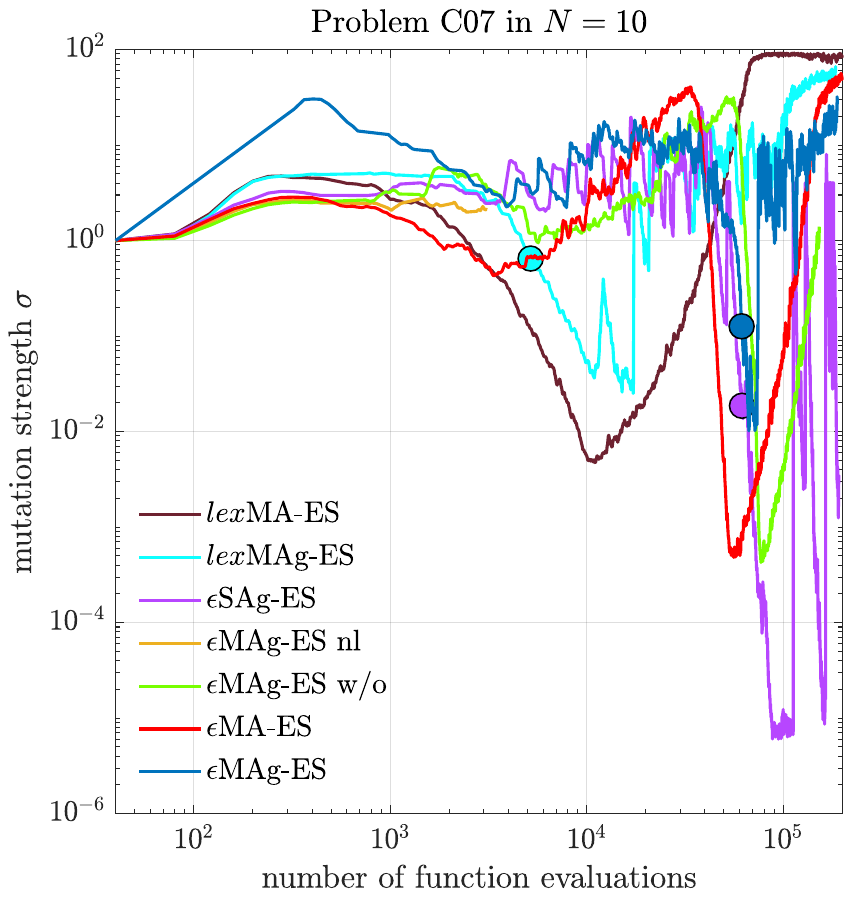} \quad
        \includegraphics[clip,trim= 80 180 120 180, width=0.425 \textwidth,height=0.35\textwidth]{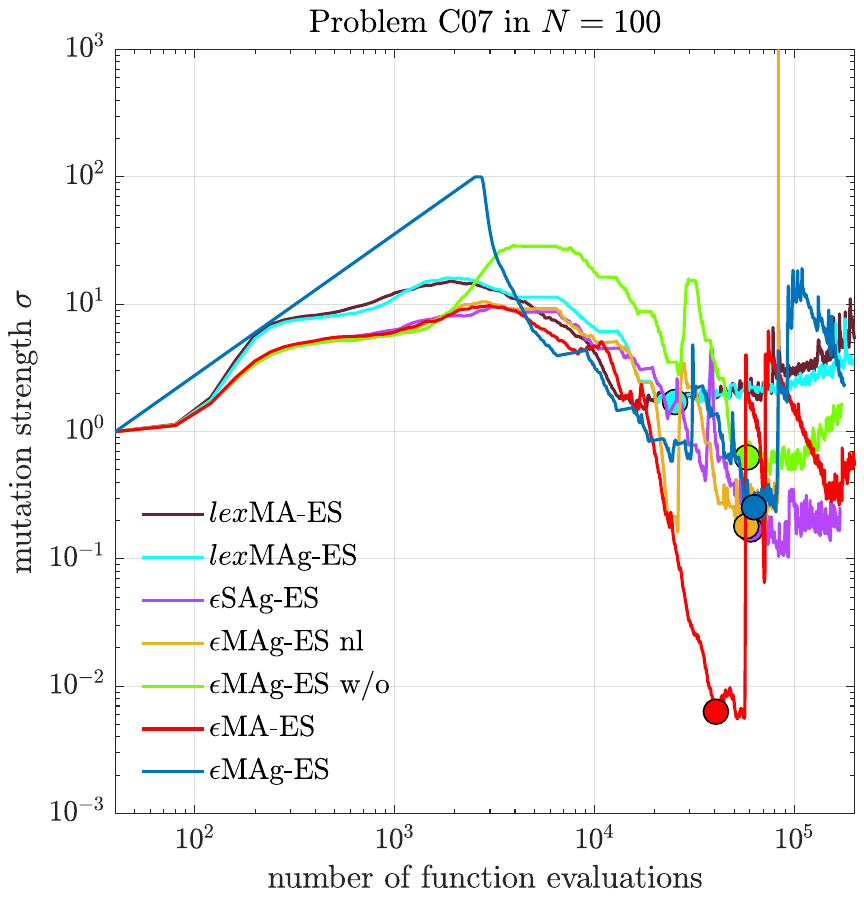}
        \Description{Mean value dynamics on problem $C07$ in dimensions $N=10$ and $N=100$.}
  \caption{Fitness plus constraint violation dynamics on problem $C07$ in dimensions $N=10$ and $N=100$. All curves represent the mean values of 25 independent algorithm runs. The circular markers display the number of function evaluations needed to satisfy all constraints on average.}
  \label{c07-sig}
 \end{figure}
    
          \begin{table}[b]
\centering
\caption{Results of the comparison of $\epsilon$MAg-ES with six different reduced algorithm variants of itself on the constrained CEC2017 benchmark function $C27$.}
\label{tabC27}
\renewcommand{\arraystretch}{1.1}
\resizebox{0.6\textwidth}{!}{%
\begin{tabular}{lccgccg}
  \multicolumn{1}{l}{$\bm{\epsilon}$\textbf{MAg-ES}}         & \multicolumn{3}{c}{$\bm{N=10}$} & \multicolumn{3}{c}{$\bm{N=100}$} \\
    \multicolumn{1}{l}{}         & \multicolumn{3}{c}{Ranking} & \multicolumn{3}{c}{Ranking} \\
       & Median & Mean & Total & Median & Mean & Total \\ \hline
$\epsilon$MA-ES       & $+$   & $+$    & $+$             & $-$  & $-$ & $-$\\
$\epsilon$MAg-ES w/o  & $+$   & $+$    & $+$             & $=$  & $=$ & $1.00$ \\
$\epsilon$MAg-ES nl   & $+$   & $+$    & $+$             & $=$  & $=$ & $1.00$\\
$\epsilon$SAg-ES      & $+$   & $+$    & $+$             & $+$  & $+$ & $+$\\
$lex$MAg-ES      & $=$   & $=$     & $0.84$                & $=$  & $+$ & $+$\\
$lex$MA-ES       & $+$   & $+$     & $+$                & $+$  & $+$ & $+$\\\hline
\end{tabular}%
}
\end{table} 
   \subsubsection*{Problem $C27$}
    One of the most difficult function descriptions is found for constrained problem $C27$. 
      It uses again Rastrigin's function as the objective, but it is applying a second transformation to the search space parameter vectors after rotation. The problem is subject to three non-linear constraints $g_1(\bm{x})$ , $g_2(\bm{x})$, and $h(\bm{x})$. The inequality constraints represent the outside of the $1$-norm's unit ball, as well as the inside of the $N-1$-dimensional hyper sphere with radius $\sqrt{100N}$. The equality constraint is non-separable. 
      
      \noindent Equation~\eqref{C27} provides a mathematical description of constrained problem $C27$.
        \begin{equation}
        \begin{aligned}
          \min  \quad & f(\bm{x}) = \sum\nolimits_{i=1}^{N} \left( z_i^2 -10 \cos(2\pi z_i)+10\right)\\ 
                    & \textrm{with } z_i = \left\{ \begin{matrix}
                                              z_i & \quad \textrm{if } \lvert z_i\rvert <0.5,\\
                                              0.5\cdot \textrm{round}(2z_i) & \textrm{else},
                                            \end{matrix} \right.
 \quad \textrm{and} \quad \bm{z}=\bm{R}(\bm{x}-\bm{o}) \\
            s.t. \quad  & g_1(\bm{x}) = 1- \sum\nolimits_{i=1}^{N} \lvert z_i \rvert  \leq  0\\
             & g_2(\bm{x}) = \sum\nolimits_{i=1}^{N} z_i^2 -100 N \leq  0\\
            & h(\bm{x}) =   \sum\nolimits_{i=1}^{N} 100 (z_i -z_{i+1})^2 + \prod_{i=1}^{N}\sin^2\left((z_i-1)\pi\right) = 0\\
            & \bm{x} \in [-50,50]^N
        \end{aligned} 
        \label{C27}
        \end{equation}

 First, inspecting Table~\ref{tabC27}, one observes differences with respect to the dimensionality. In small dimensions, the use of the lexicographic order relation in combination with repair turns out to result in significantly improved running time. Compared to all other algorithm variants, $\epsilon$MAg-ES is realizing significantly better candidate solutions in dimension $N=10$. As the dimension increases, the small advantage of the lexicographical order relation is lost.  Instead, the $\epsilon$MA-ES becomes the only algorithm variant that is able to perform better than the $\epsilon$MAg-ES. Hence, the {\color{black} Jacobian-based} repair component must not be overestimated in the context of constraint problems that include equality constraints in comparably high dimensions. At the same time the use of back-calculation and mutation strength limitation does not seem to have much impact. The respective variants perform similar to $\epsilon$MAg-ES in terms of solution quality as well as running time. These observations would also be supported by zooming into the rear part of the dynamics displayed in Fig.~\ref{c27-fit}.
      
    Note that the neglection of the matrix adaptation leads to inability of realizing a feasibility rate of 100\% in both considered dimensionalities. In dimension $N=100$, also the lexicographic algorithm versions ($lex$MAg-ES and $lex$MA-ES) can no longer detect the feasible region in 100\% of the 25 runs.
        
   \begin{figure}[b]\centering
           \includegraphics[clip,trim= 80 180 120 180, width=0.4\textwidth,height=0.35\textwidth]{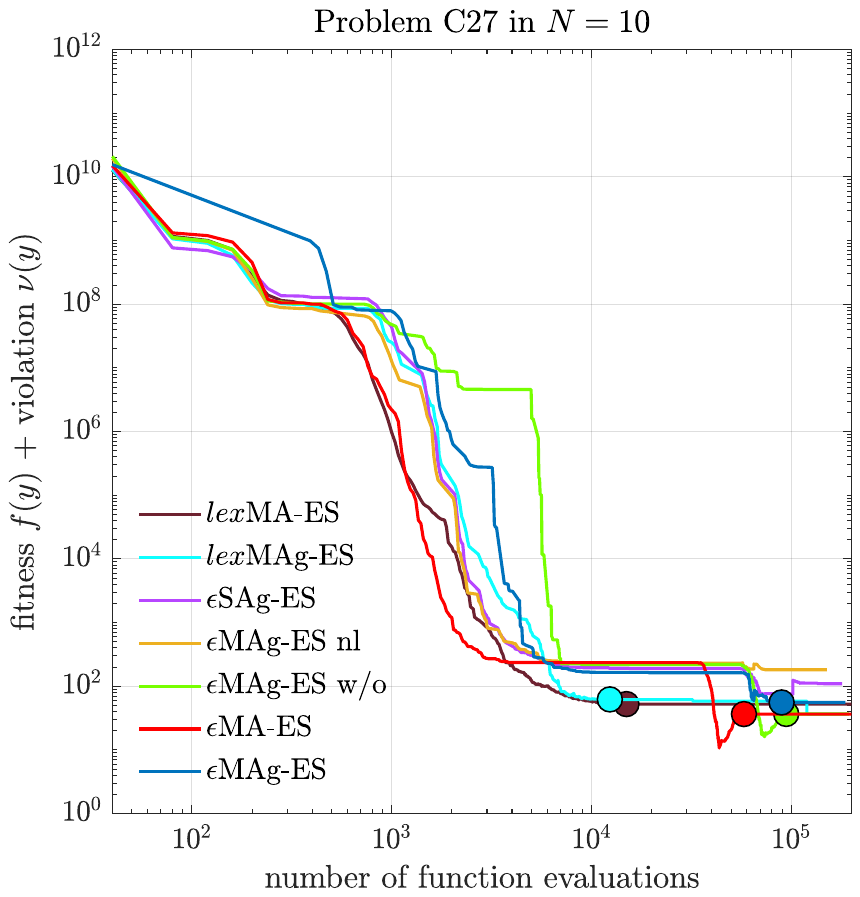} \quad
        \includegraphics[clip,trim= 80 180 120 180, width=0.4\textwidth,height=0.35\textwidth]{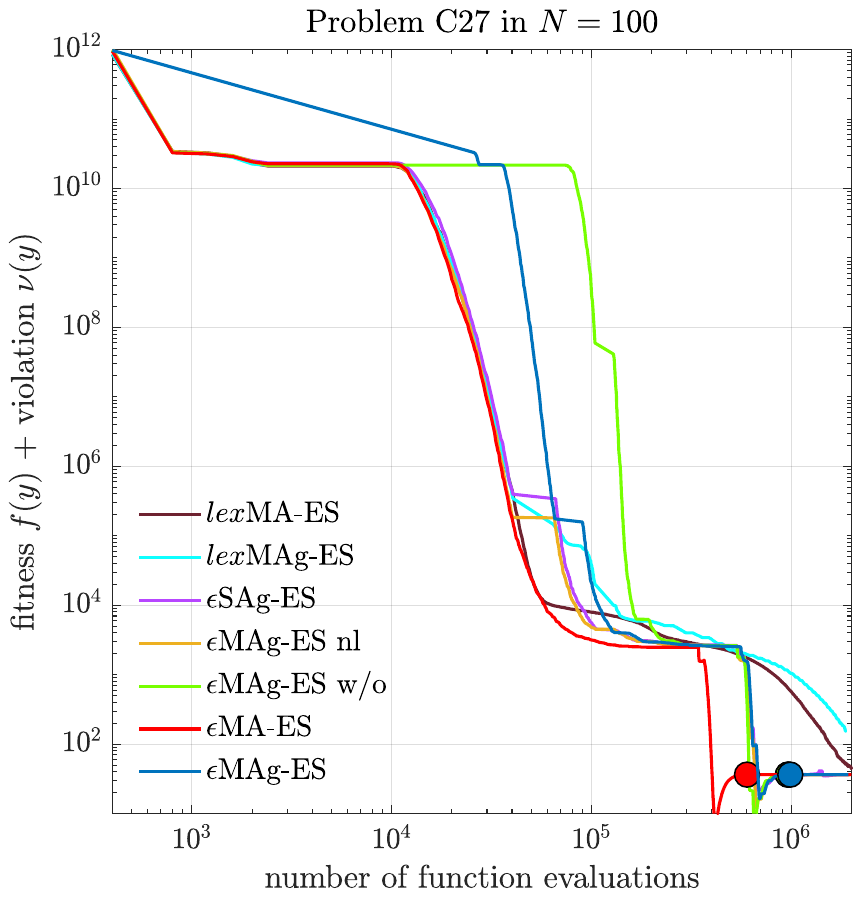} \quad
        \Description{Mean value dynamics on problem $C27$ in dimensions $N=10$ and $N=100$.}
  \caption{Fitness plus constraint violation dynamics on problem $C27$ in dimensions $N=10$ and $N=100$. All curves represent the mean values of 25 independent algorithm runs. The circular markers display the number of function evaluations needed to satisfy all constraints on average.}
  \label{c27-fit}
 \end{figure}
 
   \newpage
  \subsection{Problems of low relative performance}
   \subsubsection*{Problem $C04$}
   Test function $C04$ demands to minimize Rastrigin's function subject to two non-linear inequality constraints $g_1(\bm{x})$ and $g_2(\bm{x})$. 
      The constrained problem is displayed in Eq.~\eqref{C04}.
      Both, the objective as well as the constraint functions are separable.
        \begin{equation}
        \begin{aligned}
          \min  \quad & f(\bm{x}) = \sum\nolimits_{i=1}^{N} \left( z_i^2 -10 \cos(2\pi z_i)+10\right) \qquad \textrm{with} \quad \bm{z}=\bm{x}-\bm{o}\\
            s.t. \quad  & g_1(\bm{x}) = -\sum\nolimits_{i=1}^{N}  z_i \sin(2z_i)\leq 0\\
            & g_2(\bm{x}) =  \sum\nolimits_{i=1}^{N}  z_i \sin(z_i) \leq 0\\
            & \bm{x} \in [-10,10]^N
        \end{aligned} 
        \label{C04}
        \end{equation}
 Regarding Table~\ref{tabC04}, one notices the benefit of $\epsilon$-level ordering on this constrained test function.
 Although both lexicographic algorithm variants do find feasible candidate solutions within the initially sampled population (displayed by the circular markers on the left-hand side of the plots in Fig.~\ref{c04-fit}), they are not able to match the quality of the solutions returned by the other $\epsilon$MAg-ES versions.

\begin{table}[t]
\centering
\caption{Results of the comparison of $\epsilon$MAg-ES with six different reduced algorithm variants of itself on the constrained CEC2017 benchmark function $C04$.}
\label{tabC04}
\renewcommand{\arraystretch}{1.1}
\resizebox{0.6\textwidth}{!}{%
\begin{tabular}{lccgccg}
  \multicolumn{1}{l}{$\bm{\epsilon}$\textbf{MAg-ES}}         & \multicolumn{3}{c}{$\bm{N=10}$} & \multicolumn{3}{c}{$\bm{N=100}$} \\
    \multicolumn{1}{l}{}         & \multicolumn{3}{c}{Ranking} & \multicolumn{3}{c}{Ranking} \\
       & Median & Mean & Total & Median & Mean & Total \\ \hline
$\epsilon$MA-ES       & $-$   & $-$        & $-$             & $-$  & $-$ & $-$\\
$\epsilon$MAg-ES w/o  & $-$   & $-$    & $-$                 & $-$  & $+$ & $1.35$ \\
$\epsilon$MAg-ES nl   & $=$   & $=$        & ${ 0.51}$             & $=$  & $=$ & $1.10$\\
$\epsilon$SAg-ES      & $=$   & $=$    & ${0.85}$                 & $=$  & $=$ & $0.97$\\
$lex$MAg-ES      & $+$   & $+$     & $+$                     & $+$  & $+$ & $+$\\
$lex$MA-ES       & $+$   & $+$     & $+$                     & $+$  & $+$ & $+$\\\hline
\end{tabular}%
}
\end{table}  
\begin{figure}[b]\centering
           \includegraphics[clip,trim= 80 180 120 180, width=0.465\textwidth,height=0.4\textwidth]{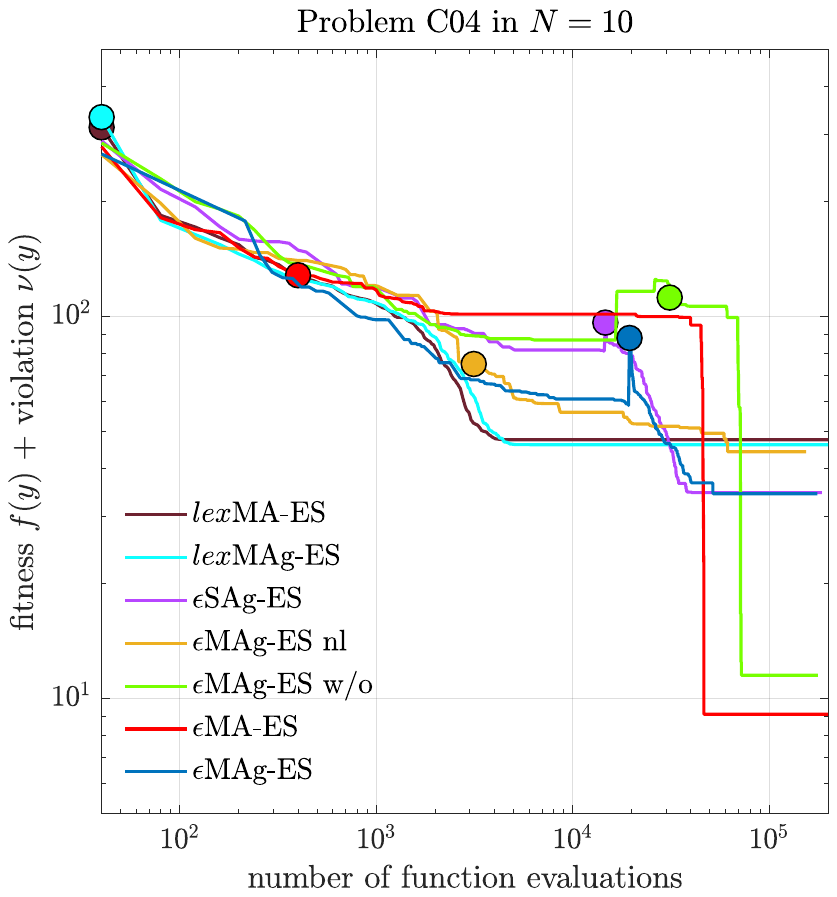} \quad
        \includegraphics[clip,trim= 80 180 120 180, width=0.465\textwidth,height=0.4\textwidth]{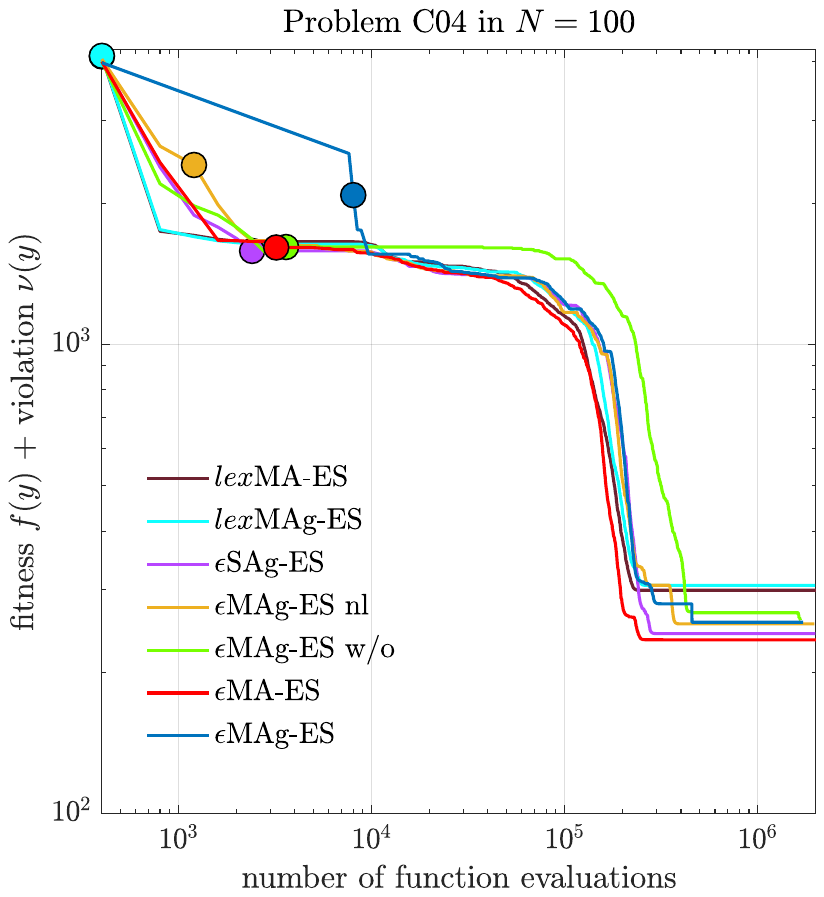} \quad
        \Description{Mean value dynamics on problem $C04$ in dimensions $N=10$ and $N=100$.}
  \caption{Fitness plus constraint violation dynamics on problem $C04$ in dimensions $N=10$ and $N=100$. All curves represent the mean values of 25 independent algorithm runs. The circular markers display the number of function evaluations needed to satisfy all constraints on average.}
  \label{c04-fit}
 \end{figure}
  Further, the mutation strength limitation as well as the matrix adaptation do not have a significant contribution to the algorithm effectiveness.
 
 While these algorithmic components result in improved running times in low dimensions, this effect vanishes with growing search space dimensionality. {\color{black} Jacobian-based} repair in combination with \emph{back-calculation} appears to have a negative influence on the performance in both considered dimensions. Neglecting the \emph{back-calculation} can balance the disadvantage of the repair method in low dimensions, but it impairs the mean value realizations in larger dimensions. Hence, in the total ranking, it is performing similarly to $\epsilon$MAg-ES but exhibits a worse average running time, see Table~\ref{tab2}.

 The illustration of the algorithm dynamics in Fig.~\ref{c04-fit} indicates the advantage of the $\epsilon$MA-ES on probelm $C04$. While the improvement in terms of running time can be regarded as negligible, the repair step prevents the algorithm from realizing significantly better candidate solutions. Apart from the slight differences in the mean value dynamics displayed in Fig.~\ref{c04-fit} for $N=100$, the algorithm realizations of $\epsilon$MAg-ES, $\epsilon$SAg-ES, and $\epsilon$MAg-ES nl are considered to be similarly distributed with respect to the Wilcoxon Signed-Rank test.

    \subsubsection*{Problem $C20$}
     The objective of constrained test function $C20$ represents the sum of $N$ Schaffer's F6 functions $h(y_1,y_2)$. Note that the denominator is slightly modified by taking into account the square root of the sum. The global minimum of these components is at $h(0,0) = 0$. Again, two non-linear inequality constraints must be satisfied.  It is non-separable.
        \begin{equation}
        \begin{aligned}
          \min  \quad & f(\bm{x}) = \sum\nolimits_{i=1}^{N-1}  h(z_i,z_{i+1}) + h(z_N,z_1)
           \qquad \textrm{with} \quad \bm{z}=\bm{x}-\bm{o}, \\
          &\qquad \textrm{and} \quad h(y_1,y_2) =\left(0.5 + \cfrac{\sin^2\left( \sqrt{ y_1^2 + y_2^2 } \right)-0.5}{\left(1+10^{-3}\left(\sqrt{ y_1^2 + y_2^2 } \right)  \right)^2 } \right) \\
            s.t. \quad  & g_1(\bm{x}) = \cos^2\left(\sum\nolimits_{i=1}^{N} z_i \right) - 0.25 \cos\left(\sum\nolimits_{i=1}^{N} z_i  \right) - 0.125 \leq 0\\
            & g_2(\bm{x}) = \exp\left( \cos\left(\sum\nolimits_{i=1}^{N} z_i  \right) \right) - \exp(0.25) \leq 0\\
            & \bm{x} \in [-100,100]^N
        \end{aligned} 
        \label{C20}
        \end{equation} 
        \begin{table}[b]
    \centering
    \caption{Results of the comparison of $\epsilon$MAg-ES with six different reduced algorithm variants of itself on the constrained CEC2017 benchmark function $C20$.}
    \label{tabC20}
    \renewcommand{\arraystretch}{1.2}
    \resizebox{0.6\textwidth}{!}{%
        \begin{tabular}{lccgccg}
        \multicolumn{1}{l}{$\bm{\epsilon}$\textbf{MAg-ES}}         & \multicolumn{3}{c}{$\bm{N=10}$} & \multicolumn{3}{c}{$\bm{N=100}$} \\
            \multicolumn{1}{l}{}         & \multicolumn{3}{c}{Ranking} & \multicolumn{3}{c}{Ranking} \\
            & Median & Mean & Total & Median & Mean & Total \\ \hline
        $\epsilon$MA-ES       & $-$   & $-$    & $-$                 & $-$  & $-$ & $-$\\
        $\epsilon$MAg-ES w/o  & $-$   & $-$    & $-$                 & $-$  & $-$ & $-$ \\
        $\epsilon$MAg-ES nl   & $-$   & $-$    & $-$                 & $-$  & $-$ & $-$\\
        $\epsilon$SAg-ES      & $-$   & $-$    & $-$                 & $-$  & $-$ & $-$\\
        $lex$MAg-ES             & $=$   & $=$     & $1.05$                     & $-$  & $-$ & $-$\\
        $lex$MA-ES              & $-$   & $-$     & $-$                     & $-$  & $-$ & $-$\\\hline
        \end{tabular}%
    }
\end{table}
      Interestingly, independent of the dimensionality the $\epsilon$MAg-ES is always performing only similarly or worse than the reduced algorithm variants. This observation is in line with the inferior performance of the $\epsilon$MAg-ES that was observed in the CEC2018 competition. While the lexicographic algorithm variants do observe feasible solutions immediately after initialization of the starting population (see the corresponding markers in Fig.~\ref{c20-fit}), the variants that use $\epsilon$-level ordering account the fitness of infeasible candidates for the best result so far due to the relaxation of the feasible region. With reduction of the $\epsilon$-level, these versions of the $\epsilon$MAg-ES do also always find the feasible region.
      Yet, the combination of back-calculation and {\color{black} Jacobian-based} repair appears to have a negative impact on the strategy's ability to find the feasible region. The algorithmic variants that omit either one of those components do perform best on constrained problem $C20$. As illustrated in Fig.~\ref{c20-fit}, the best algorithm variants do improve the fitness observed by the original $\epsilon$MAg-ES by a factor of $10$ within the given function evaluation budget.

   \begin{figure}[t]\centering
           \includegraphics[clip,trim= 80 180 120 180, width=0.465\textwidth,height=0.44\textwidth]{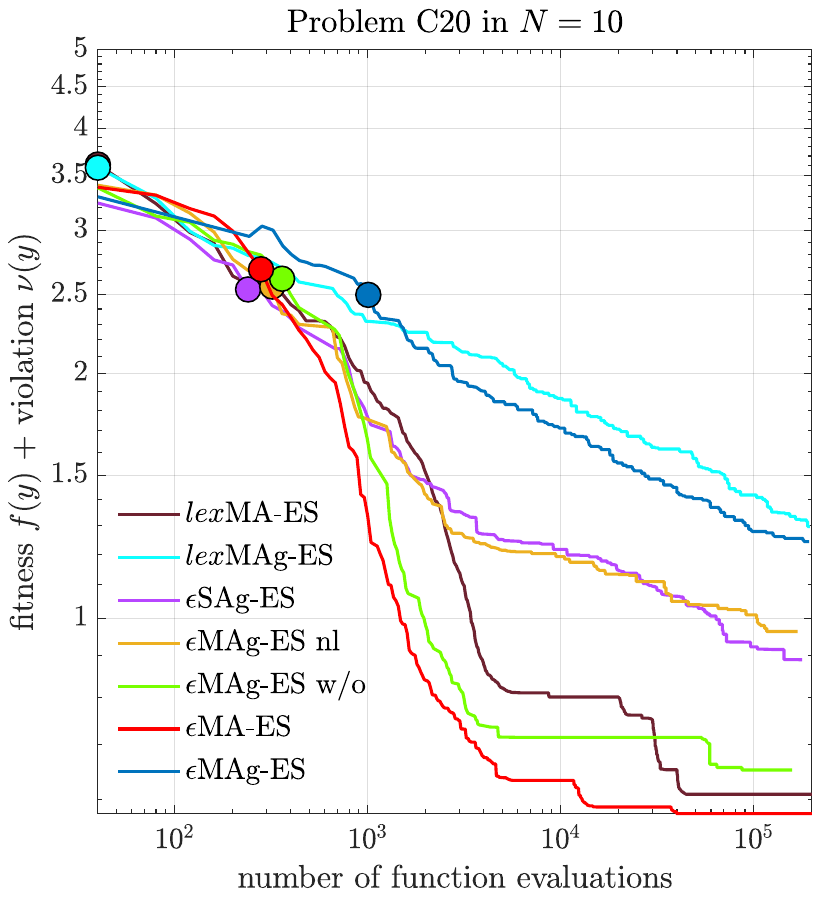} \quad
        \includegraphics[clip,trim= 80 180 120 180, width=0.465\textwidth,height=0.44\textwidth]{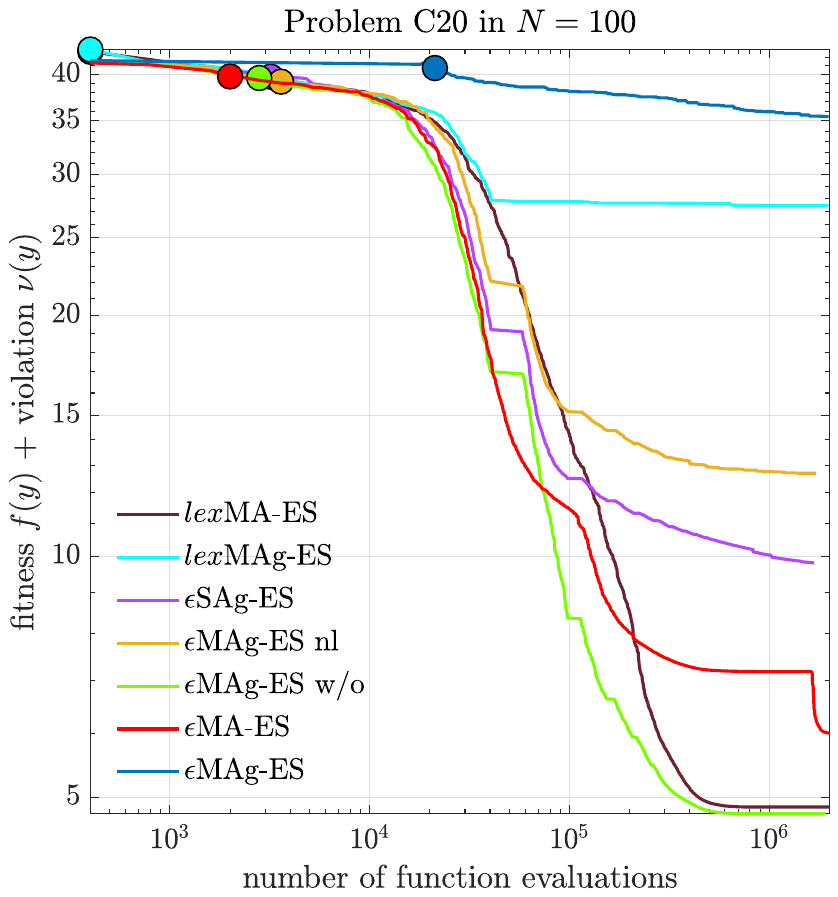} \quad
        \Description{Mean value dynamics on problem $C20$ in dimensions $N=10$ and $N=100$.}
  \caption{Fitness plus constraint violation dynamics on problem $C20$ in dimensions $N=10$ and $N=100$. All curves represent the mean values of 25 independent algorithm runs. The circular markers display the number of function evaluations needed to satisfy all constraints on average.}
  \label{c20-fit}
 \end{figure}
 
    \subsubsection*{Problem $C21$}
    This constrained test function again considers Rastrigin's function as its objective. The global minimum of the Rastrigin function is found at $f(\bm{0}) = 0$. In contrast to Eq.~\eqref{C04}, two different non-linear inequality constraints $g_1(\bm{x})$ and $g_2(\bm{x})$ are introduced and different box-constraints are used. 
      The objective as well as the constraint functions are separable.
        \begin{equation}
        \begin{aligned}
          \min  \quad & f(\bm{x}) = \sum\nolimits_{i=1}^{N} \left( z_i^2 -10 \cos(2\pi z_i)+10\right) \qquad \textrm{with} \quad \bm{z}=\bm{M}(\bm{x}-\bm{o}) \\
             s.t. \quad  & g_1(\bm{x}) = 4-\sum\nolimits_{i=1}^{N} \lvert z_i \rvert \leq 0\\
            & g_2(\bm{x}) =  \sum\nolimits_{i=1}^{N}  z_i^2  - 4 \leq 0\\
            & \bm{x} \in [-100,100]^N
        \end{aligned} 
        \label{C21}
        \end{equation}
        
    Considering the low dimensional results in Tab.~\ref{tabC21}, one observes that all strategy variants show a similar performance with respect to the median ranking. Only when considering the mean value dynamics, the $lex$MA-ES can yield a significant improvement. When taking into account large dimensions, the slight advantage of the $lex$MA-ES over the $\epsilon$MAg-ES vanishes. However, the lexicographic variants do approach their final realizations reasonably faster on average (see Fig.~\ref{c21-fit}).
    In dimension $N=100$ the use of back-calculation as well as the matrix adaptation component do turn out to be clearly beneficial
    regardless of the dimension. The limitation of the mutation strength does not have a considerable impact on the $\epsilon$MAg-ES performance on problem $C21$.    
    \begin{table}[hbtp]
    \centering
    \caption{Results of the comparison of $\epsilon$MAg-ES with six different reduced algorithm variants of itself on the constrained CEC2017 benchmark function $C21$.}
    \label{tabC21}
    \renewcommand{\arraystretch}{1.2}
    \resizebox{0.6\textwidth}{!}{%
        \begin{tabular}{lccgccg}
        \multicolumn{1}{l}{$\bm{\epsilon}$\textbf{MAg-ES}}         & \multicolumn{3}{c}{$\bm{N=10}$} & \multicolumn{3}{c}{$\bm{N=100}$} \\
            \multicolumn{1}{l}{}         & \multicolumn{3}{c}{Ranking} & \multicolumn{3}{c}{Ranking} \\
            & Median & Mean & Total & Median & Mean & Total \\ \hline
        $\epsilon$MA-ES         & $=$   & $=$    & $0.99$               & $=$  & $=$ & $0.80$\\
        $\epsilon$MAg-ES w/o    & $=$   & $=$    & $1.09$               & $+$  & $+$ & $+$ \\
        $\epsilon$MAg-ES nl     & $=$   & $=$    & $0.95$               & $=$  & $=$ & $0.98$\\
        $\epsilon$SAg-ES        & $=$   & $=$    & $0.79$               & $+$  & $+$ & $+$\\
        $lex$MAg-ES             & $=$   & $=$     & $0.66$              & $=$  & $=$ & $0.73$\\
        $lex$MA-ES              & $=$   & $-$     & $-$              & $=$  & $=$ & $0.67$\\\hline
        \end{tabular}%
        }
    \end{table}       
        
   \begin{figure}[htbp]\centering
           \includegraphics[clip,trim= 80 180 120 180, width=0.465\textwidth,height=0.44\textwidth]{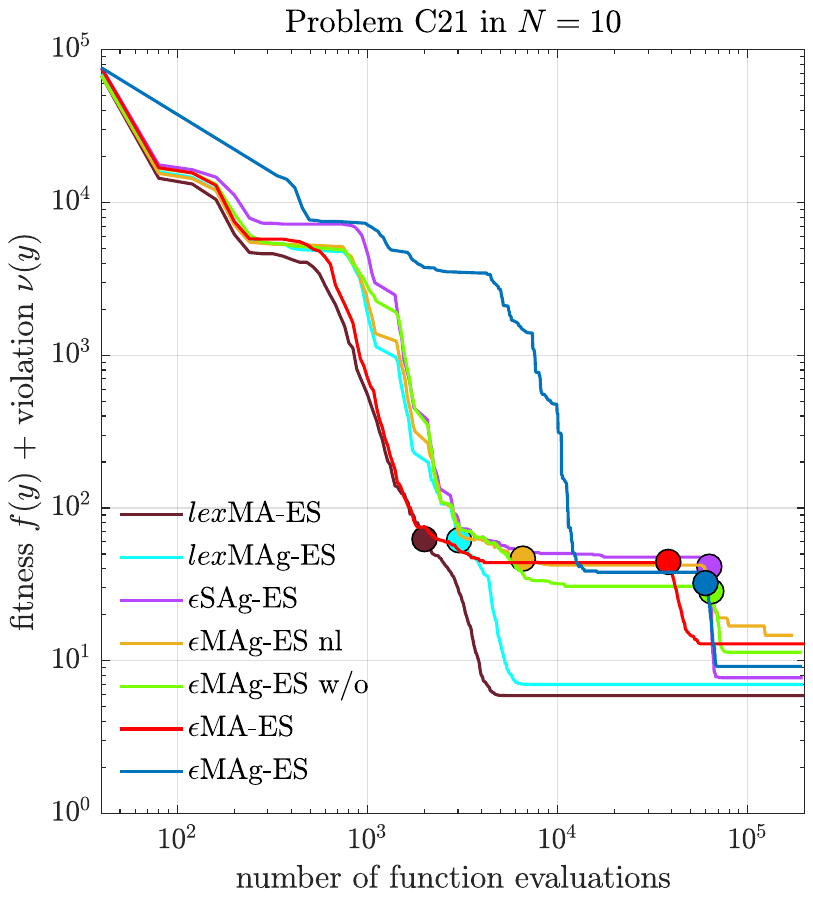} \quad
        \includegraphics[clip,trim= 80 180 120 180, width=0.465\textwidth,height=0.44\textwidth]{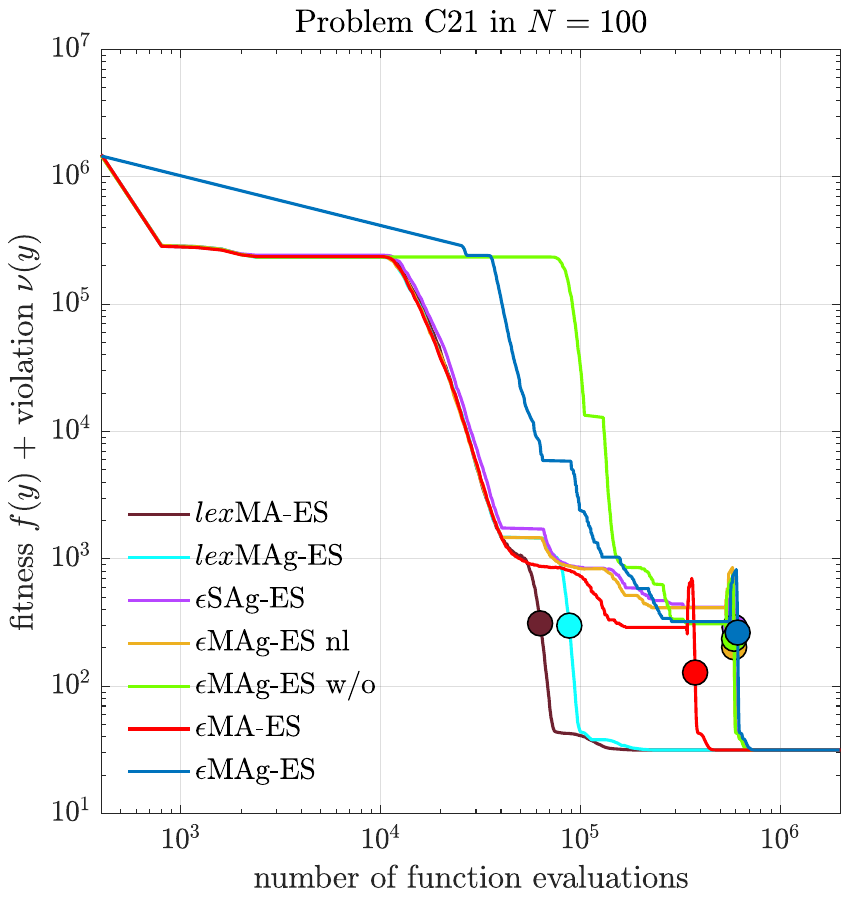} \quad
        \Description{Mean value dynamics on problem $C21$ in dimensions $N=10$ and $N=100$.}
  \caption{Fitness plus constraint violation dynamics on problem $C21$ in dimensions $N=10$ and $N=100$. All curves represent the mean values of 25 independent algorithm runs. The circular markers display the number of function evaluations needed to satisfy all constraints on average.}
  \label{c21-fit}
 \end{figure}

    \section{Impact of selected strategy parameters}
    \label{sec06} 
     Having concentrated on the impact of the individual algorithm components above, one is concerned with the empirical examination of a useful population size for the $\epsilon$MAg-ES on the constrained CEC2017 benchmarks in Sec.\ref{sec061}. Additionally, in Sec.~\ref{sec062} the impact of the number of repeated repair steps $\theta_r$ (see Alg.~\ref{alg_MAES} in line 17) is considered. 
 
    \subsection{On the population size parameters}
    \label{sec061}   
    Having concentrated on the impact of the individual algorithm components above, this section is concerned with the empirical examination of a useful population size for the $\epsilon$MAg-ES on the constrained CEC2017 benchmarks.  
    
    The $\epsilon$MAg-ES in Alg.~\ref{alg_MAES} is configured in accordance with the standard strategy parameter choices for the Matrix Adaptation Evolution Stragtegy (MA-ES)~\cite{BeyerS2017,LoshchilovGlasmachersBeyer2018} which are adopted from the CMA-ES~\cite{Hansen2003} context. These strategy parameter recommendations for MA-ES are based on elaborated empirical experiments in unconstrained settings. The recommendations mostly turn out to be useful choices for constrained optimization problems. The CMA-ES specification recommends to use an offspring population size of $\lambda = 4 + \lfloor 3 \ln(N)\rfloor$ and a parental population size of $\mu = \lfloor\lambda/ 2\rfloor$. In this respect, it differs from the default population sizes used by the $\epsilon$MAg-ES, Alg.~\ref{alg_MAES}, $\lambda = 4N$ and $\mu =  \lfloor\lambda /3\rfloor$, respectively. Table~\ref{tabPop1} illustrates the difference in magnitude of the two population size settings. The choice of such large populations for $\epsilon$MAg-ES in~\cite{HellwigB2018} is based on the fact that significant performance improvements have been observed during the design process of the $\epsilon$MAg-ES in small dimensions.
    \begin{table}[htbp]
    \centering
    \caption{Difference in the population size recommendations.}
    \label{tabPop1}
    \renewcommand{\arraystretch}{1.1}
        \begin{tabular}{lcc}
            Dimension & MA-ES~\cite{LoshchilovGlasmachersBeyer2018} & $\epsilon$MAg-ES~\cite{HellwigB2018} \\\hline
            N          & $\lambda = 4 + \lfloor 3 \ln(N)\rfloor$, $\mu = \lfloor\lambda/ 2\rfloor$ & $\lambda = 4N$, $\mu = \lfloor\lambda /3\rfloor$  \\ 
            10         & $\lambda = 10$, $\mu =5$ & $\lambda = 40$, $\mu =13 $ \\
            100        & $\lambda = 17$, $\mu =8$ & $\lambda = 400$, $\mu =133 $
        \end{tabular}
    \end{table}
    
    In order to properly assess the effect of different population sizes on the $\epsilon$MAg-ES performance, we consider the following setting. The $\epsilon$MAg-ES is run with 7 different offspring population size choices in dimension $N=10$ and $N=100$, respectively.
    For $N=10$, the population sizes $\lambda \in \{3, 10, 20 , 40 , 80, 100, 200\}$ are considered. For $N=100$, we use the population sizes $\lambda \in \{3, 10, 100, 200, 400, 1000, 2000\}$. The truncation ratio of about $1/3$ is kept constant. After having executed 25 independent $\epsilon$MAg-ES runs with each considered populations size, the algorithm configurations are compared for median and mean  realization effectiveness. 
    \begin{figure}[t]
      \includegraphics[clip,trim= 40 225 50 230, width=0.8\textwidth]{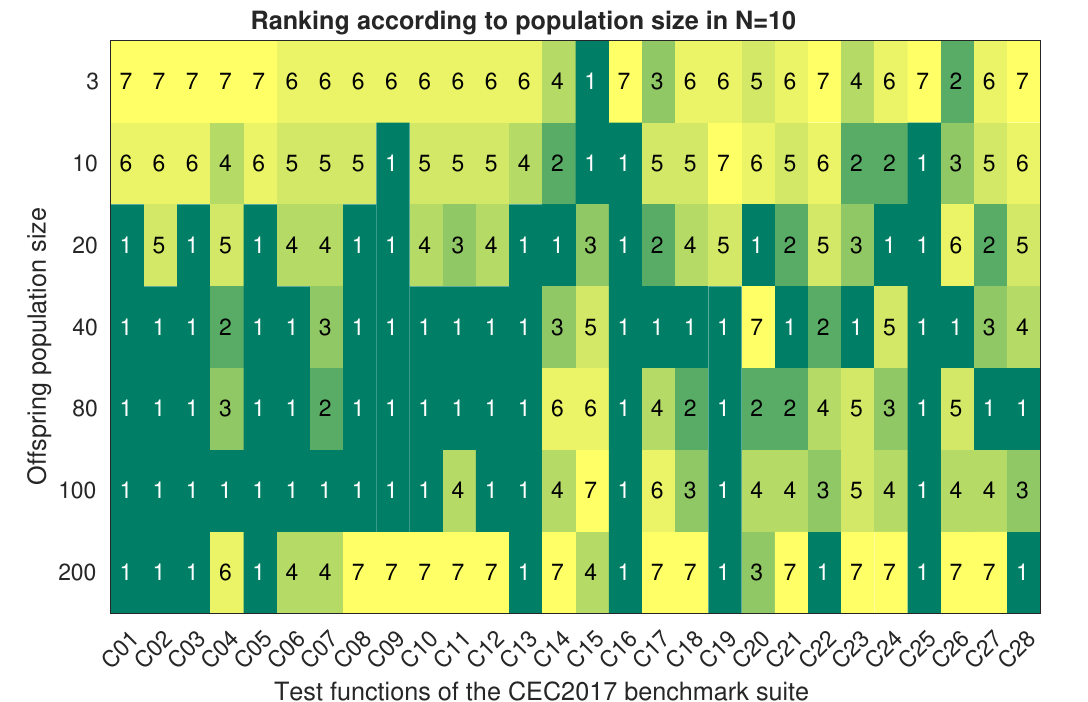}
     \caption{Impact of the offspring population size on the $\epsilon$MAg-ES performance in $N=10$. Note that the standard choice is $\lambda =40$ and $\mu = 13$ in this dimension.}
     \Description{Impact of the offspring population size on the $\epsilon$MAg-ES performance in $N=10$. Note that the standard choice is $\lambda =40$ and $\mu = 13$ in this dimension.}
     \label{heatN10}
    \end{figure}
    {
    We calculate the configuration rankings according to the CEC2017 recommendations. 
    Hence, all configurations are ranked with respect to the median and mean value comparison mentioned above. For the final ranking, both  ranking scores are aggregated. The configuration with the lowest aggregated ranking score receives the total rank 1 and so on.
    Algorithm configurations that obtain similar overall ranking scores obtain the same ranking number.
    }
    The total ranking results are presented by heat maps displaying the configuration ranking per constrained problem. In this representation, the configurations are graded by their rank values as well as the color of the tiles. Darker tiles correspond to better configurations.
    The results for $N=10$ and $N=100$ are displayed in Fig.~\ref{heatN10} and Fig.~\ref{heatN100}, respectively.

     \begin{figure}[htbp]
     \includegraphics[clip,trim= 40 225 50 230, width=0.8\textwidth]{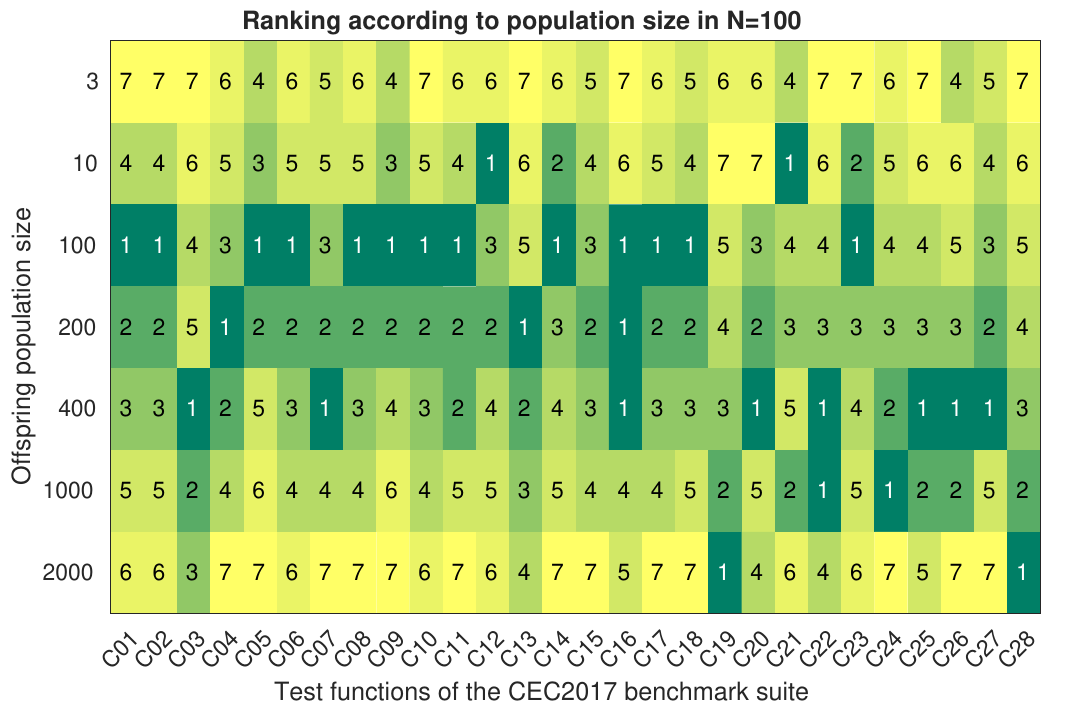}
     \caption{Impact of the offspring population size on the $\epsilon$MAg-ES performance in $N=100$. Note that the standard choice is $\lambda =400$ and $\mu = 133$ in this dimension.}
     \Description{Impact of the offspring population size on the $\epsilon$MAg-ES performance in $N=100$.}
     \label{heatN100}
    \end{figure}
    In Fig.~\ref{heatN10} we observe a particular tendency towards relatively large population sizes (compared to the MA-ES standard recommendation) over the first half of the constrained benchmark problems. On eight functions, the choice of the population size greater than $\lambda=20$ has no further influence on the effectiveness of the $\epsilon$MAg-ES. 
    However, on 13 out of 28 constrained functions too large population sizes appear to have a negative effect on the performance as the $\epsilon$MAg-ES with $\lambda=200$ receives the lowest ranks.
    While very large $\lambda$ values seem advantageous on problems $C22$ and $C28$, constrained function $C15$ gives preference to small values of $\lambda$. However, these three functions are the exception. In most cases, medium population sizes must be preferred. The default choice of $\lambda=4N$ represents a reasonable compromise for $N=10$. The $\epsilon$MAg-ES with $\lambda=40$ receives the best ranking on 19 out of 28 constrained problems. By summing up the problem-specific ranks and dividing by the number of problems, the average ranks of each configuration can be computed. According to the average ranking the $\epsilon$MAg-ES with $\lambda=40$ does also show the best realizations. Hence, the experiment supports the choice of $\lambda=4N$ introduced in~\cite{HellwigB2018}. 
   
    Regarding larger dimensions, the tendency towards medium sized populations increases. While the $\epsilon$MAg-ES configuration with $\lambda=100$ is most successful on the first half of the benchmark problems in $N=100$, on the second half of the problems the distribution of ranks is more divers. Still the constrained functions $C22$ and $C28$ appear to reward very large population sizes.
    In contrast to the situation for $N=10$, this is also observed on problem $C19$ in $N=100$. 
    
    Counting the number of first ranks, with eleven first places the configuration with $\lambda=100$ is best.
    It turns out that the $\epsilon$MAg-ES configuration with population size $\lambda=200$ is more successful when considering the average ranks. That is, it is expected to have improved overall performance on the CEC2017 benchmark suite. However, its advantage over the choice of $\lambda=100$ or the standard recommendation of $\lambda=400$ is rather small.
    The two latter configurations show a similar overall performance on the constrained CEC2017 benchmarks. Consequently, the standard recommendation of $\lambda=4N$ appears to also be a veritable choice in large dimensions. Yet, a tendency towards smaller population sizes can be observed. That is, by making use of slightly smaller populations, an overhead in terms of function evaluations for needless candidate solutions should be reduced and the performance of the $\epsilon$MAg-ES can be further increased.


    \subsection{On the maximal number of consecutively repeated repair steps}
    \label{sec062}   
    
    Considering the constraint handling components of the $\epsilon$MAg-ES, the corresponding strategy parameters also represent standard recommendations~\cite{TakahamaS10}. The {\color{black} Jacobian-based} repair step is applied every $N$ iterations with probability $\theta_p = 0.2$.
    By default, it is repeated at most $\theta_r= 3$ times unless a feasible candidate solution is generated earlier.
    Taking into account the empirical investigations above, the {\color{black} Jacobian-based} repair step turned out to be beneficial on several constrained test functions. Yet, in some cases (e.g. on problem $C20$) repair appeared to prevent approaching better regions of the search space.
    
    By varying the number of repetitions over the set $\theta_r \in \{1,3,10,20,100,1000\}$, we aim at finding a beneficial configuration for the original  $\epsilon$MAg-ES on the constrained CEC2017 benchmarks. To this end, the algorithm realizations after 25 independent runs in each configuration are compared. The resulting ranking is again illustrated by a heatmap displaying the ranks for each individual constrained problem in Fig.~\ref{heatRepN10}.
       \begin{figure}[b]
      \includegraphics[clip,trim= 40 225 50 230, width=0.8\textwidth]{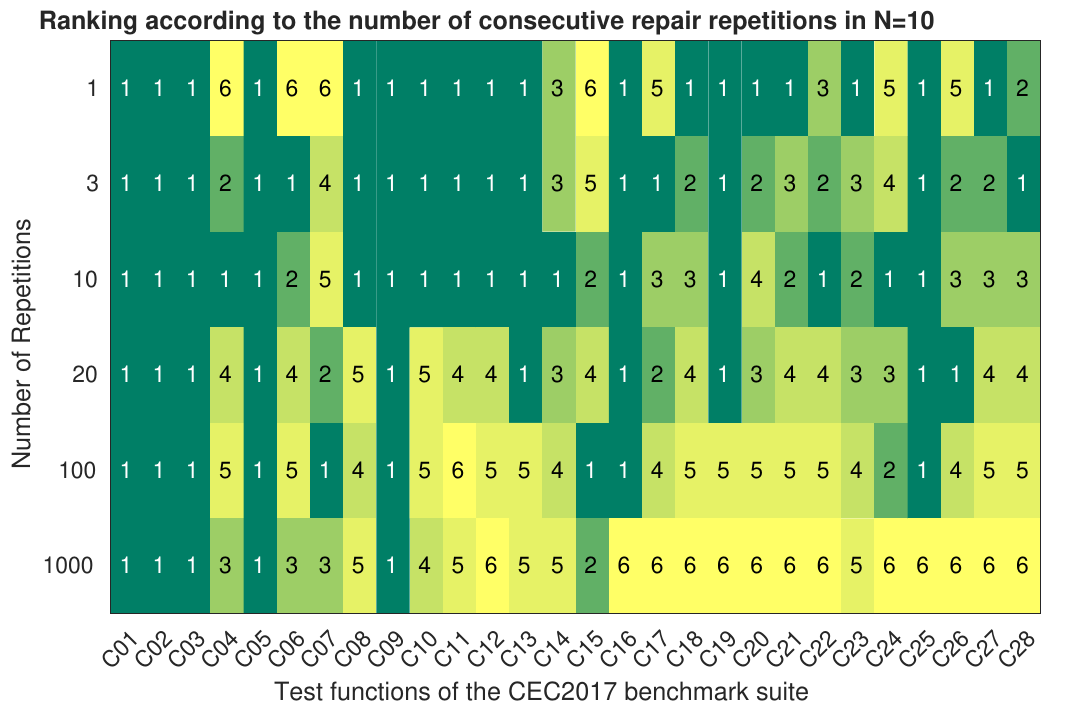}
     \caption{Impact of the number of consecutively repeated repair steps $\theta_r$ on the $\epsilon$MAg-ES performance in $N=10$.}
     \Description{Impact of the maximal number of repeated repair steps $\theta_r$ on the $\epsilon$MAg-ES performance in $N=10$.}
     \label{heatRepN10}
    \end{figure}
    
    A plain trend towards lower $\theta_r$ values can be inferred. This corresponds to fewer consecutive repair operations to be carried out in Alg.~\ref{alg_MAES}. While the $\epsilon$MAg-ES approaches candidate solutions of similar quality regardless of the  $\theta_r$ choice on the first three functions, especially on the second half of the constrained benchmark functions lower $\theta_r$ values are rewarded. Counting only the number of first ranks indicates that the $\epsilon$MAg-ES with $\theta_r=1$ is the most successful configuration. It receives the best rank on 19 out of 28 constrained benchmarks.
    Yet, that respective configuration is rather unsuccessful on the remaining 9 problems. Relating the best overall configuration for {\color{black}the constrained CEC2017 benchmarks} to the average ranking, one identifies $\theta_r=10$ as the most beneficial parameter choice. There do not exist many constrained functions that clearly reward larger $\theta_r$ values in dimension $N=10$. Taking into account Fig.~\ref{heatRepN10}, excessive repair repetitions do not have a positive impact on the algorithm realization even on most of those constrained problems that have only equality constraints, i.e. problems $C06$, $C08$, and $C10$. The only exception is constrained function $C07$, see Eq.~\eqref{C07}.
    
    A similar situation can be seen in higher dimensions. In most cases, the choice of a small number of repetitions seems more appropriate. While the number of repetitions had no influence on the algorithm realization on problems $C01$, $C02$, $C03$, $C05$, and $C09$ in N=10, the ranking is more accentuated in higher dimensions. Yet, only few problems show distinct differences compared to the ranking in low-dimensional functions. These problems include $C01$ and $C06$. 
    
    For $N=100$, the choice of $\theta_r =1$ represents the configuration that receives the largest number of first ranks. The standard configuration of $\theta_r =3$ is observed to be runner-up in this context. However, taking into account the average ranks, the standard $\epsilon$MAg-ES configuration represents the best choice for the whole benchmark suite.
    After all, the standard choice of $\theta_r=3$ appears to be a good compromise on the Constrained CEC2017 benchmark suite in both considered dimensions.

    \begin{figure}[t]
      \includegraphics[clip,trim= 40 225 50 230, width=0.8\textwidth]{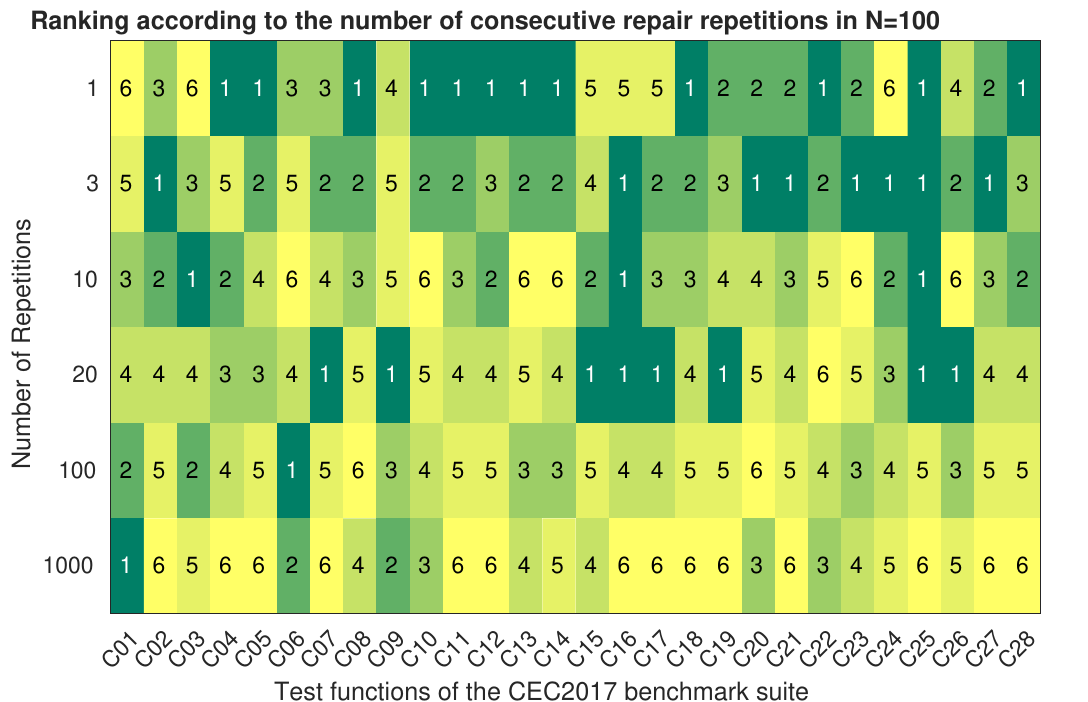}
     \caption{Impact of the number of consecutively repeated repair steps $\theta_r$ on the $\epsilon$MAg-ES performance in $N=100$.}
     \Description{Impact of the maximal number of repeated repair steps $\theta_r$ on the $\epsilon$MAg-ES performance in $N=100$.}
     \label{heatRepN100}
    \end{figure}

\section{Discussion}
\label{disc}

The $\epsilon$MAg-ES working principles have been empirically investigated on the constrained CEC2017 benchmarks in different settings. The results of these examinations will now be classified along the original goal of this paper. In addition to the proposal of suitable ways for thorough algorithm assessment, an improved understanding of the algorithmic working principles has been intended.
The latter includes identifying the contribution of algorithm components to the overall performance of the $\epsilon$MAg-ES as well as characterizing beneficial and/or challenging constrained problem features.

{
Aiming at the sound assessment of novel algorithmic ideas, the contribution of new operators and components (as well as their integration into well-known algorithms) requires to be investigated thoroughly. 
}
This is especially necessary in cases where several new algorithm components are introduced in combination. Such papers often lack elaborate examinations with respect to the usefulness of the single operators. Therefore, the paper proposes to test new developments against reduced versions of themselves. This way, the actual impact of a novel operator can be isolated and might be identified with certain problem characteristics.

Also, more in-depth investigations of the algorithm's mean value dynamics can be considered to cast light on the working principles on specific problems. Of course, it is very cumbersome to compare the dynamics over the whole benchmark suite. To limit this effort to some extend, the present paper only considers those constrained problems on which $\epsilon$MAg-ES has proven to be comparatively successful or unsuccessful in the context of the CEC2017 competition.

For the comparison of the experimental results, we included the non-parametric Wilcoxon Signed Rank into the ranking process.
This hypothesis test estimates whether two sets of algorithm realizations are similarly distributed. It is widely-used to make a decision about the significance of observed differences~\cite{DERRAC20113}. Making use of this hypothesis test within the ranking is expected to avoid rewarding or punishing algorithms for marginal differences. 
Eventually, the inclusion of the non-parametric significance test into the ranking process might not be enough. The distribution of the realizations is estimated based on the median differences. Hence, the decision returned by the Wilcoxon Signed Rank test might not match the observed mean value dynamics due to the impact of single outliers. On the other hand, the use of a parametric hypothesis test that provides a decision for the significance of the mean differences of two realizations is not possible because the distributions of the algorithm realizations do follow any fixed distribution in general.
As a consequence, other measures of effectiveness and robustness could be helpful in this context~\cite{bartz2006experimental}.

Unfortunately, {\color{black}the constrained CEC2017 benchmarks} are not grouped into constrained function classes with similar characteristics. Actually, most function definitions are very special. This complicates the deduction of general statements concerning the usefulness of distinct algorithmic components. Still, some principal findings could be obtained.
As expected, a clear benefit of the matrix adaptation can also be observed on the constrained problems of the CEC2017 benchmark suite.
Further, the combination of $\epsilon$-level constraint handling and {\color{black} Jacobian-based} repair appears to be beneficial in the presence of equality constraints. In this situation, the limitation of the mutation strength can be regarded usefull especially in small dimensions.
In high dimensions, the {\color{black} Jacobian-based} repair as well as the back-calculations cause disadvantages on individual functions. Unfortunately, it is very difficult to identify these drawbacks with associated problem features. In some of these cases, the repair approach is simply slowing down the algorithm's progress towards candidate solutions of similar quality.

The use of \emph{back-calculation} also appears particularly useful in smaller dimensions. In the presence of equality constraints, it can be regarded an indispensable algorithm component. However, for the reasons already mentioned it is difficult to derive general statements for large dimensions.

On problems where the lexicographic variants show improvements over the $\epsilon$MAg-ES performance, a faster reduction of the $\epsilon$-threshold within $\epsilon$MAg-ES would be beneficial.

An examination of beneficial strategy parameter choices was conducted in Sec.~\ref{sec06}. It revealed good performance results with respect to the standard settings. Yet, especially in larger dimensions the population size of $\lambda=4N$ which was inspired by the standard population size recommendations for Differential Evolution seems to be at the upper limit. However, our experimental analysis supported the original design decision~\cite{HellwigB2018} to let go the standard CMA-ES population size recommendations for the $\epsilon$MAg-ES in the constrained setting of {\color{black}the constrained CEC2017 benchmarks}. In fact, population sizes of about $\lambda=2N$ yields the best performance over the complete benchmark suite.

The design of the original $\epsilon$MAg-ES~\cite{HellwigB2018} represents an appropriate combination of the individual algorithm components for the complete CEC2017 benchmark suite. Particularly for low dimensional problems, the original $\epsilon$MAg-ES turned out to be the right combination of components. However, the rankings among the algorithm components are no longer so clearly distributed in higher dimensions. This might indicate additional potential for improvement in high dimensions, where $\epsilon$MAg-ES has already proven to be comparatively competitive in the CEC2018 competition. 
The performance on individual constrained problems can still be improved. To this end, the algorithm needs specific tuning on each problem. Another option would be to integrate some kind of mechanism that can identify certain problem features and chooses the algorithmic components and strategy parameters accordingly.

%% file: telo-appendix.tex
\renewcommand\thesubsection{\Alph{subsection}}
\pagenumbering{gobble}
\subsection{The impact of significance testing on the algorithm ranking}
\label{AppendixA}
 {
 This section illustrates the changes in the comparative results of the seven $\epsilon$MAg-ES variants 
 caused by the inclusion of the Wilcoxon Signed Rank test into the mean and median ranking procedure. 
 The related steps are described in detail in Sec.~\ref{sec04}.
 
 One observes that the number of draws between algorithms is increasing considerably when using the Wilcoxon Signed Rank test for identification of statistically significant differences among the algorithm realizations. That is, the algorithmic variants must be regarded to be producing similarly distributed output values on many constrained problems of the CEC2017 benchmarks.
 
 While the number of ties rises, the general ranking tendency is preserved. Hence, the use of the Wilcoxon Signed Rank Test is not producing a completely new picture. By removing insignificant differences it is only changing the contrast sharpness of the ranking results. This might be caused by the only marginally different algorithm variants that are compared in the presented investigation. At best, the contrast will be further sharpened by making use of the hypothesis test as it is able to filter out pointlessly small differences among the algorithm outputs.
}
 \begin{table}[htbp]
\centering
\caption{Comparison results according to the mean, median and total ranking as recommended for CEC2018~\cite{CEC2017}, i.e. \emph{without} making use of statistical hypothesis testing.}
\label{tabA1}
\renewcommand{\arraystretch}{1.2}
\resizebox{0.7\textwidth}{!}{%
\begin{tabular}{lccgccg}
  $\bm{\epsilon}$\textbf{MAg-ES}          & \multicolumn{3}{c}{$\bm{N=10}$} & \multicolumn{3}{c}{$\bm{N=100}$} \\
            & \multicolumn{3}{c}{Ranking} & \multicolumn{3}{c}{Ranking} \\
 $\bm{+}$/$\bm{=}$/$\bm{-}$      & Median & Mean & Total & Median & Mean   & Total\\\hline
$\epsilon$MA-ES       & 11/13/4    & 12/8/8     & 11/11/6           & 8/2/18  & 11/1/16 & 8/6/14\\
$\epsilon$MAg-ES w/o  & 14/9/5    & 21/0/7     & 19/5/4          & 15/0/13  & 15/0/13 & 13/4/11\\
$\epsilon$MAg-ES nl   & 15/9/4   & 20/5/3    & 18/7/3         & 11/2/15  & 12/1/15 & 9/6/13\\
$\epsilon$SAg-ES      & 23/3/2    & 23/2/3     & 22/5/1          & 21/0/7   & 22/0/6 & 20/3/5\\
$lex$MAg-ES      & 7/13/8    & 15/5/8          & 11/9/8          & 15/0/13  & 14/0/14 & 13/3/12\\
$lex$MA-ES       & 9/11/8    & 17/3/8          & 13/7/8          & 16/0/12  & 17/0/11 & 15/3/10\\\hline
\end{tabular}%
}
\end{table}

\begin{table}[htbp]
\centering
\caption{Comparison results with respect to the significance test based ranking, cf. Table~\ref{tab1} in Sec.~\ref{sec04}.}
\label{tabA2}
\renewcommand{\arraystretch}{1.2}
\resizebox{0.7\textwidth}{!}{%
\begin{tabular}{lccgccg}
  $\bm{\epsilon}$\textbf{MAg-ES}          & \multicolumn{3}{c}{$\bm{N=10}$} & \multicolumn{3}{c}{$\bm{N=100}$} \\
            & \multicolumn{3}{c}{Ranking} & \multicolumn{3}{c}{Ranking} \\
 $\bm{+}$/$\bm{=}$/$\bm{-}$      & Median & Mean & Total & Median & Mean   & Total\\\hline
$\epsilon$MA-ES       & 7/19/2    & 7/17/4     & 7/18/3           & 5/12/11  & 7/11/10 & 5/13/10\\
$\epsilon$MAg-ES w/o  & 8/17/2    & 10/15/3     & 10/15/3          & 10/8/10  & 13/6/9 & 11/8/9\\
$\epsilon$MAg-ES nl   & 12/14/2   & 14/12/2    & 14/12/2         & 5/15/8   & 7/14/7 & 5/16/7\\
$\epsilon$SAg-ES      & 18/9/1    & 20/7/1     & 20/7/1          & 18/8/2   & 18/8/2 & 17/10/1\\
$lex$MAg-ES      & 6/20/2    & 7/18/3          & 7/18/3          & 9/10/9   & 10/9/9 & 9/11/8\\
$lex$MA-ES       & 8/16/4    & 8/14/6          & 8/14/6          & 14/7/7   & 14/7/7 & 13/9/6\\\hline
\end{tabular}%
}
\end{table}

  \clearpage 
    
\subsection{Detailed comparison of the algorithm variants}
        \label{AppendixC}

The detailed comparison results corresponding to Table~\ref{tab1} are displayed in Table~\ref{tab:my-table1} and Table~\ref{tab:my-table2} for dimensions $N=10$ and $N=100$, respectively. 
For each constrained function in the benchmark set, the tables indicate whether neglecting a specific component does affect the algorithm performance. Accordingly, ($+$) corresponds to a positive contribution of the neglected component within the $\epsilon$MAg-ES, and ($-$) to a negative contribution, respectively.
The equal sign ($=$) implies that the component has no significant effect on the algorithm effectiveness on that particular problem. The aggregated results are displayed in the bottom row of the tables by using the notation ``+/=/-'' representing the sum of positive, neutral and negative observations.

\begin{table}[b]
\centering
\caption{Results of the comparison of $\epsilon$MAg-ES with six different reduced algorithm variants of itself on the constrained CEC2017 benchmark functions in dimension $N=10$. The highlighted rows correspond to benchmark problems on which $\epsilon$MAg-ES performed particularly well (light gray) or bad (dark gray), respectively. The algorithm behavior on these problems is more closely examined in Sec.~\ref{sec05}.}
\label{tab:my-table1}
\renewcommand{\arraystretch}{1.1}
\resizebox{0.95\textwidth}{!}{%
\begin{tabular}{l|cccccc|cccccc}
\multicolumn{1}{l|}{\textbf{N =10}} & \multicolumn{6}{c|}{\textbf{Median Performance}} & \multicolumn{6}{c}{\textbf{Mean Performance}} \\[0ex]
\multicolumn{1}{l|}{}               & \multicolumn{6}{c|}{$\epsilon$MAg-ES vs.}              & \multicolumn{6}{c}{epsMAg vs.}            \\[1ex]
\multirow{ 2}{*}{Problem}        & \multirow{ 2}{*}{$\epsilon$MA}     &  $\epsilon$MAg  & $\epsilon$MAg    & \multirow{ 2}{*}{$\epsilon$SAg}  & \multirow{ 2}{*}{lexMAg}   & \multirow{ 2}{*}{lexMA}    & \multirow{ 2}{*}{$\epsilon$MA}    & $\epsilon$MAg  & $\epsilon$MAg & \multirow{ 2}{*}{$\epsilon$SAg}  & \multirow{ 2}{*}{lexMAg }  & \multirow{ 2}{*}{lexMA }   \\[-1ex] 
           & &  w/o & nl & & & & & w/o & nl & & & \\ \hline 
C01 & $=$& $=$& $=$& $+$& $=$& $=$& $=$& $=$& $=$& $+$& $=$& $=$\\
C02 & $=$& $=$& $=$& $+$& $=$& $=$& $=$& $=$& $=$& $+$& $=$& $=$\\\rowcolor{gray!25}
C03 & \multicolumn{1}{|c}{$=$}& $=$& $=$& $+$& $+$& $+$& $=$& $=$& $=$& $+$& $+$& $+$\\\rowcolor{gray!60}
C04 & \multicolumn{1}{|c}{$-$}& $-$& $=$& $=$& $+$& $+$& $-$& $-$& $=$& $=$& $+$& $+$\\
C05 & $=$& $=$& $=$& $+$& $=$& $=$& $=$& $=$& $=$& $+$& $=$& $=$\\
C06 & $+$& $+$& $+$& $+$& $=$& $+$& $+$& $+$& $+$& $+$& $+$& $+$\\\rowcolor{gray!25}
C07 & \multicolumn{1}{|c}{$+$}& $+$& $+$& $+$& $+$& $+$& $+$& $+$& $+$& $+$& $+$& $+$\\
C08 & $=$& $=$& $=$& $+$& $=$& $=$& $=$& $=$& $=$& $+$& $=$& $=$\\
C09 & $=$& $+$& $+$& $+$& $=$& $=$& $=$& $+$& $+$& $+$& $=$& $=$\\
C10 & $=$& $+$& $=$& $+$& $=$& $=$& $=$& $+$& $=$& $+$& $=$& $=$\\
C11 & $+$& $+$& $+$& $+$& $=$& $=$& $+$& $+$& $+$& $+$& $=$& $=$\\
C12 & $=$& $=$& $+$& $=$& $=$& $=$& $=$& $=$& $+$& $=$& $=$& $=$\\
C13 & $=$& $=$& $=$& $+$& $=$& $=$& $=$& $=$& $=$& $+$& $=$& $=$\\
C14 & $+$& $+$& $+$& $+$& $-$& $-$& $+$& $+$& $+$& $+$& $-$& $-$\\
C15 & $=$& $=$& $=$& $=$& $-$& $-$& $+$& $+$& $+$& $+$& $-$& $-$\\
C16 & $=$& $=$& $=$& $=$& $+$& $+$& $=$& $=$& $=$& $=$& $+$& $+$\\
C17 & $=$& $=$& $=$& $=$& $=$& $=$& $=$& $=$& $=$& $=$& $=$& $=$\\
C18 & $+$& $+$& $+$& $+$& $=$& $=$& $-$& $+$& $+$& $+$& $=$& $=$\\
C19 & $=$& $=$& $+$& $+$& $=$& $=$& $=$& $=$& $+$& $+$& $=$& $=$\\\rowcolor{gray!60}
C20 & \multicolumn{1}{|c}{$-$}& $-$& $-$& $-$& $=$& $-$& $-$& $-$& $-$& $-$& $=$& $-$\\\rowcolor{gray!60}
C21 & \multicolumn{1}{|c}{$=$}& $=$& $=$& $=$& $=$& $=$& $=$& $=$& $=$& $=$& $=$& $-$\\
C22 & $=$& $=$& $-$& $+$& $=$& $=$& $=$& $=$& $-$& $+$& $=$& $=$\\
C23 & $=$& $=$& $+$& $=$& $=$& $=$& $-$& $-$& $+$& $+$& $-$& $-$\\
C24 & $+$& $+$& $=$& $+$& $+$& $+$& $+$& $+$& $+$& $+$& $+$& $+$\\
C25 & $=$& $=$& $=$& $=$& $+$& $+$& $=$& $=$& $=$& $=$& $+$& $+$\\
C26 & $=$& $=$& $+$& $=$& $=$& $=$& $=$& $=$& $+$& $=$& $=$& $=$\\\rowcolor{gray!25}
C27 & \multicolumn{1}{|c}{$+$}& $+$& $+$& $+$& $=$& $+$& $+$& $+$& $+$& $+$& $=$& $+$\\
C28 & $=$& $=$& $+$& $+$& $=$& $-$& $=$& $=$& $+$& $+$& $=$& -1\\\hline
\bf +/=/-         & 7/19/2    & 8/17/2 & 12/14/2 & 18/9/1   & 6/20/2   & 8/16/4  
& 7/17/4   & 10/15/3 & 14/12/2 & 20/7/1   & 7/18/3  & 8/14/6 
\end{tabular}%
}
\end{table}
    
    
\begin{table}[t]
\centering
\caption{Results of the comparison of $\epsilon$MAg-ES with six different reduced algorithm variants of itself on the constrained CEC2017 benchmark functions in dimension $N=100$. The highlighted rows correspond to benchmark problems on which $\epsilon$MAg-ES performed particularly well (light gray) or bad (dark gray), respectively. As for $N=10$, the algorithm behavior on these problems is more closely examined in Sec.~\ref{sec05}.}
\label{tab:my-table2}
\renewcommand{\arraystretch}{1.1}
\resizebox{0.95\textwidth}{!}{%
\begin{tabular}{l|cccccc|cccccc}
\multicolumn{1}{l|}{\textbf{N =100}} & \multicolumn{6}{c|}{\textbf{Median Performance}} & \multicolumn{6}{c}{\textbf{Mean Performance}} \\[0ex]
\multicolumn{1}{l|}{}               & \multicolumn{6}{c|}{$\epsilon$MAg-ES vs.}              & \multicolumn{6}{c}{epsMAg vs.}            \\[1ex]
\multirow{ 2}{*}{Problem}        & \multirow{ 2}{*}{$\epsilon$MA}     &  $\epsilon$MAg  & $\epsilon$MAg    & \multirow{ 2}{*}{$\epsilon$SAg}  & \multirow{ 2}{*}{lexMAg}   & \multirow{ 2}{*}{lexMA}    & \multirow{ 2}{*}{$\epsilon$MA}    & $\epsilon$MAg  & $\epsilon$MAg & \multirow{ 2}{*}{$\epsilon$SAg}  & \multirow{ 2}{*}{lexMAg }  & \multirow{ 2}{*}{lexMA }   \\[-1ex] 
           & &  w/o & nl & & & & & w/o & nl & & & \\ \hline 
C01 & $=$& $+$& $=$& $+$& $=$& $=$& $=$& $+$& $=$& $+$& $=$& $=$\\
C02 & $=$& $+$& $=$ & $+$& $=$ & $=$ & $=$ & $+$& $=$ & $+$& $=$ & 0 \\\rowcolor{gray!25}
C03 & \multicolumn{1}{|c}{$=$} & $+$& $=$ & $+$& $+$& $+$& $=$ & $+$& $=$ & $+$& $+$& $+$\\\rowcolor{gray!60}
C04 & \multicolumn{1}{|c}{$-$}& $-$& $=$ & $=$ & $+$& $+$& $-$& $+$& $=$ & $=$ & $+$& $+$\\
C05 & $-$& $+$& $-$& $+$& $-$& $=$ & $-$& $+$& $-$& $+$& $-$& 0 \\
C06 & $+$& $+$& $=$ & $+$& $+$& $+$& $+$& $+$& $=$ & $+$& $+$& $+$\\\rowcolor{gray!25}
C07 & \multicolumn{1}{|c}{$=$} & $=$ & $=$ & $=$ & $+$& $+$& $=$ & $=$ & $=$ & $=$ & $+$& $+$\\
C08 & $-$& $=$ & $=$ & $+$& $-$& $-$& $-$& $=$ & $=$ & $+$& $-$& $-$\\
C09 & $-$& $=$ & $+$& $+$& $-$& $-$& $-$& $-$& $+$& $-$& $-$& $-$\\
C10 & $-$& $-$& $-$& $+$& $-$& $-$& $-$& $-$& $-$& $+$& $-$& $-$\\
C11 & $=$ & $=$ & $=$ & $+$& $-$& $-$& $-$& $+$& $-$& $+$& $-$& $-$\\
C12 & $-$& $-$& $-$& $+$& $-$& $-$& $+$& $+$& $+$& $+$& $+$& $+$\\
C13 & $-$& $=$ & $-$& $+$& $+$& $+$& $-$& $=$ & $-$& $+$& $+$& $+$\\
C14 & $=$ & $-$& $+$& $=$ & $=$ & $+$& $=$ & $-$& $+$& $=$ & $=$ & $+$\\
C15 & $+$& $+$& $+$& $+$& $+$& $+$& $+$& $+$& $+$& $+$& $-$& $-$\\
C16 & $=$ & $+$& $=$ & $+$& $+$& $+$& $=$ & $+$& $=$ & $+$& $+$& $+$\\
C17 & $-$& $-$& $-$& $=$ & $=$ & $+$& $+$& $-$& $-$& $=$ & $=$ & $+$\\
C18 & $=$ & $=$ & $=$ & $+$& $=$ & $=$ & $=$ & $=$ & $=$ & $+$& $=$ & 0 \\
C19 & $=$ & $-$& $=$ & $=$ & $=$ & $=$ & $=$ & $-$& $=$ & $=$ & $=$ & 0 \\\rowcolor{gray!60}
C20 & \multicolumn{1}{|c}{$-$}& $-$& $-$& $-$& $-$& $-$& $-$& $-$& $-$& $-$& $-$& $-$\\\rowcolor{gray!60}
C21 & \multicolumn{1}{|c}{$=$} & $+$& $=$ & $+$& $=$ & $=$ & $=$ & $+$& $=$ & $+$& $=$ & 0 \\
C22 & $-$& $-$& $-$& $-$& $-$& $-$& $-$& $-$& $+$& $+$& $-$& $-$\\
C23 & $+$& $=$ & $+$& $=$ & $=$ & $+$& $+$& $=$ & $+$& $=$ & $=$ & $+$\\
C24 & $+$& $+$& $+$& $+$& $+$& $+$& $+$& $+$& $+$& $+$& $+$& $+$\\
C25 & $=$ & $+$& $=$ & $+$& $+$& $+$& $=$ & $+$& $=$ & $+$& $+$& $+$\\
C26 & $+$& $-$& $-$& $=$ & $-$& $+$& $+$& $-$& $-$& $=$ & $-$& $+$\\\rowcolor{gray!25}
C27 & \multicolumn{1}{|c}{$-$}& $=$ & $=$ & $+$& $=$ & $+$& $-$& $=$ & $=$ & $+$& $+$& $+$\\
C28 & $=$ & $-$& $=$ & $=$ & $=$ & $=$ & $=$ & $-$& $=$ & $=$ & 0 & 0\\\hline
\bf +/=/-         & 5/12/11    & 10/8/10 & 5/15/8 & 18/8/2   & 9/10/9   & 14/7/7  
& 7/11/10   & 13/6/9 & 7/14/7 & 18/8/2   & 10/9/9  & 14/7/7 
\end{tabular}%
}
\end{table}

\FloatBarrier
\clearpage
\subsection{Results on the CEC2017 constrained benchmark functions}
\label{AppendixB}
    This sections provides the detailed results of all $\epsilon$MAg-ES variants with respect to the test functions specified in the context of the IEEE CEC 2017 competition on constrained single objective real-parameter optimization~\cite{CEC2017}.    
    {
    All results have been recalculated using Matlab (2018b) on a Intel Haswell Desktop PC with Intel Core i7-4770 3.40GHz\texttimes8 CPU and 16 GB RAM.
    They confirm the results published in the competition except for expected stochastically justified errors. 
    
    In addition to the performance data required by the CEC2017 specifications, Tables~\ref{CEC18_MAESD10} and~\ref{CEC18_MAESD100} comprise the mean running time in terms of function evaluations that have been consumed until the finally returned best observed candidate solution was found.
    This information can be used to distinguish the efficiency of the $\epsilon$MAg-ES algorithm.
    }
    
    \begin{table}[b]
 \centering
 \caption{Final $\epsilon$MAg-ES results of 25 independent runs at a budget of $2 \times 10^5$ function evaluations in $N=10$.}
 \renewcommand{\arraystretch}{0.9}
\resizebox{.9\textwidth}{!}{%
\begin{tabular}{lccccccc}\hline
\textbf{} & \textbf{C01} & \textbf{C02} & \textbf{C03} & \textbf{C04} & \textbf{C05} & \textbf{C06} & \textbf{C07} \\\hline
Best & 0.00000e+00 & 0.00000e+00 & 0.00000e+00 & 0.00000e+00 & 0.00000e+00 & 0.00000e+00 & -3.96309e+02 \\
Median & 0.00000e+00 & 0.00000e+00 & 0.00000e+00 & 3.58184e+01 & 0.00000e+00 & 0.00000e+00 & -3.25196e+02 \\
c & (0,0,0) & (0,0,0) & (0,0,0) & (0,0,0) & (0,0,0) & (0,0,0) & (0,0,0) \\
$\overline{\nu}$ & 0 & 0 & 0 & 0 & 0 & 0 & 0 \\
Mean & 0.00000e+00 & 0.00000e+00 & 5.67980e-31 & 4.65723e+01 & 0.00000e+00 & 1.11013e+01 & -3.17618e+02 \\
Std & 0.00000e+00 & 0.00000e+00 & 2.83990e-30 & 4.94502e+01 & 0.00000e+00 & 2.60937e+01 & 5.55184e+01 \\
Worst & 0.00000e+00 & 0.00000e+00 & 1.41995e-29 & 2.67635e+02 & 0.00000e+00 & 7.87983e+01 & -1.90165e+02 \\
FR & 100 & 100 & 100 & 100 & 100 & 100 & 100 \\
$\overline{vio}$ & 0 & 0 & 0 & 0 & 0 & 0 & 0 \\
\textit{mRTgb} & \textit{1.61480e+04} & \textit{1.60008e+04} & \textit{1.66999e+04} & \textit{3.71440e+04} & \textit{2.07134e+04} & \textit{9.03061e+04} & \textit{7.06426e+04} \\\hline
\textbf{} & \textbf{C08} & \textbf{C09} & \textbf{C10} & \textbf{C11} & \textbf{C12} & \textbf{C13} & \textbf{C14} \\\hline
Best & -1.34840e-03 & -4.97525e-03 & -5.09647e-04 & -1.68819e-01 & 3.98790e+00 & 0.00000e+00 & 2.37633e+00 \\
Median & -1.34840e-03 & -4.97525e-03 & -5.09647e-04 & -1.68819e-01 & 3.98790e+00 & 0.00000e+00 & 2.37633e+00 \\
c & (0,0,0) & (0,0,0) & (0,0,0) & (0,0,0) & (0,0,0) & (0,0,0) & (0,0,0) \\
$\overline{\nu}$ & 0 & 0 & 0 & 0 & 0 & 0 & 0 \\
Mean & -1.34840e-03 & -4.97525e-03 & -5.09647e-04 & -1.67794e-01 & 1.07586e+01 & 0.00000e+00 & 2.60340e+00 \\
Std & 0.00000e+00 & 0.00000e+00 & 1.56212e-15 & 5.12824e-03 & 8.71661e+00 & 0.00000e+00 & 4.30338e-01 \\
Worst & -1.34840e-03 & -4.97525e-03 & -5.09647e-04 & -1.43178e-01 & 2.27853e+01 & 0.00000e+00 & 3.23817e+00 \\
FR & 100 & 100 & 100 & 100 & 100 & 100 & 84 \\
$\overline{vio}$ & 0 & 0 & 0 & 0 & 0 & 0 & 1.10031e-01 \\
\textit{mRTgb} & \textit{7.91829e+04} & \textit{8.22368e+04} & \textit{7.96384e+04} & \textit{1.27640e+05} & \textit{1.04593e+05} & \textit{3.05779e+04} & \textit{1.35929e+05} \\\hline
\textbf{} & \textbf{C15} & \textbf{C16} & \textbf{C17} & \textbf{C18} & \textbf{C19} & \textbf{C20} & \textbf{C21} \\\hline
Best & 2.35619e+00 & 0.00000e+00 & 8.35706e-01 & 3.65977e+01 & 0.00000e+00 & 3.22510e-01 & 3.98790e+00 \\
Median & 2.33418e+00 & 0.00000e+00 & 9.21244e-01 & 3.65977e+01 & 0.00000e+00 & 1.36690e+00 & 3.98790e+00 \\
c & (0,1,0) & (0,0,0) & (1,0,0) & (0,0,0) & (1,0,0) & (0,0,0) & (0,0,0) \\
$\overline{\nu}$ & 1.55638e-02 & 0  & 5.50000e+00 & 0  & 6.63359e+03 & 0  & 0  \\
Mean & 8.00800e+00 & 0.00000e+00 & 7.51421e-01 & 3.66228e+01 & 1.25524e+00 & 1.24122e+00 & 9.19458e+00 \\
Std & 6.14078e+00 & 0.00000e+00 & 3.37835e-01 & 1.24187e-01 & 2.48714e+00 & 3.84294e-01 & 1.07983e+01 \\
Worst & 8.40414e+00 & 0.00000e+00 & 1.00687e+00 & 3.72189e+01 & 6.67172e+00 & 1.58379e+00 & 3.96544e+01 \\
FR & 28 & 100 & 0 & 100 & 0 & 100 & 100 \\
$\overline{vio}$ & 2.87505e-02 & 0 & 5.46406e+00 & 0 & 6.63544e+03 & 0 & 0 \\
\textit{mRTgb} & \textit{4.27572e+03} & \textit{1.50758e+04} & \textit{1.28030e+05} & \textit{1.34718e+05} & \textit{5.12316e+04} & \textit{1.00448e+05} & \textit{1.15331e+05} \\\hline
\textbf{} & \textbf{C22} & \textbf{C23} & \textbf{C24} & \textbf{C25} & \textbf{C26} & \textbf{C27} & \textbf{C28} \\\hline
Best & 3.46248e-27 & 2.37633e+00 & 2.35612e+00 & 0.00000e+00 & 1.15234e+00 & 3.65977e+01 & 0.00000e+00 \\
Median & 3.95606e-27 & 2.37633e+00 & 2.35619e+00 & 0.00000e+00 & 9.33222e-01 & 3.65977e+01 & 1.27767e+01 \\
c & (0,0,0) & (0,0,0) & (0,0,0) & (0,0,0) & (1,0,0) & (0,0,0) & (1,0,0) \\
$\overline{\nu}$ & 0  & 0  & 0  & 0  & 5.50000e+00 & 0  & 6.64271e+03 \\
Mean & 1.59463e-01 & 2.82990e+00 & 4.36680e+00 & 0.00000e+00 & 7.13006e-01 & 5.43499e+01 & 7.40651e+00 \\
Std & 7.97316e-01 & 6.73197e-01 & 2.54578e+00 & 0.00000e+00 & 3.76973e-01 & 6.31146e+01 & 6.25520e+00 \\
Worst & 3.98658e+00 & 3.19016e+00 & 8.63937e+00 & 0.00000e+00 & 9.82088e-01 & 3.08500e+02 & 1.58320e+01 \\
FR & 100 & 72 & 100 & 100 & 0 & 100 & 0 \\
$\overline{vio}$ & 0 & 2.43341e-01 & 0 & 0 & 5.40731e+00 & 0 & 6.63975e+03 \\
\textit{mRTgb} & \textit{9.32575e+04} & \textit{1.17856e+05} & \textit{1.03273e+05} & \textit{1.59107e+04} & \textit{8.76907e+04} & \textit{1.46554e+05} & \textit{1.08451e+05}      \\\hline           
\end{tabular}%
}
\label{CEC18_MAESD10}
\end{table}

    \begin{table}[t]
 \centering
 \caption{Final $\epsilon$MAg-ES results of 25 independent runs at a budget of $2 \times 10^6$ function evaluations in $N=100$.}
 \renewcommand{\arraystretch}{0.9}
\resizebox{0.9\textwidth}{!}{%
\begin{tabular}{lccccccc}\hline
\textbf{} & \textbf{C01} & \textbf{C02} & \textbf{C03} & \textbf{C04} & \textbf{C05} & \textbf{C06} & \textbf{C07} \\\hline
Best & 3.10360e-26 & 2.81474e-26 & 3.39580e-26 & 1.93021e+02 & 2.59752e+01 & 1.95966e+02 & -3.35642e+03 \\
Median & 4.14232e-26 & 3.83181e-26 & 4.20179e-26 & 2.53713e+02 & 2.69755e+01 & 1.10410e+03 & -2.79273e+03 \\
c & (0,0,0) & (0,0,0) & (0,0,0) & (0,0,0) & (0,0,0) & (0,0,0) & (0,0,0) \\
$\overline{\nu}$ & 0 & 0 & 0 & 0 & 0 & 0 & 0 \\
Mean & 4.13000e-26 & 3.92321e-26 & 4.21586e-26 & 2.53355e+02 & 2.69935e+01 & 9.02181e+02 & -2.51197e+03 \\
Std & 4.87701e-27 & 5.70146e-27 & 5.18128e-27 & 2.97793e+01 & 5.77895e-01 & 4.17264e+02 & 8.29608e+02 \\
Worst & 5.08859e-26 & 4.95259e-26 & 5.27463e-26 & 3.04456e+02 & 2.81042e+01 & 1.36320e+03 & -7.01511e+02 \\
FR & 100 & 100 & 100 & 100 & 100 & 100 & 100 \\
$\overline{vio}$ & 0 & 0 & 0 & 0 & 0 & 0 & 0 \\
\textit{mRTgb} & \textit{1.21246e+06} & \textit{1.25758e+06} & \textit{1.36123e+06} & \textit{3.92639e+05} & \textit{1.99822e+06} & \textit{1.73964e+06} & \textit{1.05396e+06} \\\hline
\textbf{} & \textbf{C08} & \textbf{C09} & \textbf{C10} & \textbf{C11} & \textbf{C12} & \textbf{C13} & \textbf{C14} \\\hline
Best & -4.76002e-05     & 1.34102e-04 & -7.58732e-06 & -5.73017e+00 & 1.88577e+01 & 4.01395e+01 & 9.29361e-01 \\
Median & -4.53776e-05   & 2.01341e+00 & -9.50623e-07 & -5.72792e+00 & 3.15768e+01 & 4.09403e+01 & 9.66522e-01 \\
c & (0,0,0) & (0,0,0)   & (0,0,0)        & (0,0,0) & (0,0,0) & (0,0,0) & (0,0,0) \\
$\overline{\nu}$ & 0    & 0              & 0 & 0 & 0 & 0 & 0 \\
Mean & -4.53505e-05     & 3.61141e+00 & -1.15694e-06 & -5.56414e+00 & 3.00505e+01 & 4.13271e+01 & 9.69503e-01 \\
Std & 1.42742e-06       & 3.94819e+00 & 2.53082e-06 & 5.74166e-01 & 4.21847e+00 & 1.03009e+00 & 2.18962e-02 \\
Worst & -4.19790e-05    & 9.22061e+00 & 3.08926e-06 & -5.10189e+00 & 3.15768e+01 & 4.39995e+01 & 1.01813e+00 \\
FR & 100                & 92            & 100     & 100 & 100 & 100 & 100 \\
$\overline{vio}$ & 0    & 3.36830e-04 & 0 & 0 & 0 & 0 & 0 \\
\textit{mRTgb} & \textit{1.99800e+06} & \textit{1.89356e+06} & \textit{1.99746e+06} & \textit{1.97715e+06} & \textit{1.97440e+06} & \textit{1.99880e+06} & \textit{1.51669e+06} \\\hline
\textbf{} & \textbf{C15} & \textbf{C16} & \textbf{C17} & \textbf{C18} & \textbf{C19} & \textbf{C20} & \textbf{C21} \\\hline
Best & 2.35612e+00 & 0.00000e+00 & 1.09220e+00 & 3.63770e+01 & 2.52948e+01 & 3.16900e+01 & 3.15768e+01 \\
Median & 8.63938e+00 & 0.00000e+00 & 1.09687e+00 & 3.63770e+01 & 7.65363e+01 & 3.55191e+01 & 3.15768e+01 \\
c & (0,0,0) & (0,0,0) & (1,0,0) & (0,0,0) & (1,0,0) & (0,0,0) & (0,0,0) \\
$\overline{\nu}$ & 0 & 0 & 5.05000e+01 & 0 & 7.30489e+04 & 0 & 0 \\
Mean & 1.02729e+01 & 0.00000e+00 & 1.09366e+00 & 3.63770e+01 & 7.50169e+01 & 3.54012e+01 & 3.15768e+01 \\
Std & 5.21916e+00 & 0.00000e+00 & 2.58251e-03 & 3.12396e-06 & 4.98116e+01 & 1.06111e+00 & 5.93952e-14 \\
Worst & 1.49211e+01 & 0.00000e+00 & 1.09375e+00 & 3.63770e+01 & 2.42357e+02 & 3.69680e+01 & 3.15768e+01 \\
FR & 96 & 100 & 0 & 100 & 0 & 100 & 100 \\
$\overline{vio}$ & 4.02523e-05 & 0 & 5.05000e+01 & 0 & 7.30546e+04 & 0 & 0 \\
\textit{mRTgb} & \textit{1.08402e+06} & \textit{7.91872e+05} & \textit{1.71046e+06} & \textit{1.99394e+06} & \textit{1.99765e+06} & \textit{1.01919e+06} & \textit{1.76169e+06} \\\hline
\textbf{} & \textbf{C22} & \textbf{C23} & \textbf{C24} & \textbf{C25} & \textbf{C26} & \textbf{C27} & \textbf{C28} \\\hline
Best & 5.96452e+01 & 9.23985e-01        & 2.35612e+00 & 0.00000e+00 & 1.09645e+00 & 3.63770e+01 & 1.13360e+02 \\
Median & 3.81015e+03 & 9.61070e-01      & 1.17810e+01 & 0.00000e+00 & 1.09492e+00 & 3.63770e+01 & 1.61397e+02 \\
c & (0,0,0) & (0,0,0) & (0,0,0)         & (0,0,0)       & (1,0,0) & (0,0,0) & (1,0,0) \\
$\overline{\nu}$ & 0 & 0                & 0            & 0 & 5.05000e+01 & 0 & 7.31192e+04 \\
Mean & 3.73613e+03 & 9.67071e-01        & 1.01473e+01 & 2.28092e-14 & 1.09440e+00 & 3.63770e+01 & 1.65114e+02 \\
Std & 1.54660e+03 & 2.34153e-02         & 5.52541e+00 & 9.60814e-14 & 1.65230e-03 & 3.42842e-06 & 2.43884e+01 \\
Worst & 6.41242e+03 & 1.00797e+00       & 1.80641e+00 & 4.74832e-13 & 1.09646e+00 & 3.63770e+01 & 2.15224e+02 \\
FR & 100 & 100                          & 100 & 100 & 0 & 100 & 0 \\
$\overline{vio}$ & 0 & 0                & 0 & 0 & 5.05000e+01 & 0 & 7.31146e+04 \\
\textit{mRTgb} & \textit{1.95883e+06} & \textit{1.50016e+06} & \textit{1.17679e+06} & \textit{1.25249e+06} & \textit{1.66838e+06} & \textit{1.99549e+06} & \textit{1.89616e+06}\\\hline
\end{tabular}%
}
\label{CEC18_MAESD100}
\end{table}
\FloatBarrier


%% file: telo-main.bbl

\begin{thebibliography}{16}


\ifx \showCODEN    \undefined \def \showCODEN     #1{\unskip}     \fi
\ifx \showDOI      \undefined \def \showDOI       #1{#1}\fi
\ifx \showISBNx    \undefined \def \showISBNx     #1{\unskip}     \fi
\ifx \showISBNxiii \undefined \def \showISBNxiii  #1{\unskip}     \fi
\ifx \showISSN     \undefined \def \showISSN      #1{\unskip}     \fi
\ifx \showLCCN     \undefined \def \showLCCN      #1{\unskip}     \fi
\ifx \shownote     \undefined \def \shownote      #1{#1}          \fi
\ifx \showarticletitle \undefined \def \showarticletitle #1{#1}   \fi
\ifx \showURL      \undefined \def \showURL       {\relax}        \fi
\providecommand\bibfield[2]{#2}
\providecommand\bibinfo[2]{#2}
\providecommand\natexlab[1]{#1}
\providecommand\showeprint[2][]{arXiv:#2}

\bibitem[\protect\citeauthoryear{Bartz-Beielstein}{Bartz-Beielstein}{2006}]%
        {bartz2006experimental}
\bibfield{author}{\bibinfo{person}{Thomas Bartz-Beielstein}.}
  \bibinfo{year}{2006}\natexlab{}.
\newblock \bibinfo{booktitle}{\emph{Experimental Research in Evolutionary
  Computation: The New Experimentalism}}.
\newblock \bibinfo{publisher}{Springer Science \& Business Media}.
\newblock


\bibitem[\protect\citeauthoryear{Beyer and Sendhoff}{Beyer and
  Sendhoff}{2017}]%
        {BeyerS2017}
\bibfield{author}{\bibinfo{person}{Hans-Georg Beyer} {and}
  \bibinfo{person}{Bernhard Sendhoff}.} \bibinfo{year}{2017}\natexlab{}.
\newblock \showarticletitle{Simplify Your Covariance Matrix Adaptation
  Evolution Strategy}.
\newblock \bibinfo{journal}{\emph{IEEE Transactions on Evolutionary
  Computation}} \bibinfo{volume}{21}, \bibinfo{number}{5} (\bibinfo{date}{Oct}
  \bibinfo{year}{2017}), \bibinfo{pages}{746--759}.
\newblock
\showISSN{1089-778X}
\urldef\tempurl%
\url{https://doi.org/10.1109/TEVC.2017.2680320}
\showDOI{\tempurl}


\bibitem[\protect\citeauthoryear{Collange, Delattre, Hansen, Quinquis, and
  Schoenauer}{Collange et~al\mbox{.}}{2010}]%
        {collange2010multidisciplinary}
\bibfield{author}{\bibinfo{person}{Guillaume Collange},
  \bibinfo{person}{Nathalie Delattre}, \bibinfo{person}{Nikolaus Hansen},
  \bibinfo{person}{Isabelle Quinquis}, {and} \bibinfo{person}{Marc
  Schoenauer}.} \bibinfo{year}{2010}\natexlab{}.
\newblock \showarticletitle{Multidisciplinary optimization in the design of
  future space launchers}.
\newblock \bibinfo{journal}{\emph{Multidisciplinary design optimization in
  computational mechanics}} (\bibinfo{year}{2010}), \bibinfo{pages}{459--468}.
\newblock


\bibitem[\protect\citeauthoryear{Deb}{Deb}{2000}]%
        {DEB2000}
\bibfield{author}{\bibinfo{person}{Kalyanmoy Deb}.}
  \bibinfo{year}{2000}\natexlab{}.
\newblock \showarticletitle{An efficient constraint handling method for genetic
  algorithms}.
\newblock \bibinfo{journal}{\emph{Computer Methods in Applied Mechanics and
  Engineering}} \bibinfo{volume}{186}, \bibinfo{number}{2}
  (\bibinfo{year}{2000}), \bibinfo{pages}{311 -- 338}.
\newblock
\showISSN{0045-7825}
\urldef\tempurl%
\url{https://doi.org/10.1016/S0045-7825(99)00389-8}
\showDOI{\tempurl}


\bibitem[\protect\citeauthoryear{Derrac, Garc\'{i}a, Molina, and
  Herrera}{Derrac et~al\mbox{.}}{2011}]%
        {DERRAC20113}
\bibfield{author}{\bibinfo{person}{Joaqu\'{i}n Derrac},
  \bibinfo{person}{Salvador Garc\'{i}a}, \bibinfo{person}{Daniel Molina}, {and}
  \bibinfo{person}{Francisco Herrera}.} \bibinfo{year}{2011}\natexlab{}.
\newblock \showarticletitle{A practical tutorial on the use of nonparametric
  statistical tests as a methodology for comparing evolutionary and swarm
  intelligence algorithms}.
\newblock \bibinfo{journal}{\emph{Swarm and Evolutionary Computation}}
  \bibinfo{volume}{1}, \bibinfo{number}{1} (\bibinfo{year}{2011}),
  \bibinfo{pages}{3 -- 18}.
\newblock
\showISSN{2210-6502}
\urldef\tempurl%
\url{https://doi.org/10.1016/j.swevo.2011.02.002}
\showDOI{\tempurl}


\bibitem[\protect\citeauthoryear{Hansen, M\"uller, and Koumoutsakos}{Hansen
  et~al\mbox{.}}{2003}]%
        {Hansen2003}
\bibfield{author}{\bibinfo{person}{N. Hansen}, \bibinfo{person}{S.~D.
  M\"uller}, {and} \bibinfo{person}{P. Koumoutsakos}.}
  \bibinfo{year}{2003}\natexlab{}.
\newblock \showarticletitle{Reducing the Time Complexity of the Derandomized
  Evolution Strategy with Covariance Matrix Adaptation ({CMA-ES})}.
\newblock \bibinfo{journal}{\emph{Evolutionary Computation}}
  \bibinfo{volume}{11}, \bibinfo{number}{1} (\bibinfo{date}{March}
  \bibinfo{year}{2003}), \bibinfo{pages}{1--18}.
\newblock
\showISSN{1063-6560}
\urldef\tempurl%
\url{https://doi.org/10.1162/106365603321828970}
\showDOI{\tempurl}


\bibitem[\protect\citeauthoryear{{Hellwig} and {Beyer}}{{Hellwig} and
  {Beyer}}{2018}]%
        {HellwigB2018}
\bibfield{author}{\bibinfo{person}{Michael {Hellwig}} {and}
  \bibinfo{person}{Hans-Georg {Beyer}}.} \bibinfo{year}{2018}\natexlab{}.
\newblock \showarticletitle{A Matrix Adaptation Evolution Strategy for
  Constrained Real-Parameter Optimization}. In \bibinfo{booktitle}{\emph{2018
  IEEE Congress on Evolutionary Computation (CEC)}}. \bibinfo{pages}{1--8}.
\newblock
\urldef\tempurl%
\url{https://doi.org/10.1109/CEC.2018.8477950}
\showDOI{\tempurl}


\bibitem[\protect\citeauthoryear{Hellwig and Beyer}{Hellwig and Beyer}{2019}]%
        {HellwigB2019a}
\bibfield{author}{\bibinfo{person}{Michael Hellwig} {and}
  \bibinfo{person}{Hans-Georg Beyer}.} \bibinfo{year}{2019}\natexlab{}.
\newblock \showarticletitle{Benchmarking evolutionary algorithms for single
  objective real-valued constrained optimization -- A critical review}.
\newblock \bibinfo{journal}{\emph{Swarm and Evolutionary Computation}}
  \bibinfo{volume}{44} (\bibinfo{year}{2019}), \bibinfo{pages}{927 -- 944}.
\newblock
\showISSN{2210-6502}
\urldef\tempurl%
\url{https://doi.org/10.1016/j.swevo.2018.10.002}
\showDOI{\tempurl}


\bibitem[\protect\citeauthoryear{Loshchilov, Glasmachers, and Beyer}{Loshchilov
  et~al\mbox{.}}{2018}]%
        {LoshchilovGlasmachersBeyer2018}
\bibfield{author}{\bibinfo{person}{Ilya Loshchilov}, \bibinfo{person}{Tobias
  Glasmachers}, {and} \bibinfo{person}{Hans-Georg Beyer}.}
  \bibinfo{year}{2018}\natexlab{}.
\newblock \showarticletitle{Large Scale Black-box Optimization by
  Limited-Memory Matrix Adaptation}.
\newblock \bibinfo{journal}{\emph{IEEE Transactions on Evolutionary
  Computation}}  \bibinfo{volume}{99} (\bibinfo{year}{2018}).
\newblock


\bibitem[\protect\citeauthoryear{Mora and Squillero}{Mora and
  Squillero}{2015}]%
        {mora2015applications}
\bibfield{author}{\bibinfo{person}{Antonio~M Mora} {and}
  \bibinfo{person}{Giovanni Squillero}.} \bibinfo{year}{2015}\natexlab{}.
\newblock \bibinfo{booktitle}{\emph{Applications of Evolutionary Computation:
  18th European Conference, EvoApplications 2015, Copenhagen, Denmark, April
  8-10, 2015, Proceedings}}. Vol.~\bibinfo{volume}{9028}.
\newblock \bibinfo{publisher}{Springer}.
\newblock


\bibitem[\protect\citeauthoryear{Sutton, Lunacek, and Whitley}{Sutton
  et~al\mbox{.}}{2007}]%
        {SuttonLW2007}
\bibfield{author}{\bibinfo{person}{Andrew~M Sutton}, \bibinfo{person}{Monte
  Lunacek}, {and} \bibinfo{person}{L~Darrell Whitley}.}
  \bibinfo{year}{2007}\natexlab{}.
\newblock \showarticletitle{Differential evolution and non-separability: using
  selective pressure to focus search}. In \bibinfo{booktitle}{\emph{Proceedings
  of the 9th annual conference on Genetic and evolutionary computation}}. ACM,
  \bibinfo{pages}{1428--1435}.
\newblock


\bibitem[\protect\citeauthoryear{Takahama and Sakai}{Takahama and
  Sakai}{2006}]%
        {TakahamaS06}
\bibfield{author}{\bibinfo{person}{Tetsuyuki Takahama} {and}
  \bibinfo{person}{Setsuko Sakai}.} \bibinfo{year}{2006}\natexlab{}.
\newblock \showarticletitle{Constrained Optimization by the $\epsilon$
  Constrained Differential Evolution with Gradient-Based Mutation and Feasible
  Elites}. In \bibinfo{booktitle}{\emph{{IEEE} International Conference on
  Evolutionary Computation, {CEC} 2006, part of {WCCI} 2006, Vancouver, BC,
  Canada, 16-21 July 2006}}. \bibinfo{pages}{1--8}.
\newblock
\urldef\tempurl%
\url{https://doi.org/10.1109/CEC.2006.1688283}
\showDOI{\tempurl}


\bibitem[\protect\citeauthoryear{Takahama and Sakai}{Takahama and
  Sakai}{2010}]%
        {TakahamaS10}
\bibfield{author}{\bibinfo{person}{Tetsuyuki Takahama} {and}
  \bibinfo{person}{Setsuko Sakai}.} \bibinfo{year}{2010}\natexlab{}.
\newblock \showarticletitle{Constrained optimization by the epsilon constrained
  differential evolution with an archive and gradient-based mutation}. In
  \bibinfo{booktitle}{\emph{Proceedings of the {IEEE} Congress on Evolutionary
  Computation, {CEC} 2010, Barcelona, Spain, 18-23 July 2010}}.
  \bibinfo{pages}{1--9}.
\newblock
\urldef\tempurl%
\url{https://doi.org/10.1109/CEC.2010.5586484}
\showDOI{\tempurl}


\bibitem[\protect\citeauthoryear{Wu, Mallipeddi, and Suganthan}{Wu
  et~al\mbox{.}}{2017}]%
        {CEC2017}
\bibfield{author}{\bibinfo{person}{Guohua Wu}, \bibinfo{person}{Rammohan
  Mallipeddi}, {and} \bibinfo{person}{Ponnuthurai~N. Suganthan}.}
  \bibinfo{year}{2017}\natexlab{}.
\newblock \bibinfo{title}{{Problem Definitions and Evaluation Criteria for the
  {CEC} 2017 Competition on Constrained Single Objective Real-Parameter
  Optimization}}.
\newblock
\newblock
\urldef\tempurl%
\url{https://github.com/P-N-Suganthan/CEC2017}
\showURL{%
\tempurl}
\newblock
\shownote{Technical Report, Nanyang Technological University, Singapore.}


\bibitem[\protect\citeauthoryear{Wu, Mallipeddi, and Suganthan}{Wu
  et~al\mbox{.}}{2018}]%
        {CEC2018comp}
\bibfield{author}{\bibinfo{person}{Guohua Wu}, \bibinfo{person}{Rammohan
  Mallipeddi}, {and} \bibinfo{person}{Ponnuthurai~N. Suganthan}.}
  \bibinfo{year}{2018}\natexlab{}.
\newblock \bibinfo{title}{Competition on Constrained Real Parameter
  Optimization (2017 \& 2018)}.
\newblock \bibinfo{howpublished}{IEEE CEC2018 competition results, Files
  available on GitHub}.
\newblock
\urldef\tempurl%
\url{https://github.com/P-N-Suganthan/CEC2018}
\showURL{%
\tempurl}


\bibitem[\protect\citeauthoryear{Zhang, h.~Zhan, Lin, Chen, j.~Gong, h.~Zhong,
  Chung, Li, and h.~Shi}{Zhang et~al\mbox{.}}{2011}]%
        {Zhang2011}
\bibfield{author}{\bibinfo{person}{J. Zhang}, \bibinfo{person}{Z. h. Zhan},
  \bibinfo{person}{Y. Lin}, \bibinfo{person}{N. Chen}, \bibinfo{person}{Y. j.
  Gong}, \bibinfo{person}{J. h. Zhong}, \bibinfo{person}{H.~S.~H. Chung},
  \bibinfo{person}{Y. Li}, {and} \bibinfo{person}{Y. h. Shi}.}
  \bibinfo{year}{2011}\natexlab{}.
\newblock \showarticletitle{Evolutionary Computation Meets Machine Learning: A
  Survey}.
\newblock \bibinfo{journal}{\emph{IEEE Computational Intelligence Magazine}}
  \bibinfo{volume}{6}, \bibinfo{number}{4} (\bibinfo{date}{Nov}
  \bibinfo{year}{2011}), \bibinfo{pages}{68--75}.
\newblock
\showISSN{1556-603X}
\urldef\tempurl%
\url{https://doi.org/10.1109/MCI.2011.942584}
\showDOI{\tempurl}


\end{thebibliography}
